%% file: main.tex
\documentclass[10pt,twocolumn,letterpaper]{article}

\usepackage[pagenumbers]{cvpr} %

\input{preamble}

\definecolor{cvprblue}{rgb}{0.21,0.49,0.74}
\usepackage[pagebackref,breaklinks,colorlinks,allcolors=cvprblue]{hyperref}

\title{\vspace{-0.6cm} \includegraphics[height=2\fontcharht\font`f]{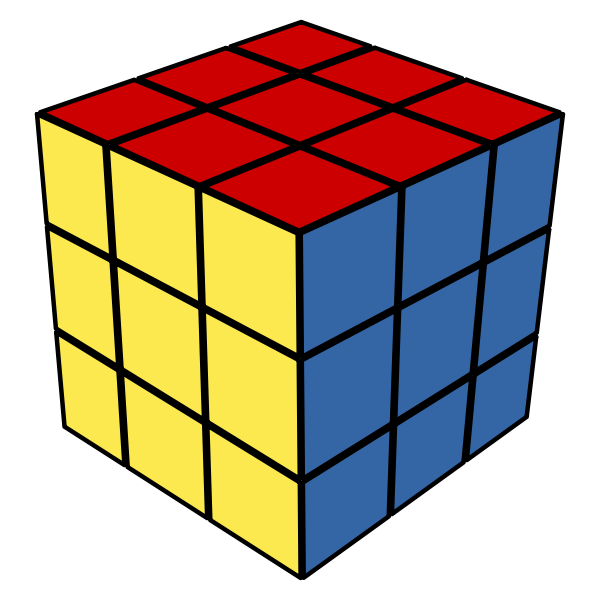} \BN: A Structured Benchmark for Image Matching \\ across Geometric Challenges
\vspace{-0.5cm}
}

\author{
Thibaut Loiseau$^{\dagger}$
\hspace{20pt} 
Guillaume Bourmaud$^{\ddagger}$
\\
$^{\dagger}$ LIGM, Ecole des Ponts, Université Gustave Eiffel, CNRS, France \\
$^{\ddagger}$ Laboratoire IMS, Université de Bordeaux, France \\
{\tt\small thibaut.loiseau@enpc.fr} 
\hspace{5pt}
{\tt\small guillaume.bourmaud@u-bordeaux.fr}
}

\begin{document}

\twocolumn[{
    \renewcommand\twocolumn[1][]{#1}   
    \maketitle
    \vspace{-0.8cm}
    \centering 
    \includegraphics[width=0.95\linewidth]{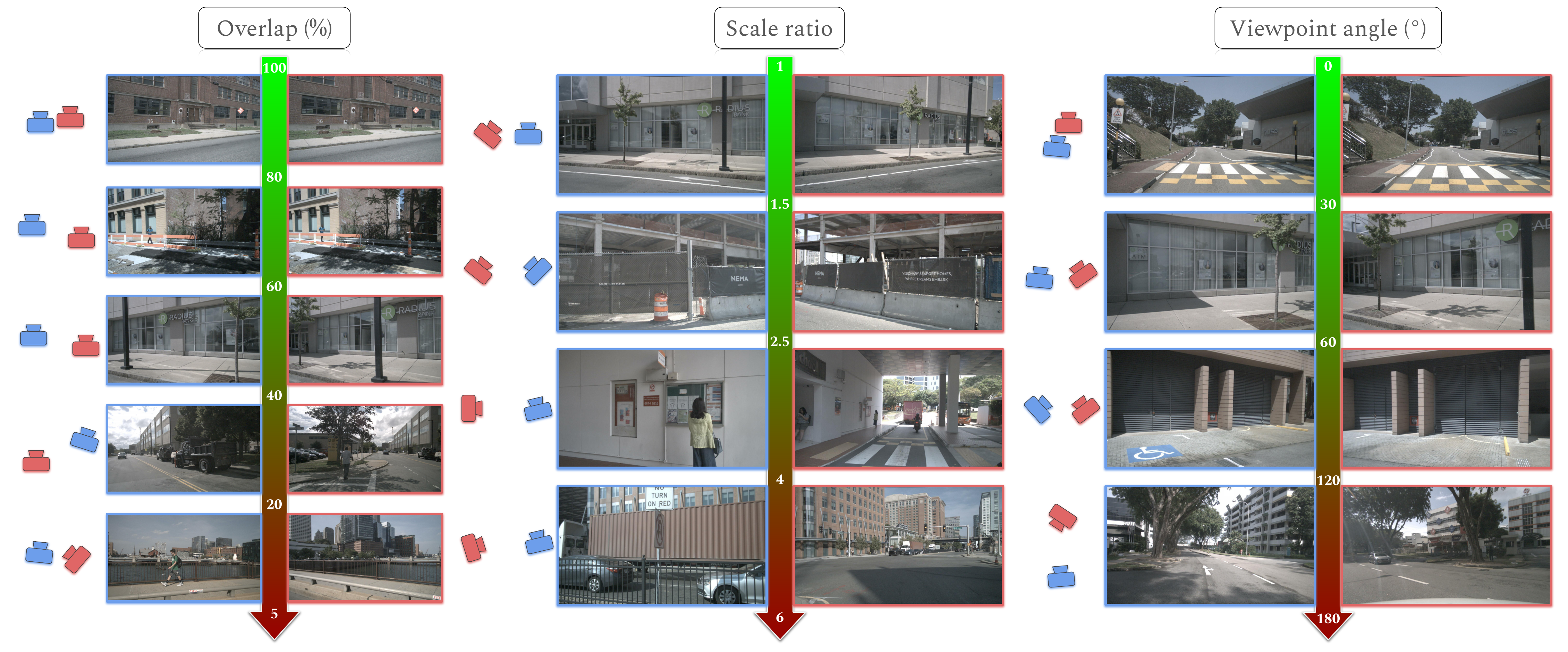}
    \vspace{-0.2cm}
    \captionof{figure}{
    \textbf{We introduce \BN{} -- }a benchmark based on the images from nuScenes for fine grain evaluation of camera pose estimations methods. \BN is made of image pairs spanning three difficulty criteria, in terms of scene overlap, scale ratio, and difference of viewpoint angles. 
    It contains 16.5K image pairs across 33 difficulty levels. We use it to provide a comprehensive benchmarking of 14 methods.}
    \label{fig:teaser}
    \vspace{0.3cm}
}]

\input{sec/0_abstract}    
\input{sec/1_intro}
\input{sec/2_related}

\input{sec/3_benchmark}
\input{sec/4_results}
\input{sec/5_conclusion}

{
    \small
    \bibliographystyle{ieeenat_fullname}
    \bibliography{main}
}

\input{sec/X_suppl}

\end{document}

%% file: preamble.tex
\usepackage{xspace}
\def\BN{RUBIK\xspace} 

\definecolor{mygreen}{RGB}{182,215,168}
\definecolor{myorange}{RGB}{249,203,156}
\definecolor{mygrey}{RGB}{102,102,102}
\definecolor{myred}{RGB}{224,102,102}
\definecolor{myblue}{RGB}{109,158,235}

\usepackage{pgfplots}
\pgfplotsset{compat=1.18}
\usepackage{tikz}
\usepackage[dvipsnames]{xcolor}

\def\m#1{\ensuremath{\mathtt{#1}}}

\def\vec#1{\ensuremath{\mathbf{#1}}}
\def\vp{\vec{p}}
\def\vt{\vec{t}}
\def\mr{\m{R}}
\def\tr{^{\top}}

%% file: sec/0_abstract.tex
\begin{abstract}
Camera pose estimation is crucial for many computer vision applications, yet existing benchmarks offer limited insight into method limitations across different geometric challenges. We introduce \BN, a novel benchmark that systematically evaluates image matching methods across well-defined geometric difficulty levels. Using three complementary criteria - overlap, scale ratio, and viewpoint angle - we organize 16.5K image pairs from nuScenes into 33 difficulty levels. Our comprehensive evaluation of 14 methods reveals that while recent detector-free approaches achieve the best performance ($>$47\% success rate), they come with significant computational overhead compared to detector-based methods (150-600ms vs. 40-70ms). Even the best performing method succeeds on only 54.8\% of the pairs, highlighting substantial room for improvement, particularly in challenging scenarios combining low overlap, large scale differences, and extreme viewpoint changes.
Benchmark will be made publicly available.
\end{abstract}

%% file: sec/1_intro.tex
\section{Introduction}
\label{sec:intro}
\vspace{-0.2cm}
Camera pose estimation is a cornerstone of many computer vision applications, including augmented reality~\cite{azuma2001recent}, robotics~\cite{cadena2016past}, 3D reconstruction~\cite{schonberger2016structure,wu2013towards,heinly2015reconstructing}, and autonomous navigation~\cite{taira2018inloc,mur2015orb,mur2017orb}.
Camera pose estimators often rely on state-of-the-art image matching methods~\cite{potje2024xfeat,edstedt2024dedode,lindenberger2023lightglue,zhao2023aliked, sun2021loftr,wang2024efficient,edstedt2024roma,chen2022aspanformer,wang2024dust3r,leroy2024grounding, wang2022matchformer, giang2023topicfm, mao20223dg, tangquadtree, dusmanu2019d2, revaud2019r2d2, edstedt2023dkm, fan2023occ, chen2021learning, yin2024srpose, gleize2023silk, hedlin2024unsupervised, kim2025learning, germainvisual, germain2020s2dnet, germain2021neural, jiang2021cotr, tan2022eco}, which have achieved great performance in challenging scenarios, such as occlusions, limited overlap, and significant viewpoint changes.
The creation of benchmarks~\cite{toft2020long,sattler2018benchmarking,zhang2021reference,taira2018inloc,arnold2022map} significantly contributed to these advancements by allowing a fair comparison between the proposed methods and pushing the development for more performing methods.

In this paper, we propose \BN, a benchmark based on the images from the nuScenes dataset~\cite{caesar2020nuscenes}, specifically designed to provide a more granular understanding of the limitations of current  methods compared to existing benchmarks. Such understanding is critical to identify the weaknesses and to keep improving the performance of camera pose estimation methods.
The nuScenes dataset offers an important diversity, featuring both broad and narrow streets, large and small buildings, as well as vegetation and rivers. Additionally, the images were taken by cameras mounted on a car, oriented in multiple directions. As a result, each scene was captured from numerous viewpoints and at varying distances, making these images an ideal testbed for camera pose estimation.

More exactly, we structured \BN along 3 different types of challenges, as illustrated in~\cref{fig:teaser}. We quantified these challenges in terms of (1) overlap percentage between the two input views, (2) difference of apparent scale between the views, and (3) the difference of view angles: Small overlaps, large scale ratios, and large perspective differences make estimating the camera motion between the images challenging, and can happen alone or simultaneously. Let us highlight that we specifically focus on scenes recorded in good weather conditions to isolate and evaluate geometric difficulties without the confounding effect of adverse weathers.

\noindent We design \BN as follows:
\begin{itemize}
    \item \textbf{Camera registration -- }We carefully estimate the camera poses for the images in nuScenes. The nuScenes dataset already provides the camera poses but only within the ground plane, and we thus used COLMAP~\cite{schonberger2016structure} to obtain full 6 degrees of freedom (DoF) poses. We still use the nuScenes 3~DoF camera poses to ensure the recovered 6~DoF poses are correct. 
    
    \item \textbf{Dense co-visibility maps -- }For each pair of nuScenes images from the same scene, we estimate co-visibility maps, as illustrated in~\cref{fig:pipeline_dataset}. This gives us a fine measure of the co-visible regions between the two images. To do so, we developed a surprisingly simple and efficient method using the camera poses for each image pair and their depth and normal maps as predicted by state-of-the-art monocular depth estimators~\cite{piccinelli2024unidepth,yang2024depth}. 
    
    \item \textbf{Difficulty criteria -- }For each pair, we evaluate our three criteria---overlap, ratio of the distances to the scene, and viewpoint angle difference---to quantify the difficulty of estimating the relative pose of a given image pair. Examples of pairs and their criteria are shown in~\cref{fig:teaser}. We then quantize the range of each criterion into a few bins, which results in a 3D grid of 33 boxes with varying levels of difficulty. Each box is populated with 500 image pairs, for a total of 16.5K test pairs.
    
    \item \textbf{Comprehensive benchmarking -- }\BN allows us to provide an extensive evaluation of 14 methods: SIFT~\cite{lowe2004distinctive}, SuperPoint~\cite{detone2018superpoint}, ALIKED~\cite{zhao2023aliked}, DISK~\cite{tyszkiewicz2020disk}, all using the LightGlue matcher~\cite{lindenberger2023lightglue}, XFeat and its variants XFeat* and XFeat-LighterGlue~\cite{potje2024xfeat}, DeDoDe v2~\cite{edstedt2024dedode}, LoFTR~\cite{sun2021loftr}, ASpanFormer~\cite{chen2022aspanformer}, RoMa~\cite{edstedt2024roma}, Efficient LoFTR~\cite{wang2024efficient}, DUSt3R~\cite{wang2024dust3r} and MASt3R~\cite{leroy2024grounding}.
    Our results show that while most methods accurately estimate the poses under high overlap, similar scale, and small relative viewpoint angle, even the best performing method struggles to correctly estimate the pose for more than 45\% of the image pairs for a threshold of 5° (rotation) and 2m (translation). These findings highlight \BN's utility in assessing and comparing different approaches.
\end{itemize}

We believe that \BN will serve as a valuable resource for the computer vision community, encouraging the development of more robust, occlusion-aware, or curriculum learning-based camera pose estimation methods. By providing both a comprehensive dataset and demonstrating the practical benefits of our co-visibility maps, we aim to advance research in this critical area of computer vision.

%% file: sec/2_related.tex
\section{Related Work}
\label{sec:related}

\begin{figure*}[ht]
    \centering
    \includegraphics[width=\textwidth]{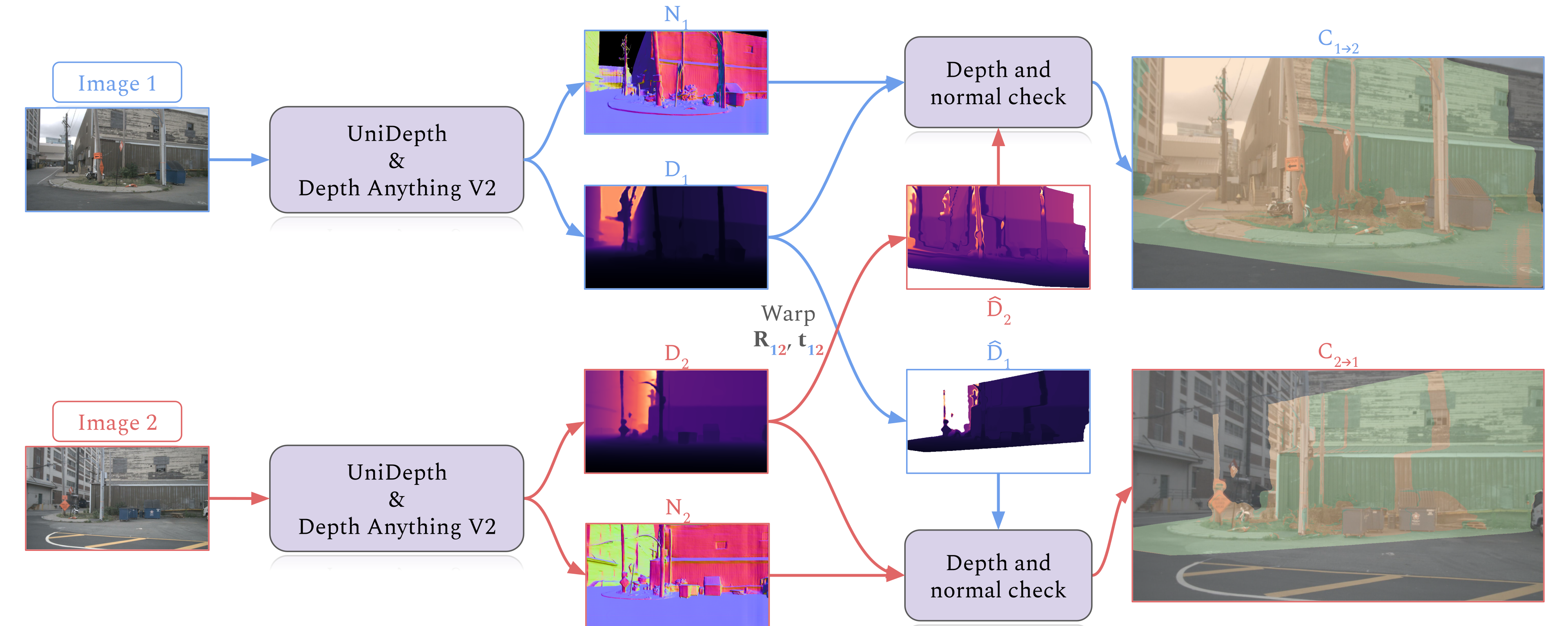}
    \caption{\textbf{Dense co-visibility map estimation -- }Using normal maps ($\textcolor{myblue}{\m{N_1}}$, $\textcolor{myred}{\m{N_2}}$) and depth maps ($\textcolor{myblue}{\m{D_1}}$, $\textcolor{myred}{\m{D_2}}$) along with relative camera poses ($\m{\textcolor{mygrey}{R}_{\textcolor{myblue}{1}\textcolor{myred}{2}}}$, $\m{\textcolor{mygrey}{t}_{\textcolor{myblue}{1}\textcolor{myred}{2}}}$), we warp depth maps between views to obtain $\textcolor{myblue}{\hat{\m{D}}_1}$ and $\textcolor{myred}{\hat{\m{D}}_2}$. Geometric consistency checks classify pixels as \textcolor{mygreen}{co-visible}, \textcolor{myorange}{occluded}, or \textcolor{mygrey}{outside field-of-view} to obtain the co-visibilty maps $\textcolor{myblue}{\m{C}_{1\rightarrow 2}}$ and $\textcolor{myred}{\m{C}_{2\rightarrow 1}}$ (see~\cref{ssec:covis_gen}). We use UniDepth~\cite{piccinelli2024unidepth} for metric depth estimation and Depth Anything V2~\cite{yang2024depth} for normal map computation.}
    \label{fig:pipeline_dataset}
\end{figure*}

\subsection{Image Matching Benchmarks}
Several benchmarks have been proposed to evaluate image matching methods. HPatches~\cite{balntas2017hpatches} focuses on homography estimation under viewpoint and illumination changes. MegaDepth1500 and ScanNet1500 are two widely used benchmarks initially proposed in~\cite{sarlin2020superglue}. MegaDepth1500 randomly sampled 1,500 image pairs from scenes “Sacre Coeur” and “St. Peter’s Square” of MegaDepth~\cite{li2018megadepth}, discarding pairs with too small or too large overlap. ScanNet1500 randomly sampled 1,500 test pairs from ScanNet~\cite{dai2017scannet} similarly. The KITTI~\cite{geiger2012we}  dataset is also frequently used~\cite{jau2020deep}, where 2,710 image pairs, from consecutive frames, are sampled from the two sequences 09-10.
The Image Matching Challenge 2024~\cite{image-matching-challenge-2024} and its previous occurrences represent a significant advancement in comprehensive evaluation, featuring six distinct categories that cover real-world challenges: from phototourism with varying viewpoints and temporal changes, to aerial-ground matching, repeated structures, natural environments, and challenging scenarios with transparencies and reflections. 

\subsection{Visual Localization Benchmarks}
Aachen Day-Night~\cite{sattler2018benchmarking,zhang2021reference} is a visual localization benchmark addressing outdoor localization in changing conditions. It consists of 4,328 sparsely sampled daytime database images, 824 daytime query images and 98 nighttime query images taken in the same environment.

InLoc~\cite{taira2018inloc} is a visual localization benchmark addressing large scale indoor localization with illumination and long-term changes, as well as repetitive patterns. It consists of 9,972 database images and 329 query images.

The Map-free Relocalization~\cite{arnold2022map} benchmark consists of 655 small places, where each place comes with a reference image. The benchmark features changing conditions and image pairs with low to no visual overlap.

Despite these advances, existing benchmarks often lack controlled geometric variations, making it difficult to systematically analyze method performance across different difficulty levels. Our \BN benchmark addresses this limitation by providing 16.5K test pairs organized according to well-defined geometric criteria to obtain controlled varying levels of difficulty.

%% file: sec/3_benchmark.tex
\section{\BN}
\label{sec:benchmark}
Our \BN benchmark is based on images from the nuScenes dataset~\cite{caesar2020nuscenes}. nuScenes is a large-scale autonomous driving dataset containing 1,000 driving scenes in Boston and Singapore, each 20 seconds long, recorded at 12Hz in various weather conditions and times of day. The dataset provides high-quality synchronized camera images from 6 cameras with complete 360° coverage, along with precise camera calibration and pose information. 

For our benchmark, we specifically focus on scenes recorded in good weather conditions to isolate and evaluate geometric difficulties without the confounding effect of adverse weathers. This deliberate choice allows us to systematically analyze how methods perform across different geometric challenges, without the additional complexity of environmental factors.

The creation of \BN consists of four main steps: (1) lifting of nuScenes ground truth 3~DoF camera poses to 6 DoF, (2) generation of depth and surface normal maps for each image in order to (3) compute the co-visibility maps between image pairs, which allow us to (4) systematically organize the image pairs based on geometric criteria.

\subsection{Lifting nuScenes ground truth camera poses}\label{sec:lifting}

While nuScenes provides high-quality metric camera poses, these are limited to 3 DoF within the ground plane, as they are primarily intended for autonomous driving applications. However, for comprehensive camera pose estimation benchmarking, we require full 6~DoF metric camera poses that account for variation in camera height and orientation.
To lift nuScenes ground truth 3 DoF metric poses to 6 DoF metric poses, we carefully process each sequence independently using a two-stage approach:
\\
\\
\noindent \textbf{1. 3D Reconstruction -- }We first perform Structure-from-Motion using COLMAP~\cite{schonberger2016structure}.
This provides us with initial 6-DoF camera pose estimates. However, the translations are not metric (i.e. the scene is reconstructed up to a scale factor) and some camera poses may be erroneous.

\noindent \textbf{2. Pose alignment and filtering -- }To obtain metric translations and filter out erroneous poses, we align the previously estimated 6 DoF poses with the nuScenes ground truth 3 DoF metric poses.
To do so, we use a custom LO-RANSAC~\cite{chum2003locally} approach to estimate a 7 DoF similarity transformation between the projected COLMAP poses and nuScenes ground truth poses.
We set the RANSAC threshold to 1 meter to filter out erroneous poses and ensure high-quality ground truth.

\begin{figure}[ht]
    \centering
    \includegraphics[width=0.9\linewidth]{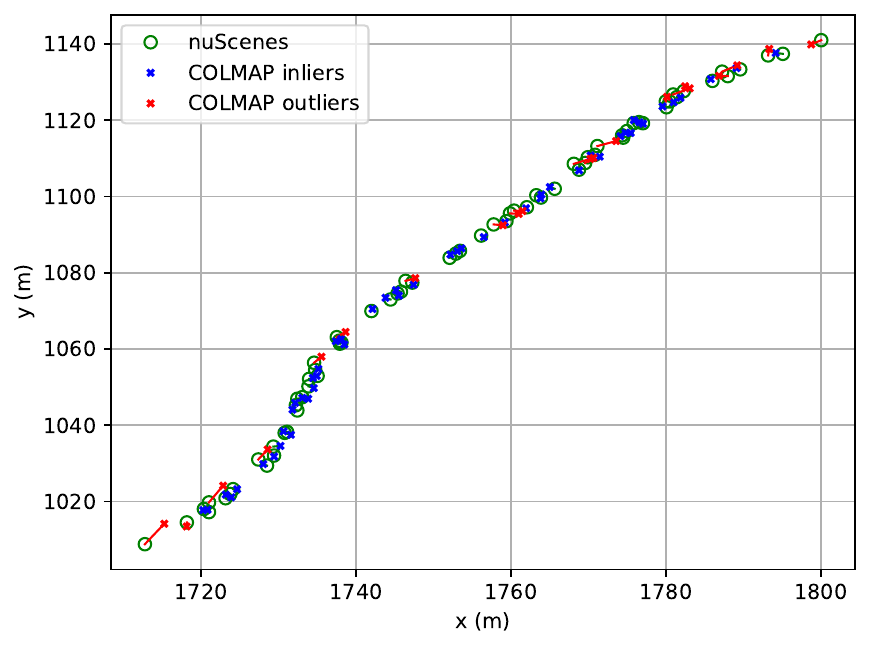}
    \vspace{-0.25cm}
    \caption{\textbf{Camera pose alignment and filtering -- }Visualization of (subsampled) camera trajectories of scene-0266 after aligning COLMAP poses with nuScenes ground truth poses.
    Blue crosses ($\textcolor{blue}{\times}$) indicate inlier poses (alignment error $<$ 1m) that are kept for our benchmark, while red crosses ($\textcolor{red}{\times}$) show outlier poses that are discarded.}
    \label{fig:pose_alignment}

\end{figure}

An example of alignment is shown in~\cref{fig:pose_alignment}.
This alignment with nuScenes' metric ground truth allows us to recover proper metric scale, which is not available from COLMAP reconstructions alone, and consequently maintain the high precision of nuScenes' original poses while adding reliable elevation and orientation information.
The resulting 6~DoF metric poses serve as the foundation for our geometric difficulty criteria and co-visibility map computations.

\subsection{Generation of metric depth and normal maps}\label{sec:depth_generation}
\begin{figure*}[ht]
    \centering
    \begin{subfigure}{.33\textwidth}
        \centering
        \fbox{\includegraphics[width=0.9\linewidth]{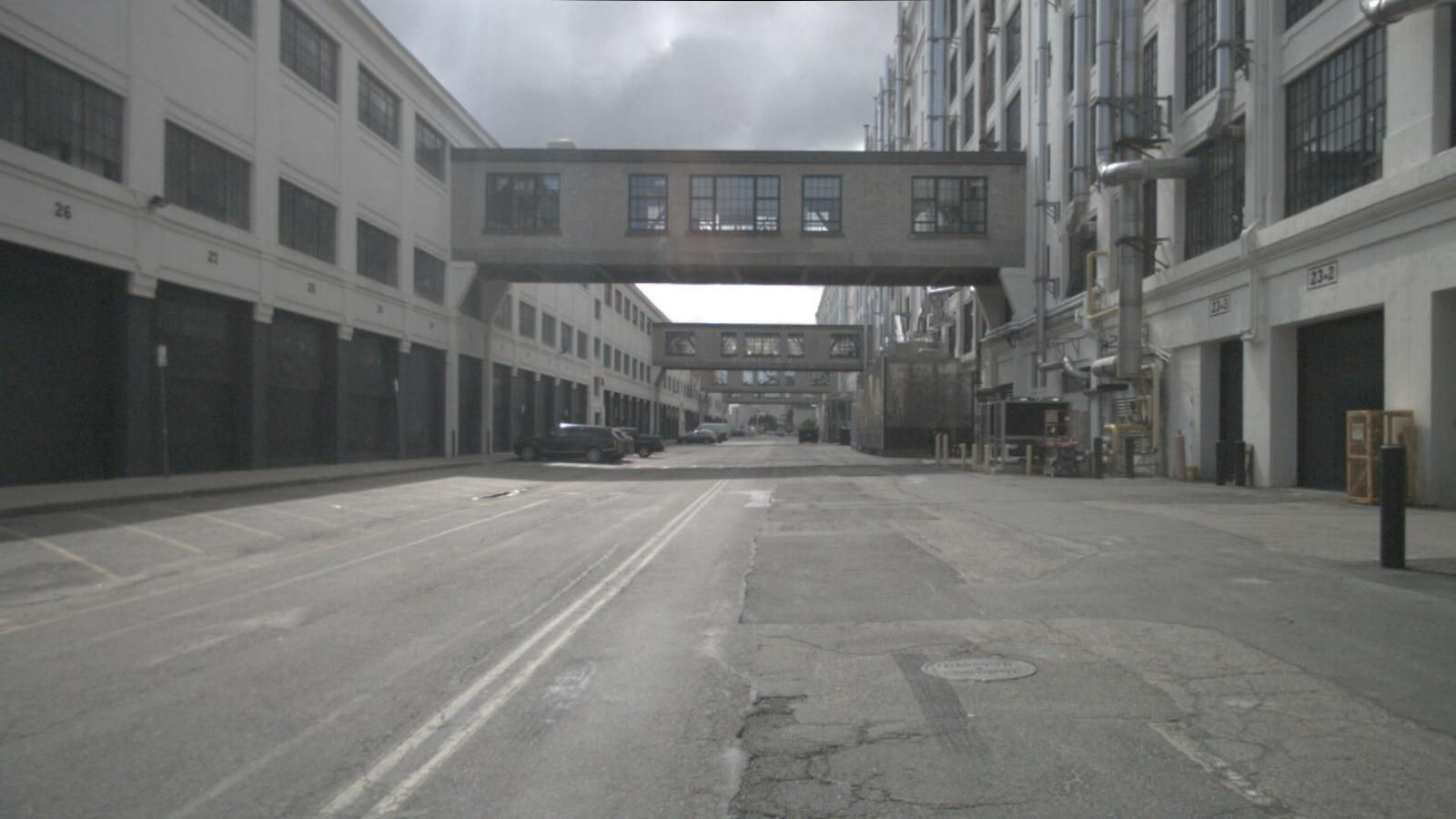}}        
        \caption{Input image}
        \label{fig:image_normals}
    \end{subfigure}
    \begin{subfigure}{.33\textwidth}
        \centering
        \fbox{\includegraphics[width=0.9\linewidth]{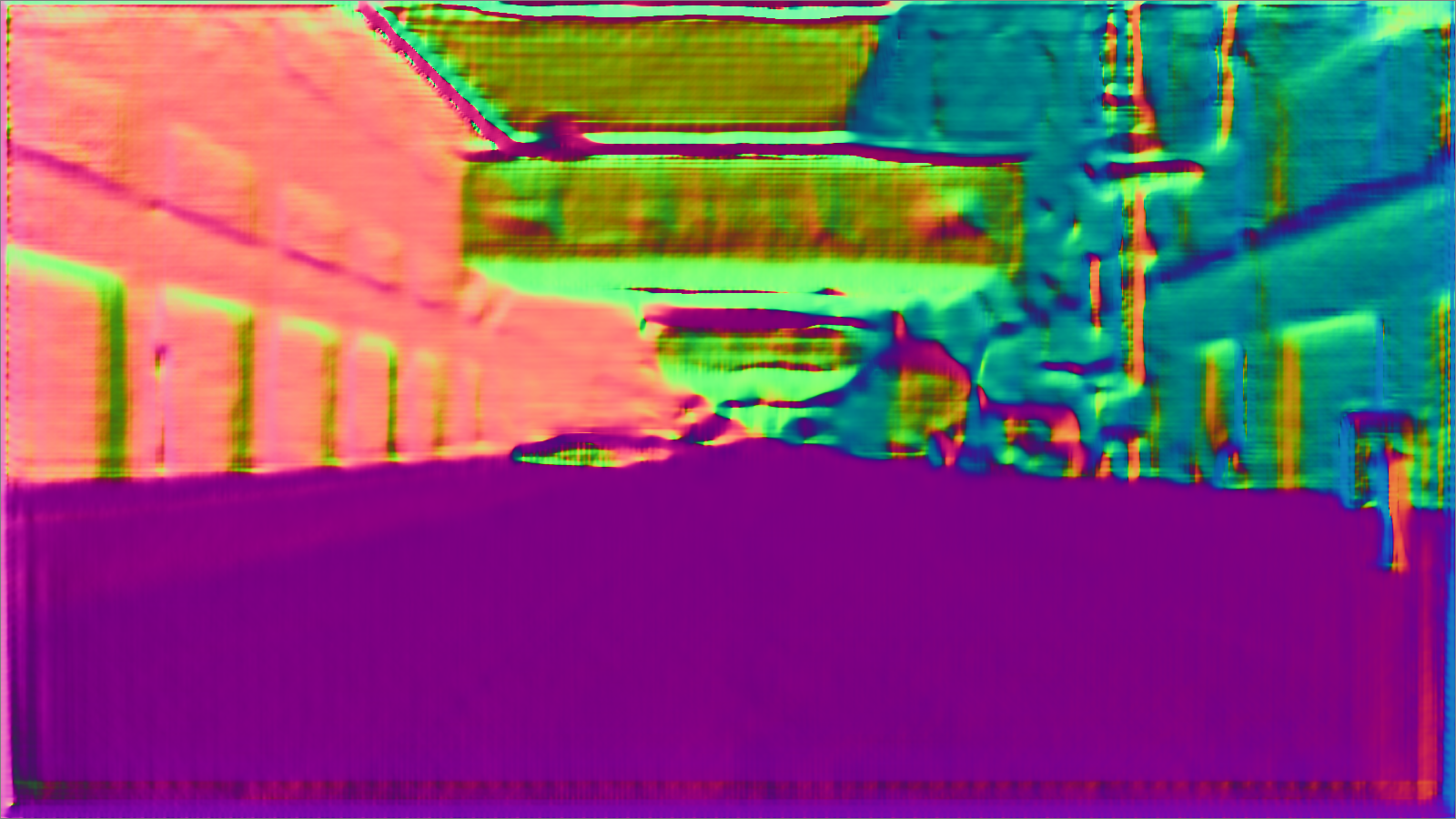}}
        \caption{Normals from UniDepth~\cite{piccinelli2024unidepth}}
        \label{fig:unidepth_normals}
    \end{subfigure}
    \begin{subfigure}{.33\textwidth}
        \centering
        \fbox{\includegraphics[width=0.9\linewidth]{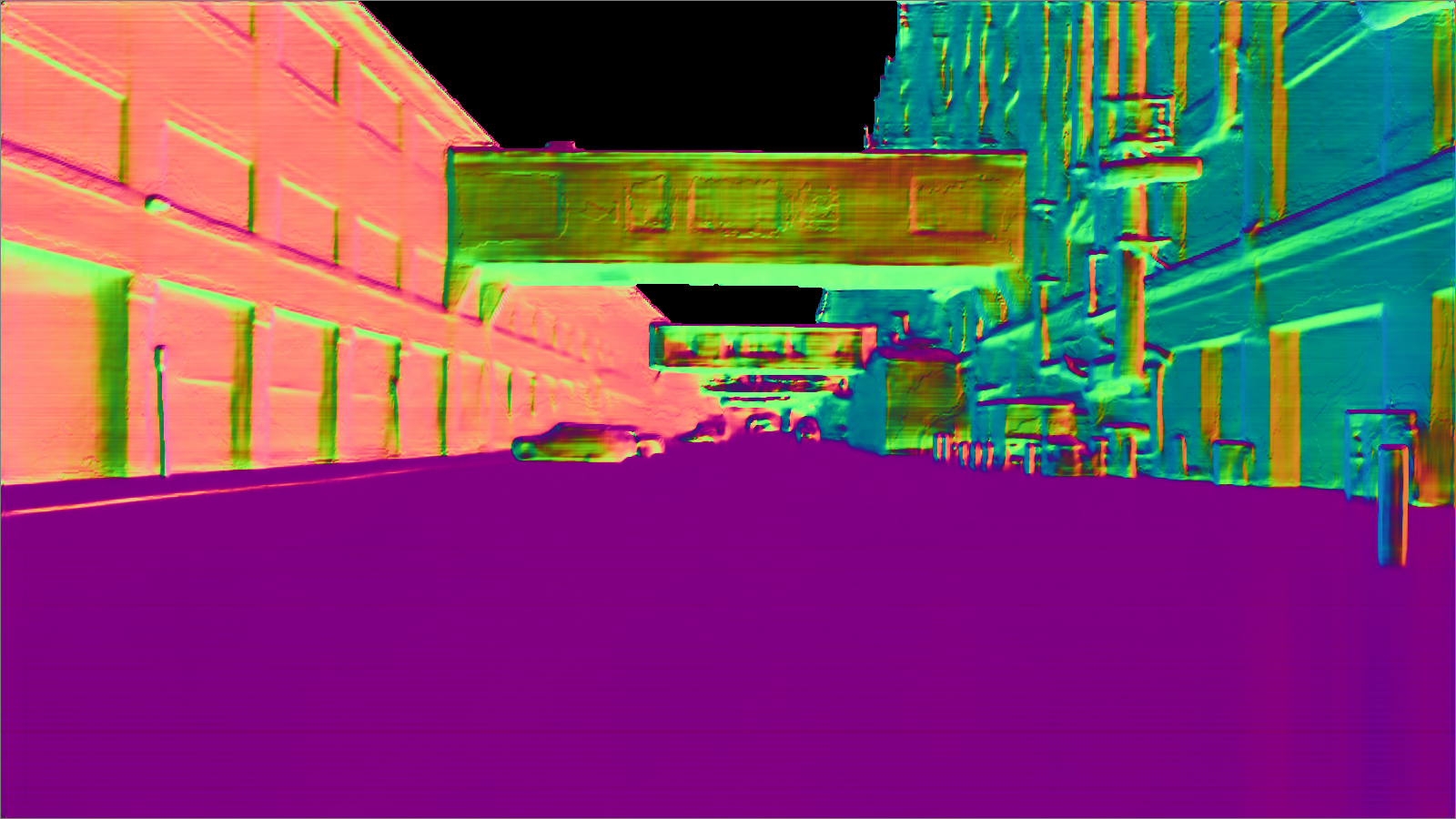}}
        \caption{Normals from DAv2~\cite{yang2024depth}}
        \label{fig:dav2_normals}
    \end{subfigure}
    \caption{\textbf{Comparison of surface normal maps -- }From left to right: input image (a), normal maps computed from UniDepth's metric depth predictions (b) and from Depth Anything V2 after alignment to UniDepth depth map (c). Note the significantly sharper object boundaries and finer geometric details in Depth Anything V2's prediction, particularly around building edges and depth discontinuities.}
    \label{fig:normal_comparison}
\end{figure*}

To compute dense co-visibility maps between image pairs, we require accurate depth and normal maps for each image. Recent advances in monocular depth estimation have enable reliable geometric information extraction from single images without expensive ground truth measurements. After evaluating several state-of-the-art models, we found that combining two complementary approaches yield optimal results:

\begin{itemize}
    \item \textbf{Metric Depth Maps -- }We use UniDepth~\cite{piccinelli2024unidepth} for its ability to predict metric depth values. The model provides well-aligned depth predictions that are crucial for consistent cross-view measurements.

    \item \textbf{Surface Normal Maps -- }Normal maps computed from UniDepth~\cite{piccinelli2024unidepth} depth predictions tend to be noisy. Instead, we compute normal maps from Depth Anything V2~\cite{yang2024depth} depth maps. Let us highlight that these depth maps are not metric, thus we align them with Unidepth depth maps before normals computation. This approach produces remarkably sharp normal maps with precise object boundaries and fine geometric details (see \cref{fig:normal_comparison} for a comparison).
\end{itemize}

\noindent This complementary approach leverages each model's strengths: UniDepth's accuracy and Depth Anything's superior normal predictions. Our experiments show that this combination provides reliable geometric information for computing co-visibility maps between views, as demonstrated in the following section.

\subsection{Generation of Co-visibility maps}
\label{ssec:covis_gen}

\begin{figure}
    \centering
    \includegraphics[width=0.9\linewidth]{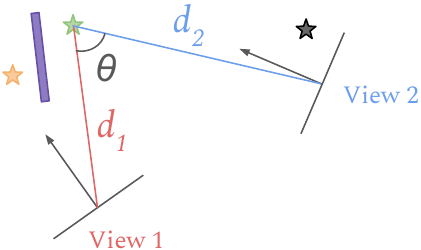}
    \caption{\textbf{Two-view setup -- }Considering two views, a 3D point can either be co-visible ($\textcolor{mygreen}{\bigstar}$), occluded ($\textcolor{myorange}{\bigstar}$), or outside the field of view ($\textcolor{mygrey}{\bigstar}$) in one of the views. For each co-visible 3D point, we compute its distances $\textcolor{myred}{\m{d_1}}$ and $\textcolor{myblue}{\m{d_2}}$ to both camera centers and the angle ${\pmb{\theta}}$ between the two lines of sight.}
    \label{fig:cov_occ_out}
\end{figure}

Given a pair of images $(\m{I}_1, \m{I}_2)$ with known relative camera pose $(\mr_{12}, \vt_{12})$ from~\cref{sec:lifting}, calibration matrices $(\m{K}_1, \m{K}_2)$, depth maps $(\m{D}_1, \m{D}_2)$, and surface normals $(\m{N}_1, \m{N}_2)$ from~\cref{sec:depth_generation}, we generate the co-visibility maps $\m{C}_{1\rightarrow 2}$ and $\m{C}_{2\rightarrow 1}$ (see Fig.~\ref{fig:pipeline_dataset}).
We start by warping the depth map $\m{D}_2$ to predict ${\m{D}}_{1}$ as follows:
 \begin{equation}
        \hat{\m{D}}_{1}(\vp_1) = \left[ 0\, 0\, 1\right] \left( \mr_{12}\m{D}_2(\vp_{1\rightarrow 2})\m{K}_2^{-1}\vp_{1\rightarrow 2} + \vt_{12} \right),
  \end{equation}
where $\vp_1$ is a pixel location (in homogeneous coordinates) in the grid of $\m{I}_1$, $\vp_{1\rightarrow 2} = \m{K}_2\pi\left( \mr_{12}\tr\left(\m{D}_1(\vp_1)\m{K}_1^{-1}\vp_1 - \vt_{12}\right) \right)$ and $\m{D}_2(\vp_{1\rightarrow 2})$ is implemented using bilinear interpolation.

Predicting  ${\m{D}}_{1}$ enables occlusion detection through a relative depth check:
\begin{equation}
        \frac{|\hat{\m{D}}_{1}(\vp_1) - \m{D}_{1}(\vp_1)|}{\m{D}_{1}(\vp_1)} > \tau,
\end{equation}
where we used  $\tau = 5\%$. Let us highlight that if $\vp_{1\rightarrow 2}$ falls outside the image boundaries, then pixel $\vp_1$ is labeled "outside the field of view". 
This occlusion detection is further refined by discarding a pixel $\vp_1$ if its normal does not point towards camera 2:
    \begin{equation}
        \angle \left(\vec{z}, \mr_{12}\tr\m{N}_1(\vp_{1})\right) < 90^{\circ} - \epsilon,
    \end{equation}
where $\vec{z}=\left[ 0\, 0\, 1\right]\tr$ and we used $\epsilon = 5^\circ$.

An example of co-visibility maps $\m{C}_{1\rightarrow 2}$ and $\m{C}_{2\rightarrow 1}$ is shown in Fig.~\ref{fig:pipeline_dataset}. We perform the previous steps in both directions ($\m{I}_1 \rightarrow \m{I}_2$ and $\m{I}_2 \rightarrow \m{I}_1$). Our results surprisingly show that metric monocular depth prediction networks are, from now on, accurate enough to perform cross-view 3D reasoning. We believe this finding opens a path towards camera pose estimation frameworks beyond classical combination of image matching and 3D geometry-based minimal solver, but we leave this as future work.

\subsection{Geometric criteria}
Using the co-visibility maps $\m{C}_{1\rightarrow 2}$ and $\m{C}_{2\rightarrow 1}$ previously computed, we evaluate three complementary criteria to quantify the geometric difficulty of estimating the relative pose between an image pair ($\m{I}_1, \m{I}_2$).
\\
\\
\noindent \textbf{1. Overlap ($\omega$) -- }The ratio of co-visible pixels to total pixels:
    \begin{equation}
        \omega = \frac{|\m{C}_{1\rightarrow2}| + |\m{C}_{2\rightarrow1}|}{|\m{I}_1| + |\m{I}_2|},
    \end{equation}
    where $|\cdot|$ is the cardinal of a set. The overlap is a classical criterion that is often used by image matching methods~\cite{sarlin2020superglue, sun2021loftr, chen2022aspanformer}  to obtain a balanced training set that includes both simple pairs (with large overlaps) and challenging pairs (with small overlaps). However, this criterion alone is somewhat limited, as an image pair with a large pure rotation (i.e. where the translation is null) may result in a small overlap, even though the underlying matching problem is not particularly difficult, since the viewpoint remains unchanged.
    \\

\noindent \textbf{2. Scale Ratio ($\delta$) -- }The median of the ratios of the camera distances to the co-visible 3D points:  
    \begin{equation}
        \delta = {\mathrm{med}} \left\{ 
        \left\{ r_{\vp_1}^{1\rightarrow 2} \right\}_{\vp_1 \in \m{C}_{1\rightarrow 2}} \cup
        \left\{ r_{\vp_2}^{2\rightarrow 1}\right\}_{\vp_2 \in \m{C}_{2\rightarrow 1}}
        \right\},
    \end{equation}
    with $r_{\vp_i}^{i\rightarrow j}=\max \left( \frac{\left\Vert \m{K_i}^{-1}\vp_i \right\Vert }{\left\Vert \m{D}_i(\vp_i)\m{K}_i^{-1}\vp_i - \vt_{ij} \right\Vert },\frac{\left\Vert \m{D}_i(\vp_i)\m{K}_i^{-1}\vp_i - \vt_{ij} \right\Vert }{\left\Vert  \m{K_i}^{-1}\vp_i \right\Vert }\right)$.
    Contrary to the overlap, the scale ratio is independent of the relative rotation between
the two cameras (i.e. rotating camera 1 and camera 2 in~\cref{fig:cov_occ_out} does not affect neither $d_1$ nor $d_2$) and only depends on the 3D geometry of the scene and the relative translation.
\\

\noindent \textbf{3. Viewpoint Angle ($\theta$) -- }The median of the co-visible line-of-sight angles:
    \begin{equation}
        \theta = {\mathrm{med}} \left\{ 
        \left\{ \theta_{\vp_1}^{1\rightarrow 2} \right\}_{\vp_1 \in \m{C}_{1\rightarrow 2}} \cup
        \left\{ \theta_{\vp_2}^{2\rightarrow 1}\right\}_{\vp_2 \in \m{C}_{2\rightarrow 1}}
        \right\},
    \end{equation}
    where $\theta_{\vp_i}^{i\rightarrow j}=\angle \left(\m{K_i}^{-1}\vp_i, \m{D}_i(\vp_i)\m{K}_i^{-1}\vp_i - \vt_{ij}\right)$ represents the angle between the two lines of sight. It is clear that this criterion is also independent of the relative rotation between the two cameras and only depends on the 3D geometry and the relative translation, just like the scale ratio. 
    However, the viewpoint angle and the scale ratio complement each other, as the viewpoint angle is independent of the scale ratio (i.e. changing $d_2$ in~\cref{fig:cov_occ_out} does not affect $\theta$), and vice versa.

The three criteria discussed above complement each other and will be used in the next section to categorize the image pairs from the nuScenes test scenes based on their difficulty level.

\subsection{Benchmark Organization}
Using the three geometric criteria defined above, we can systematically organize image pairs from nuScenes test scenes according to their difficulty level. For each possible pair within a scene, we compute its overlap $\omega$, scale ratio $\delta$, and viewpoint angle $\theta$. Based on the distributions of these values across the entire test set which comprises 4.2M image pairs across 85 successfully reconstructed and filtered scenes, we define meaningful bins for each criterion:\\
\\
\noindent \textbf{Overlap (\%) -- 5 bins:} 5 - 20 - 40 - 60 - 80 - 100\\
\\
\noindent \textbf{Scale ratio -- 4 bins:} 1.0 - 1.5 - 2.5 - 4.0 - 6.0\\
\\
\noindent \textbf{Viewpoint angle (°) -- 4 bins:} 0 - 30 - 60 - 120 - 180\\

An example of image pair for each bin is shown in~\cref{fig:teaser}.

\begin{figure}[ht]
    \centering
    \includegraphics[width=0.7\linewidth]{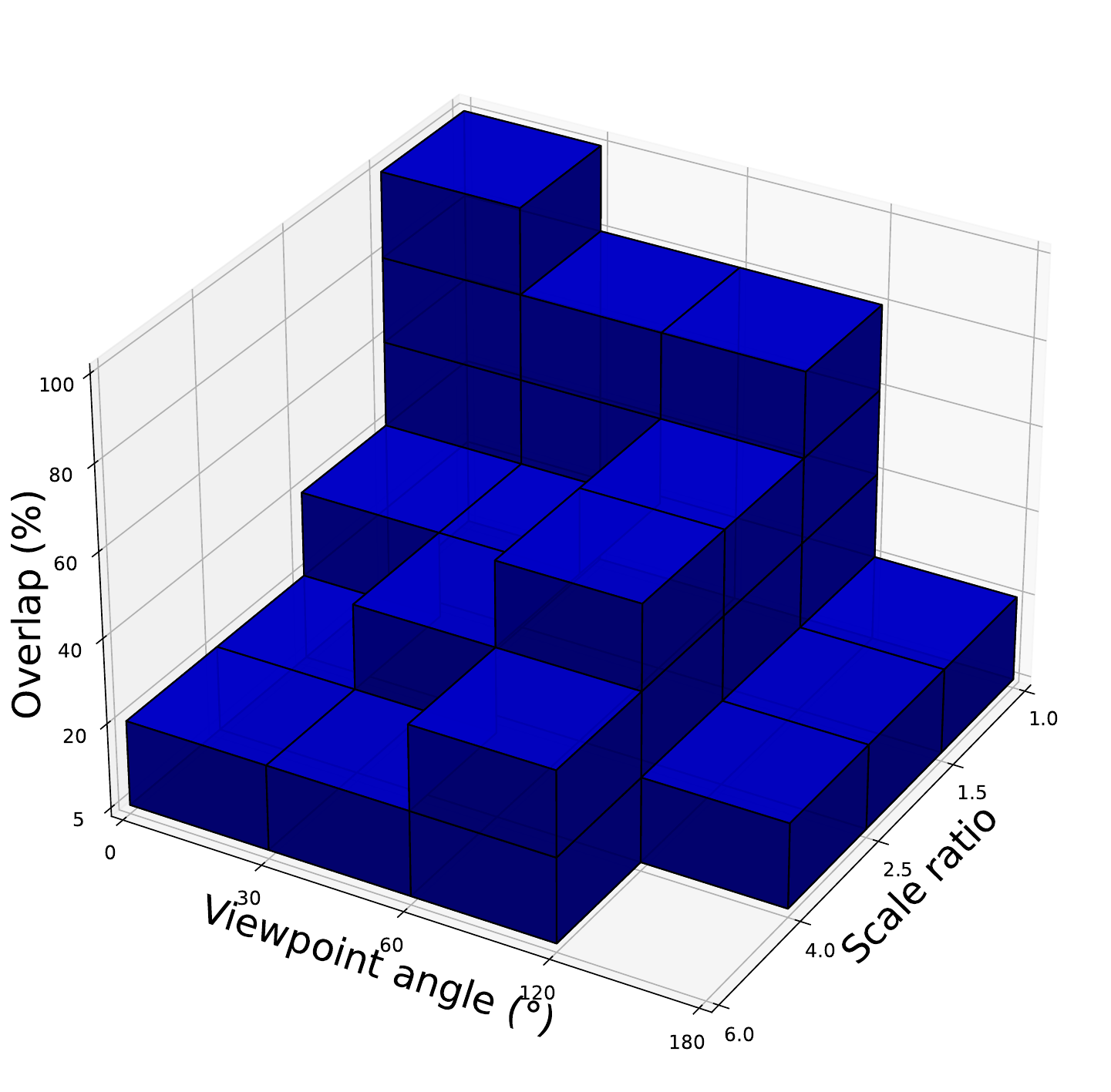}
    \caption{\textbf{Visualization of our 3D grid organization -- }Each axis represents one of our geometric criteria.}
    \label{fig:boxes_3d}
\end{figure}

While this binning strategy theoretically creates a $5\times4\times4$ grid (80 difficulty levels), not all combinations are physically possible. For instance, image pairs with both very large overlap and small scale ratio rarely exhibit large viewpoint angles, as these geometric conditions are inherently contradictory. We finally obtain 33 valid difficulty levels (see~\cref{fig:boxes_3d}).

To ensure statistical significance while maintaining a manageable dataset size, we populate each valid difficulty level with 500 randomly sampled image pairs. This results in a benchmark of 16.5K pairs, carefully curated to span the full spectrum of geometric challenges encountered in real-world scenarios. As shown in~\cref{fig:ablation_no_samples}, our choice of 500 image pairs per box ensures stable and reliable evaluation metrics, with similar conclusions holding across all evaluated methods.

This structured organization enables systematic evaluation of pose estimation methods across well-defined difficulty levels, from simple cases with large overlap and similar viewpoints to challenging scenarios with minimal overlap and extreme geometric variations.

%% file: sec/4_results.tex
\section{Results}
\label{sec:results}

We evaluate 14 image matching methods on our novel \mbox{\BN} benchmark to assess their performance across different geometric challenges. In this section, we first describe our evaluation protocol, then present a comprehensive analysis of both detector-based and detector-free approaches.

\subsection{Evaluation Protocol}
For a fair comparison of all the considered methods, for each image pair in our benchmark, we follow the same evaluation pipeline:

\begin{enumerate}
    \item \textbf{Image Matching -- }We first obtain matches between the two images using each method's default parameters (e.g. number of keypoints, backbone size, input image resolution etc.) and pre-trained weights.

    \item \textbf{Pose Estimation -- }Using these matches, we estimate the essential matrix using MAGSAC++~\cite{barath2020magsac++} from \textsc{OpenCV}, with a threshold of 0.5 pixel. From this essential matrix, we recover the relative rotation and the translation direction between the two views.

    \item \textbf{Scale Recovery -- }To obtain metric translations, we leverage depth predictions from UniDepth~\cite{piccinelli2024unidepth} at matched locations, following the approach in~\cite{leroy2024grounding}. This provides us with a metric scene scale, enabling full 6~DoF pose estimation.
\end{enumerate}

We consider a pose estimation successful when both the rotation error is less than 5° and the translation error is less than 2m. These thresholds were chosen based on typical requirements in real-world applications such as autonomous navigation and the precision of our ground truth poses, which were obtained through careful COLMAP reconstruction and alignment with nuScenes metric poses (see~\cref{sec:lifting}).

\begin{figure}[ht]
    \centering
    \includegraphics[width=\linewidth]{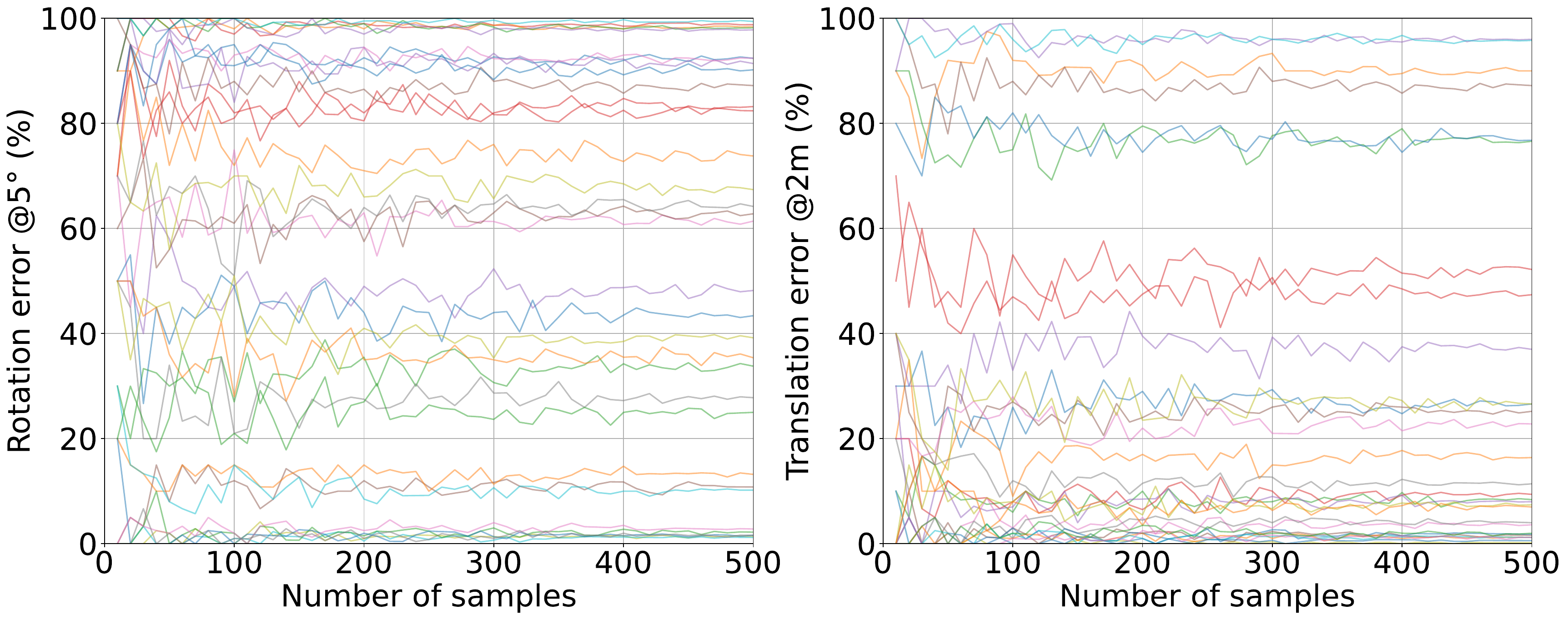}
    \caption{\textbf{Impact of sample size on evaluation reliability -- }
    We analyze how the number of image pairs per difficulty level affects the stability of performance metrics for LoFTR~\cite{sun2021loftr}. Left: Percentage of pairs with rotation error $<$ 5°. Right: Percentage of pairs with translation error $<$ 2m. The plots demonstrate that 500 pairs per difficulty level provide stable evaluation metrics, with minimal variance when increasing the sample size further than about 400 pairs. We obtain similar conclusions across all evaluated methods.
    }
    \label{fig:ablation_no_samples}
\end{figure}

\subsection{Learning-based Image Matching Methods}
Before presenting our benchmark results, we first provide a brief overview of recent advances in learning-based image matching methods. Recent years have seen significant advances in learning-based image matching approaches. These can be broadly categorized into detector-based and detector-free methods.

\paragraph{Detector-based methods} build upon traditional keypoint detection and description paradigms. SuperPoint~\cite{detone2018superpoint} pioneered self-supervised interest point detection and description. Recent works like DISK~\cite{tyszkiewicz2020disk}, ALIKED~\cite{zhao2023aliked}, and XFeat~\cite{potje2024xfeat} have further improved efficiency and accuracy. LightGlue~\cite{lindenberger2023lightglue} focuses on accelerating the matching process while maintaining high accuracy. DeDoDe v2~\cite{edstedt2024dedode} specifically addresses the challenges of keypoint detection reliability.

\paragraph{Detector-free methods} take a different approach by directly establishing dense correspondences between images. LoFTR~\cite{sun2021loftr} introduced transformer-based architectures for local feature matching without explicit keypoint detection. Recent works like RoMa~\cite{edstedt2024roma} and Efficient LoFTR~\cite{wang2024efficient} have improved both the efficiency and accuracy of dense matching approaches. DUSt3R~\cite{wang2024dust3r} introduces a paradigm shift by reformulating the matching problem as pointmap regression without requiring camera calibration or pose information, enabling joint optimization of 3D reconstruction and matching. Building upon this, MASt3R~\cite{leroy2024grounding} explicitly grounds the matching process in 3D space and introduces an efficient reciprocal matching scheme that significantly improves both speed and accuracy, particularly for challenging viewpoint changes. These 3D-aware approaches demonstrate substantial improvements over traditional 2D matching methods, especially in scenarios with extreme viewpoint variations.

\subsection{Main Results}
We evaluate the performance of 14 methods on our benchmark, including both detector-based approaches (SIFT~\cite{lowe2004distinctive}, SuperPoint~\cite{detone2018superpoint}, DISK~\cite{tyszkiewicz2020disk}, ALIKED~\cite{zhao2023aliked}, all using LightGlue~\cite{lindenberger2023lightglue}, XFeat~\cite{potje2024xfeat} and its variants, DeDoDe v2~\cite{edstedt2024dedode}) and detector-free approaches (LoFTR~\cite{sun2021loftr}, ELoFTR~\cite{wang2024efficient}, ASpanFormer~\cite{chen2022aspanformer}, RoMa~\cite{edstedt2024roma}, DUSt3R~\cite{wang2024dust3r} and MASt3R~\cite{leroy2024grounding}). For fair runtime comparison, all experiments were conducted on the same NVIDIA RTX 4090 GPU. ~\cref{tab:benchmark_results} presents the overall ranking of these methods, along with their computational efficiency.

\begin{table}[ht]
    \caption{\textbf{Benchmark Results -- }Average ranking across all difficulty levels (lower is better), based on the percentage of successful pose estimations (rotation error $<$ 5° and translation error $<$ 2m). We also report the overall percentage of successful pose estimations as well as the median runtime per image pair in milliseconds. Best and second-best values are shown in \textbf{bold} and \underline{underlined} respectively.}
    \label{tab:benchmark_results}
    \resizebox{!}{0.47\linewidth}{
        \begin{tabular}{lccc}
            \toprule
            Method & Avg. Rank & Success & Time \\
                   &           &  (\%)   & (ms) \\
            \midrule
            \multicolumn{4}{l}{\textit{Detector-based methods}} \\
            ALIKED+LightGlue~\cite{zhao2023aliked} & \textbf{5.3} & \textbf{36.8} & \underline{45} \\
            DISK+LightGlue~\cite{tyszkiewicz2020disk} & \underline{5.4} & \underline{35.9} & 69 \\
            SP+LightGlue~\cite{detone2018superpoint} & 6.1 & 35.7 & \textbf{43} \\
            SIFT+LightGlue~\cite{lowe2004distinctive} & 7.3 & 33.1 & 194 \\
            DeDoDe v2~\cite{edstedt2024dedode} & 8.6 & 30.4 & 282 \\
            XFeat~\cite{potje2024xfeat} & 13.1 & 14.2 & 54 \\
            XFeat*~\cite{potje2024xfeat} & 12.5 & 15.1 & 82 \\
            XFeat+LighterGlue~\cite{potje2024xfeat} & 9.0 & 30.1 & \textbf{43} \\
            \midrule
            \multicolumn{4}{l}{\textit{Detector-free methods}} \\
            LoFTR~\cite{sun2021loftr} & 10.8 & 24.9 & 185 \\
            ELoFTR~\cite{wang2024efficient} & 9.5 & 26.6 & \underline{124} \\
            ASpanFormer~\cite{chen2022aspanformer} & 9.8 & 24.8 & \textbf{108} \\
            RoMa~\cite{edstedt2024roma} & 2.7 & 47.3 & 614 \\
            DUSt3R~\cite{wang2024dust3r} & \textbf{2.4} & \textbf{54.8} & 257 \\
            MASt3R~\cite{leroy2024grounding} & \underline{2.5} & \underline{53.6} & 173 \\
            \bottomrule
        \end{tabular}
    }
\end{table}
Our benchmark, here aggregated, reveals several key findings:

\begin{enumerate}
    \item \textbf{Detector-free dominance -- }The top three performing methods (DUSt3R, MASt3R and RoMa) are all detector-free approaches, suggesting that direct dense matching is more robust across varying geometric conditions.

    \item \textbf{Speed-accuracy trade-off -- }While detector-free methods achieve better accuracy, they generally require more computation time. Detector-based methods, especially ALIKED or DISK, combined with LightGlue, offer competitive performance with significantly lower runtime (40-70ms vs. 150-600ms).

    \item \textbf{Impact of matching strategies -- }The quite significant performance gap between XFeat variants (with and without LighterGlue) highlights the importance of the matching strategy, even with the same feature detector.

    \item \textbf{Recent advances -- }The newest methods (DUSt3R, MASt3R, RoMa) show substantial improvements over their predecessors (LoFTR, ASpanFormer), demonstrating the rapid progress in the field.
\end{enumerate}

These results demonstrate that while recent detector-free methods achieve the best performance across our benchmark's diverse geometric challenges, detector-based approaches remain competitive, especially when computational efficiency is a priority.
\begin{figure*}[ht]
    \centering
    \begin{subfigure}{0.245\linewidth}
        \centering
        \includegraphics[width=\linewidth]{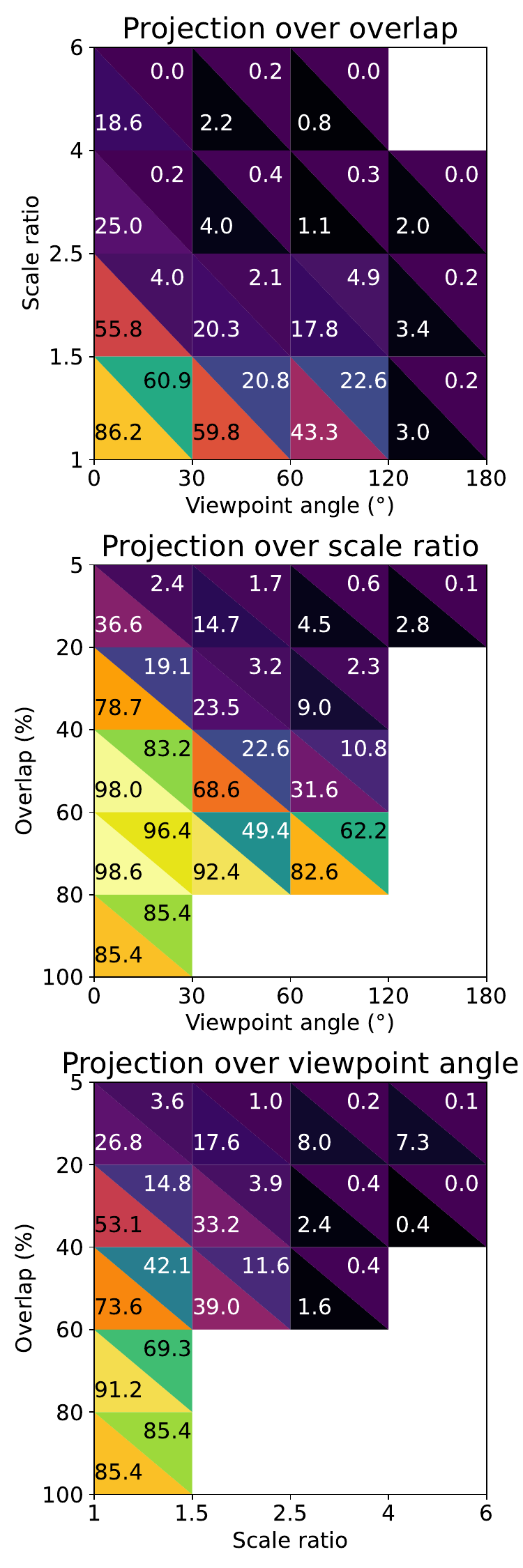}
        \caption{XFeat~\cite{potje2024xfeat}}
        \label{fig:xfeat}
    \end{subfigure}
    \begin{subfigure}{0.245\linewidth}
        \centering
        \includegraphics[width=\linewidth]{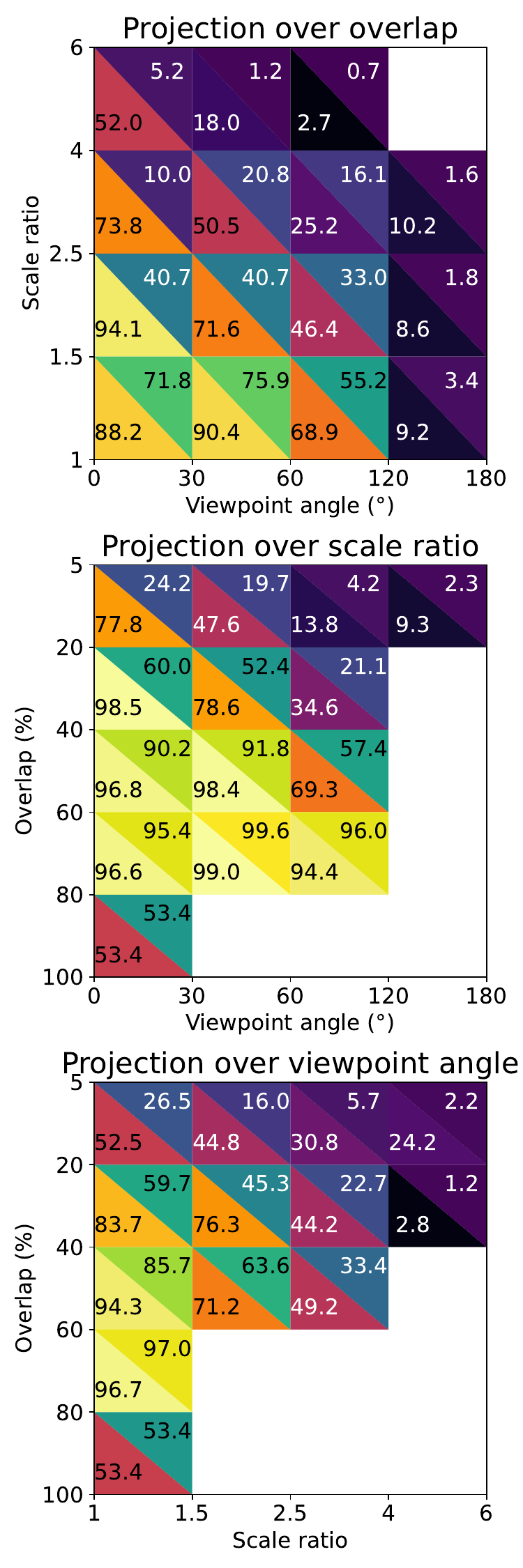}      
        \caption{ALIKED+LightGlue~\cite{zhao2023aliked}}
        \label{fig:aliked+lightglue}
    \end{subfigure}
    \begin{subfigure}{0.245\linewidth}
        \centering
        \includegraphics[width=\linewidth]{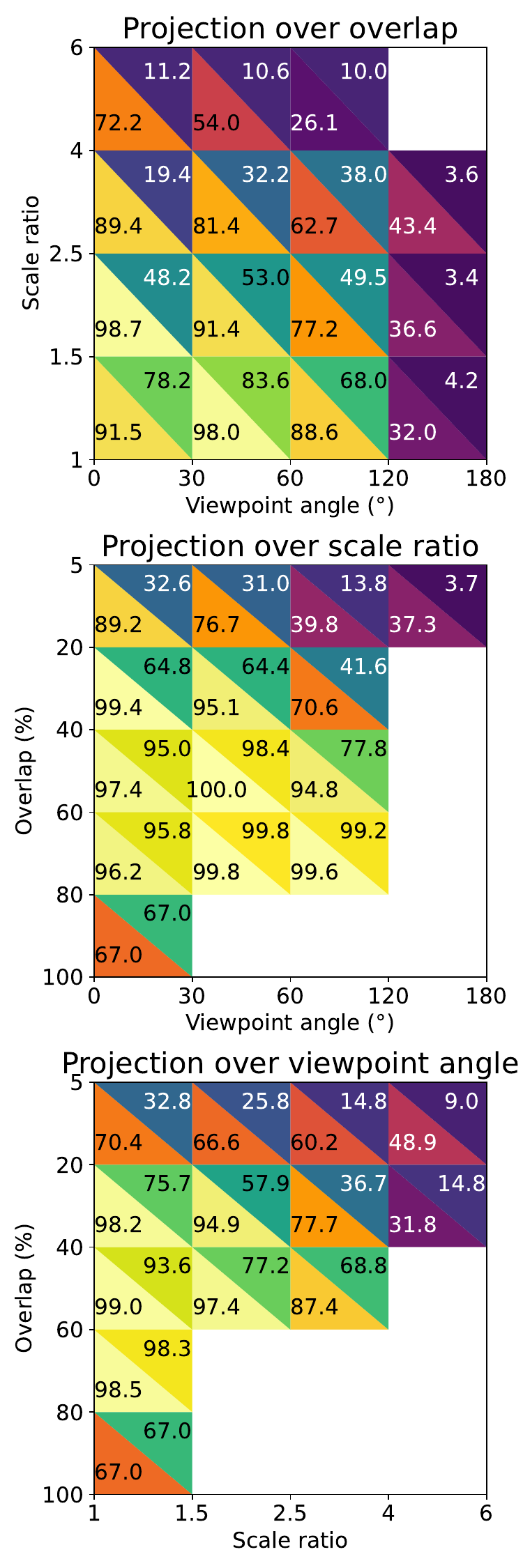}
        \caption{RoMa~\cite{edstedt2024roma}}
        \label{fig:roma}
    \end{subfigure}
    \begin{subfigure}{0.245\linewidth}
        \centering
        \includegraphics[width=\linewidth]{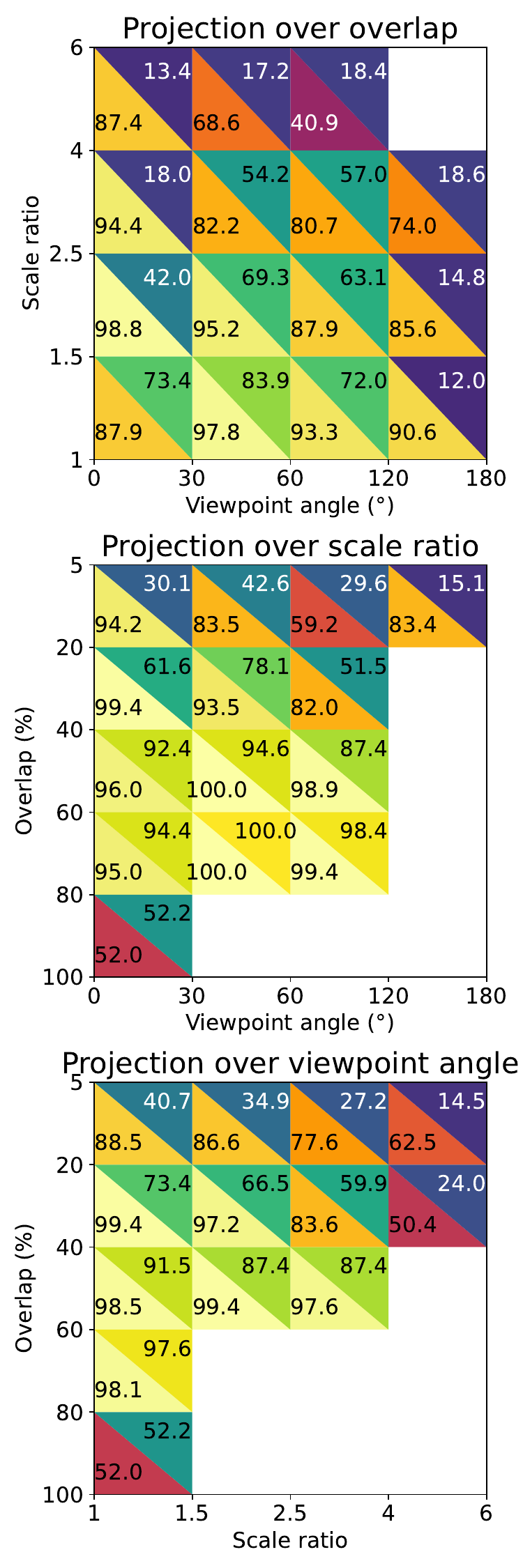}
        \caption{MASt3R~\cite{leroy2024grounding}}
        \label{fig:mast3r}
    \end{subfigure}
    \caption{\textbf{Performance analysis across geometric criteria -- }Success rate either for R\string@5° or t\string@2m (bottom-left and top-right of each triangle, respectively), for 4 methods (detector-based and detector-free), when projecting results onto individual geometric criteria (see~\cref{fig:boxes_3d}). For each method, we show three plots corresponding to the projection over overlap (top), scale ratio (middle), and viewpoint angle (bottom).}
    \label{fig:fine_grained}
\end{figure*}

In~\cref{fig:fine_grained}, we show for 4 methods how their performance varies with specific geometric challenges. Results for all other methods are in the Supplementary Material. Overall, detector-free methods show better performance than detector-based methods when dealing with high geometric challenges.
Specifically, DUSt3R and MASt3R show a better balance between speed and performance, allowing for stable results without steep degradation under tougher conditions, as is the case for most of the detector-based methods. Those fine grained findings emphasizes the recent advancements made by newer dense matching methods over traditional keypoint-based approaches. These improvements, especially in challenging scenarios, underline the value of incorporating visibility-aware features and dense matching strategies into camera pose estimation.
Our findings indicate that although most methods provide accurate pose estimates under conditions of high overlap, similar scale, and small relative viewpoint angles, even the top-performing method fails to correctly estimate the pose for over 45\% of image pairs when using thresholds of 5° (rotation) and 2m (translation). These results underscore the value of \BN in evaluating and comparing various approaches.

%% file: sec/5_conclusion.tex
\section{Limitations}
\label{sec:limitations}
Our co-visibility map generation pipeline lacks robustness to instance changes; for example, when one car is replaced by another, the depth check and normal check may still be satisfied, leading to the region being incorrectly considered co-visible. Some examples are in the Supplementary Material.

\section{Conclusion}
\label{sec:conclusion}
We introduced \BN, a novel benchmark that provides a systematic way to evaluate camera pose estimation methods across well-defined geometric challenges. By organizing 16.5K image pairs into 33 difficulty levels based on overlap, scale ratio, and viewpoint angle, our benchmark revealed several important insights. First, recent detector-free methods (DUSt3R, MASt3R, RoMa) significantly outperform traditional approaches, achieving success rates above 47\%, but at the cost of higher computational requirements (150-600ms vs. 40-70ms for detector-based methods). Second, even the best performing methods (DUSt3R and MASt3R) fail to correctly estimate poses more than 45\% of image pairs, highlighting significant room for improvement, particularly in challenging scenarios combining low overlap, large scale differences, and extreme viewpoint changes.
By providing a fine-grained understanding of method limitations, \BN opens new perspectives for developing more robust pose estimation approaches, particularly for challenging geometric configurations that current methods struggle with.

\section*{Acknowledgment}
This project has received funding from the Bosch Research Foundation
(Bosch Forschungsstiftung), and was granted access to the HPC resources of IDRIS under the allocation 2024-AD011014905R1 made by GENCI.

%% file: sec/X_suppl.tex
\clearpage
\setcounter{page}{1}

\newpage
\twocolumn[
\centering
\Large
\textbf{\includegraphics[height=2\fontcharht\font`f]{figures/rubik.png} \BN: A Structured Benchmark for Image Matching \\ across Geometric Challenges}\\
\vspace{0.5em}Supplementary Material \\
\vspace{1.0em}
] 

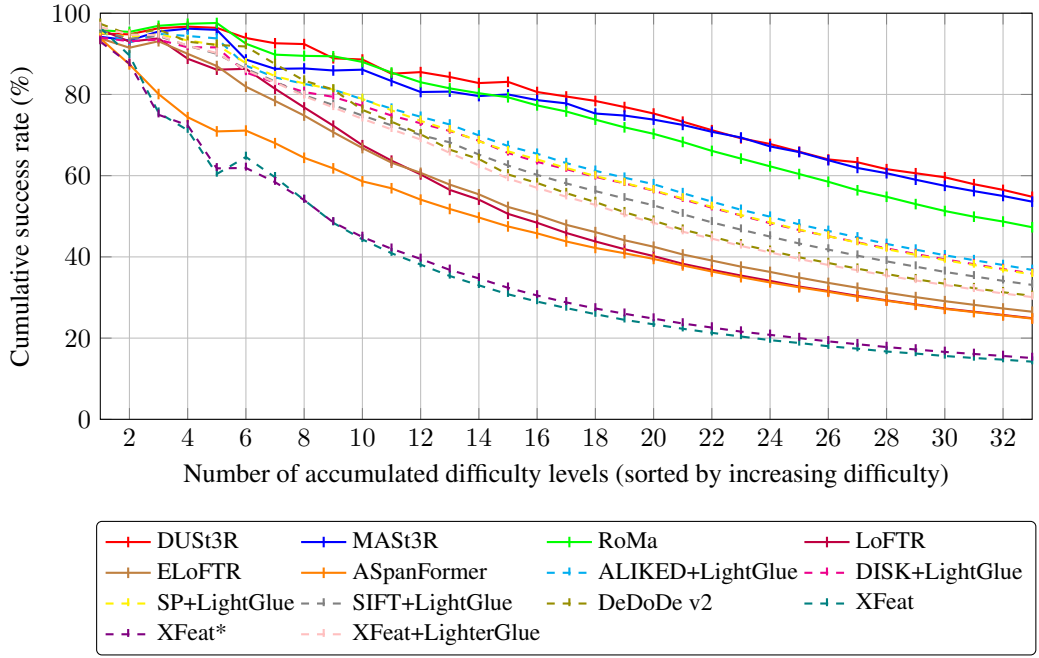
\begin{figure*}[ht]
    \centering
    \begin{tikzpicture}
    \begin{axis}[
        width=0.8\linewidth,
        height=0.4\linewidth,
        xlabel={Number of accumulated difficulty levels (sorted by increasing difficulty)},
        ylabel={Cumulative success rate (\%)},
        grid=major,
        legend style={
            at={(0.5,-0.25)},
            anchor=north,
            legend columns=4,
            font=\small,
            cells={anchor=west},
            draw=black,
            fill=white,
            rounded corners=2pt
        },
        xmin=1,
        xmax=33,
        ymin=0,
        ymax=100,
        ]
        
        \addplot[thick,color=red,mark=|] table[x expr=\coordindex+1,y index=0] {curves_acc.txt}; \addlegendentry{DUSt3R}
        \addplot[thick,color=blue,mark=|] table[x expr=\coordindex+1,y index=1] {curves_acc.txt}; \addlegendentry{MASt3R}
        \addplot[thick,color=green,mark=|] table[x expr=\coordindex+1,y index=2] {curves_acc.txt}; \addlegendentry{RoMa}
        \addplot[thick,color=purple,mark=|] table[x expr=\coordindex+1,y index=3] {curves_acc.txt}; \addlegendentry{LoFTR}
        \addplot[thick,color=brown,mark=|] table[x expr=\coordindex+1,y index=4] {curves_acc.txt}; \addlegendentry{ELoFTR}
        \addplot[thick,color=orange,mark=|] table[x expr=\coordindex+1,y index=5] {curves_acc.txt}; \addlegendentry{ASpanFormer}
        
        \addplot[thick,dashed,color=cyan,mark=|] table[x expr=\coordindex+1,y index=6] {curves_acc.txt}; \addlegendentry{ALIKED+LightGlue}
        \addplot[thick,dashed,color=magenta,mark=|] table[x expr=\coordindex+1,y index=7] {curves_acc.txt}; \addlegendentry{DISK+LightGlue}
        \addplot[thick,dashed,color=yellow,mark=|] table[x expr=\coordindex+1,y index=8] {curves_acc.txt}; \addlegendentry{SP+LightGlue}
        \addplot[thick,dashed,color=gray,mark=|] table[x expr=\coordindex+1,y index=9] {curves_acc.txt}; \addlegendentry{SIFT+LightGlue}
        \addplot[thick,dashed,color=olive,mark=|] table[x expr=\coordindex+1,y index=10] {curves_acc.txt}; \addlegendentry{DeDoDe v2}
        \addplot[thick,dashed,color=teal,mark=|] table[x expr=\coordindex+1,y index=11] {curves_acc.txt}; \addlegendentry{XFeat}
        \addplot[thick,dashed,color=violet,mark=|] table[x expr=\coordindex+1,y index=12] {curves_acc.txt}; \addlegendentry{XFeat*}
        \addplot[thick,dashed,color=pink,mark=|] table[x expr=\coordindex+1,y index=13] {curves_acc.txt}; \addlegendentry{XFeat+LighterGlue}
        
    \end{axis}
    \end{tikzpicture}
    \caption{\textbf{Cumulative success rates across difficulty levels -- }Methods are evaluated on increasingly difficult image pairs, sorted by the average success rate across all methods. Solid lines represent detector-free methods while dashed lines represent detector-based methods. The plot shows how performance degrades as more challenging pairs are included in the evaluation.}
    \label{fig:cumulative_results}
\end{figure*}

\begin{table*}[ht]
    \caption{\textbf{Detailed Results by Geometric Criterion -- }Success rate (in \%) for each method across individual geometric criterion bins. Best and second-best values for each column are shown in \textbf{bold} and \underline{underlined} respectively.}
    \label{tab:detailed_results}
    \resizebox{\textwidth}{!}{
        \begin{tabular}{l|ccccc|cccc|cccc|c}
            \toprule
            & \multicolumn{5}{c|}{Overlap (\%)} & \multicolumn{4}{c|}{Scale Ratio} & \multicolumn{4}{c|}{Viewpoint Angle (°)} & Whole \\
            & 80--100 & 60--80 & 40--60 & 20--40 & 5--20 & 1.0--1.5 & 1.5--2.5 & 2.5--4.0 & 4.0--6.0 & 0--30 & 30--60 & 60--120 & 120--180 & Dataset \\
            \midrule
            Number of boxes & 1 & 3 & 5 & 9 & 15 & 14 & 8 & 7 & 4 & 9 & 9 & 12 & 3 & 33 \\
            \midrule
            \multicolumn{14}{l}{\textit{Detector-based methods}} \\
            ALIKED+LightGlue~\cite{zhao2023aliked} & 53.4 & \textbf{95.8} & \textbf{68.2} & \underline{38.0} & \textbf{12.7} & \textbf{62.0} & \textbf{31.0} & \textbf{13.1} & 1.6 & \textbf{50.6} & \textbf{46.0} & \textbf{28.3} & \underline{2.0} & \textbf{36.8} \\
            DISK+LightGlue~\cite{tyszkiewicz2020disk} & 54.2 & 91.4 & 65.9 & \textbf{38.7} & \underline{11.8} & 60.4 & \underline{30.8} & 11.6 & \textbf{2.4} & \underline{50.3} & \underline{43.8} & 27.4 & \textbf{2.7} & \underline{35.9} \\
            SP+LightGlue~\cite{detone2018superpoint} & 64.8 & \underline{93.3} & \underline{68.0} & 36.4 & 10.9 & \underline{61.2} & 28.4 & \underline{12.5} & 1.4 & 49.9 & 43.0 & \underline{28.2} & 0.9 & 35.7 \\
            SIFT+LightGlue~\cite{lowe2004distinctive} & 68.2 & 92.1 & 61.4 & 32.3 & 9.9 & 57.3 & 26.9 & 9.6 & 1.7 & 49.8 & 39.7 & 23.7 & 0.5 & 33.1 \\
            DeDoDe v2~\cite{edstedt2024dedode} & \textbf{89.8} & \underline{93.3} & 54.8 & 26.7 & 7.9 & 60.4 & 16.3 & 3.2 & 0.9 & 49.3 & 35.4 & 19.9 & 0.3 &  30.4\\
            XFeat~\cite{potje2024xfeat} & \underline{85.4} & 67.4 & 24.3 & 5.2 & 0.9 & 32.1 & 2.4 & 0.1 & 0.0 & 34.4 & 8.3 & 7.1 & 0.0 & 14.2 \\
            XFeat*~\cite{potje2024xfeat} & 62.4 & 69.1 & 27.6 & 7.6 & 1.5 & 32.8 & 4.5 & 0.6 & 0.0 & 33.8 & 9.4 & 9.2 & 0.0 & 15.1 \\
            XFeat+LighterGlue~\cite{potje2024xfeat} & 64.6 & 91.7 & 59.1 & 26.2 & 8.1 & 56.6 & 20.9 & 4.6 & 0.2 & 48.0 & 33.4 & 21.4 & 1.2 & 30.1 \\
            \midrule
            \multicolumn{14}{l}{\textit{Detector-free methods}} \\
            LoFTR~\cite{sun2021loftr} & \textbf{87.2} & 88.4 & 47.2 & 17.5 & 5.0 & 51.6 & 10.1 & 2.3 & 0.6 & 43.2 & 27.9 & 15.1 & 0.0 & 24.9 \\
            ELoFTR~\cite{wang2024efficient} & 56.4 & 90.3 & 50.8 & 22.1 & 6.3 & 51.2 & 15.6 & 4.4 & 0.7 & 42.2 & 30.8 & 18.2 & 0.1 & 26.6 \\
            ASpanFormer~\cite{chen2022aspanformer} & 72.2 & 72.3 & 44.5 & 21.9 & 7.4 & 46.0 & 14.9 & 6.9 & 1.6 & 42.5 & 27.2 & 16.0 & 0.1 & 24.8 \\
            RoMa~\cite{edstedt2024roma} & 67.0 & \textbf{98.3} & 84.5 & 52.7 & 20.2 & \underline{71.2} & 43.2 & 26.6 & 8.3 & \underline{57.5} & \underline{56.2} & 44.1 & 3.0 & 47.3 \\
            DUSt3R~\cite{wang2024dust3r} & \underline{81.8} & 97.4 & \textbf{90.8} & \underline{58.4} & \textbf{30.4} & \textbf{73.3} & \textbf{57.9} & \underline{40.1} & \underline{9.9} & \textbf{67.4} & 55.3 & \underline{50.0} & \textbf{35.2} & \textbf{54.8} \\
            MASt3R~\cite{leroy2024grounding} & 52.0 & \underline{97.5} & \underline{89.6} & \textbf{61.0} & \underline{28.4} & \underline{71.2} & \underline{52.3} & \textbf{42.5} & \textbf{13.8} & 53.5 & \textbf{65.6} & \textbf{54.5} & \underline{14.1} & \underline{53.6} \\
            \bottomrule
        \end{tabular}
    }
\end{table*}

\section{Additional Results}

We provide detailed performance metrics for all evaluated methods across our benchmark's geometric criteria. In~\cref{tab:detailed_results}, we break down the success rates according to individual geometric bins showing the percentage of successful pose estimations for each method across the different ranges of overlap, scale ratio, and viewpoint angle. This granular analysis complements the aggregated results presented in the main paper (see~\cref{tab:benchmark_results}).

\begin{figure*}[ht]
    \centering
    \begin{subfigure}{0.195\linewidth}
        \centering
        \includegraphics[width=\linewidth]{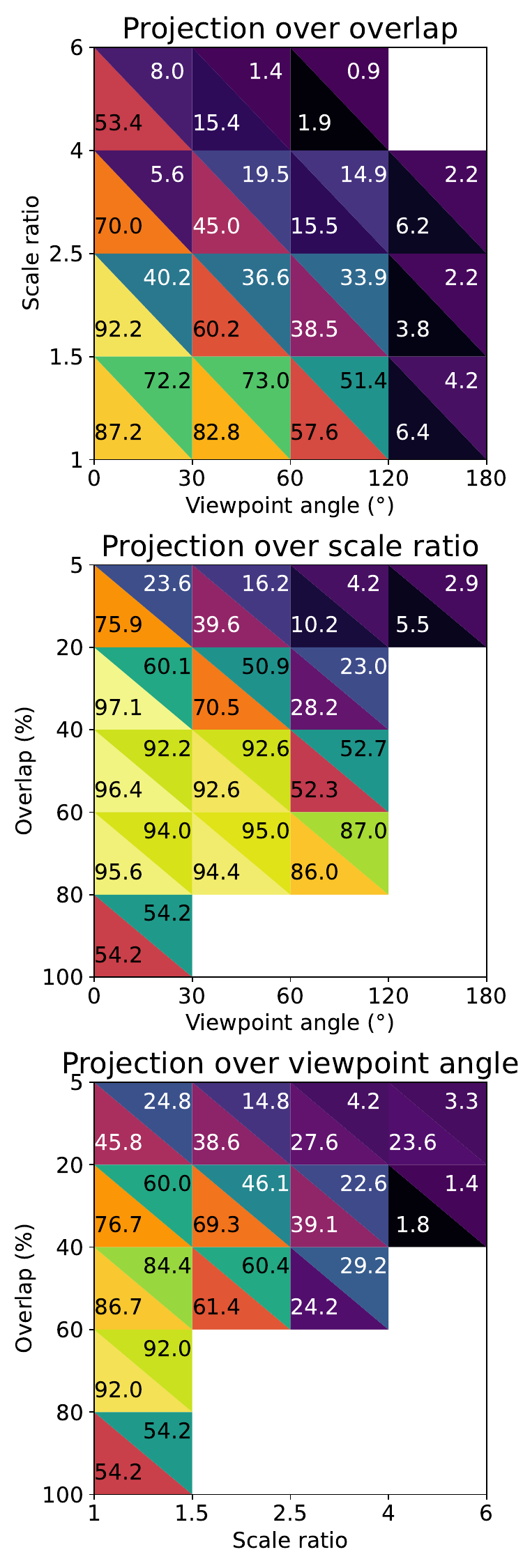}
        \caption{DISK+LightGlue~\cite{tyszkiewicz2020disk}}
        \label{fig:disk+lightglue}
    \end{subfigure}
    \begin{subfigure}{0.195\linewidth}
        \centering
        \includegraphics[width=\linewidth]{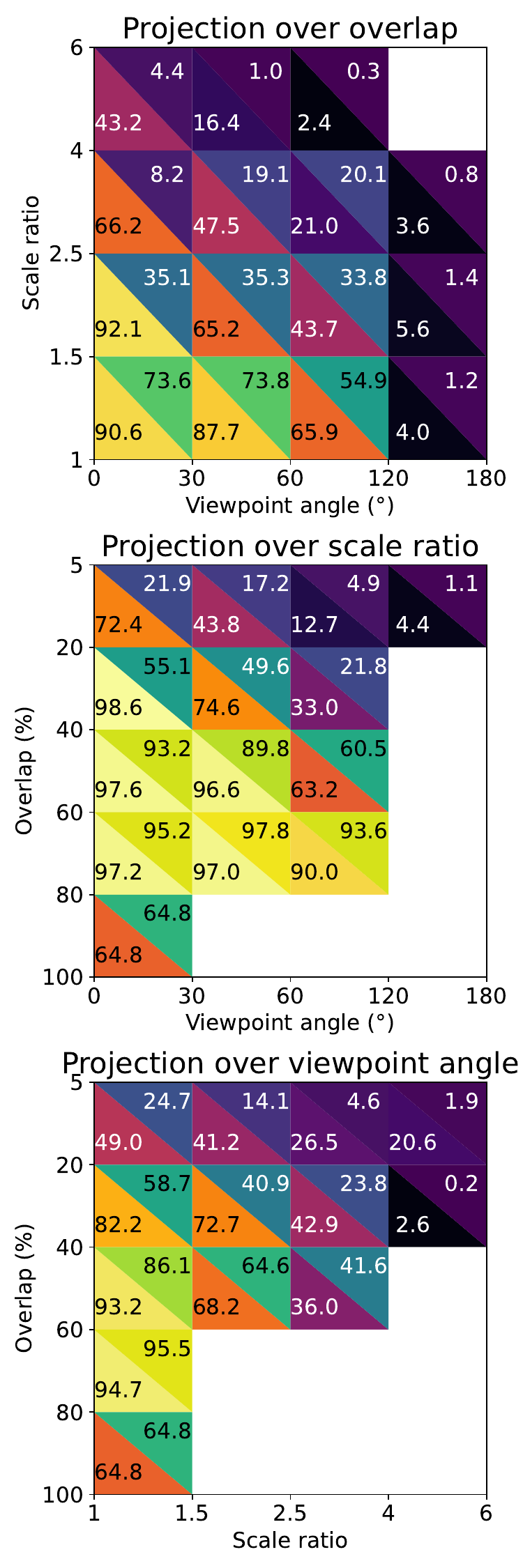}      
        \caption{SP+LightGlue~\cite{detone2018superpoint}}
        \label{fig:sp+lightglue}
    \end{subfigure}
    \begin{subfigure}{0.195\linewidth}
        \centering
        \includegraphics[width=\linewidth]{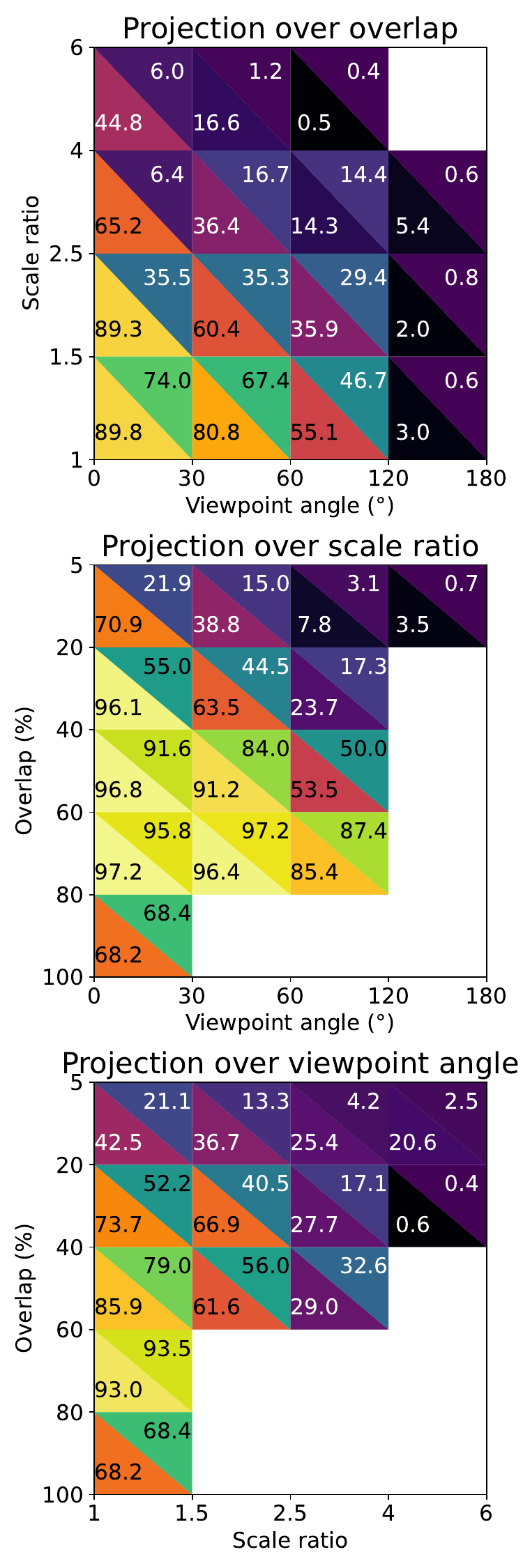}
        \caption{SIFT+LightGlue~\cite{lowe2004distinctive}}
        \label{fig:sift+lightglue}
    \end{subfigure}
    \begin{subfigure}{0.195\linewidth}
        \centering
        \includegraphics[width=\linewidth]{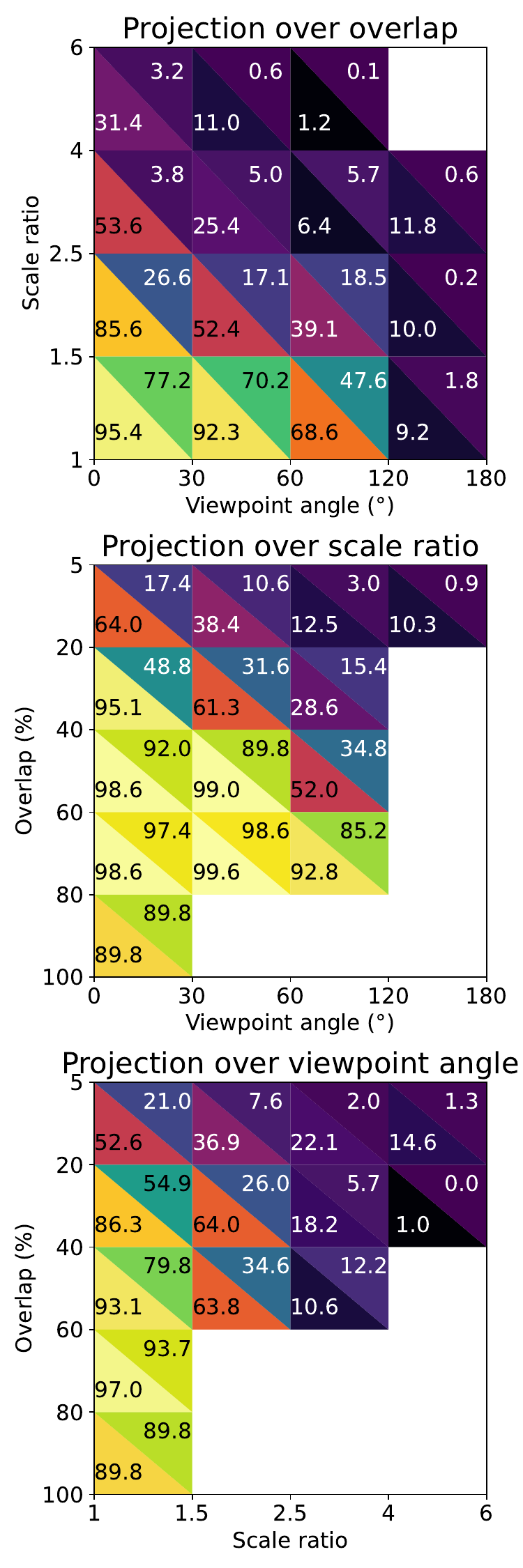}
        \caption{DeDode v2~\cite{edstedt2024dedode}}
        \label{fig:dedode}
    \end{subfigure}
    \begin{subfigure}{0.195\linewidth}
        \centering
        \includegraphics[width=\linewidth]{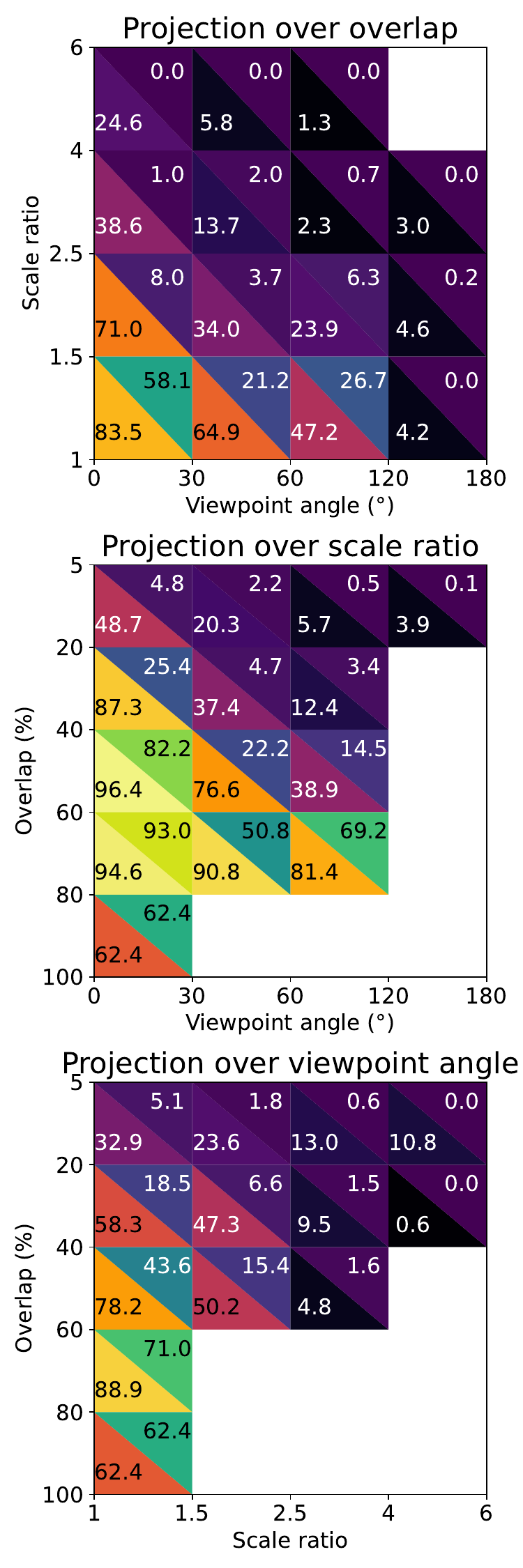}
        \caption{XFeat*~\cite{potje2024xfeat}}
        \label{fig:xfeat_star}
    \end{subfigure}
    \begin{subfigure}{0.195\linewidth}
        \centering
        \includegraphics[width=\linewidth]{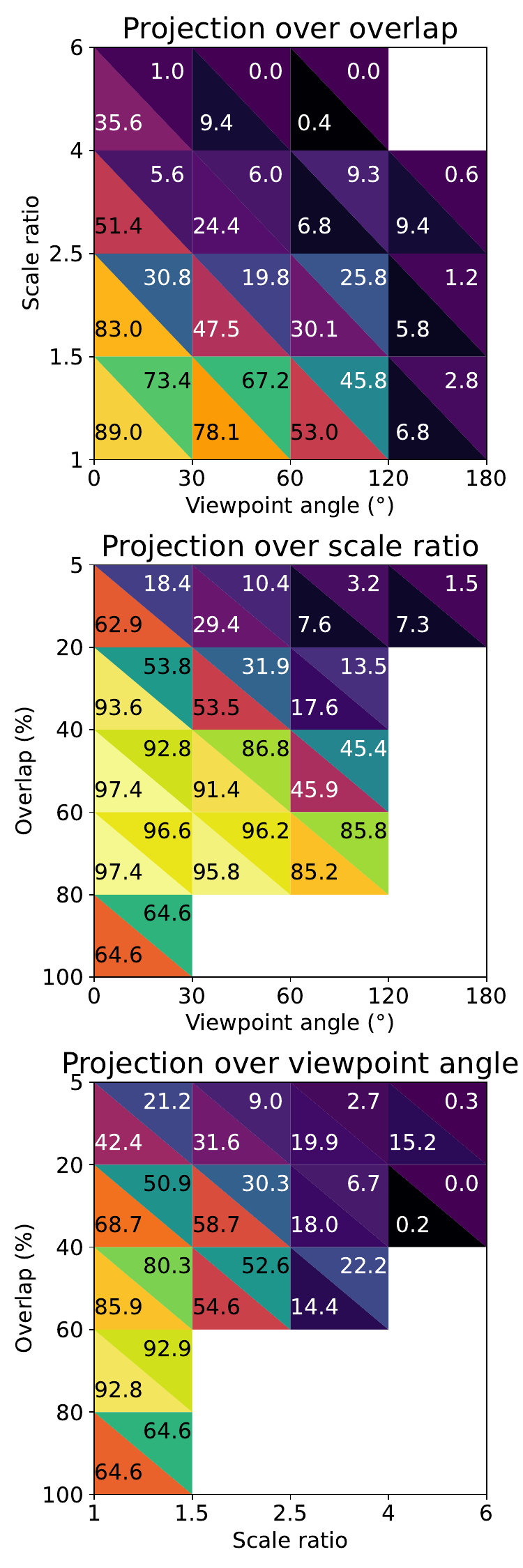}
        \caption{XFeat+LighterGlue~\cite{potje2024xfeat}}
        \label{fig:xfeat+lighterglue}
    \end{subfigure}
    \begin{subfigure}{0.195\linewidth}
        \centering
        \includegraphics[width=\linewidth]{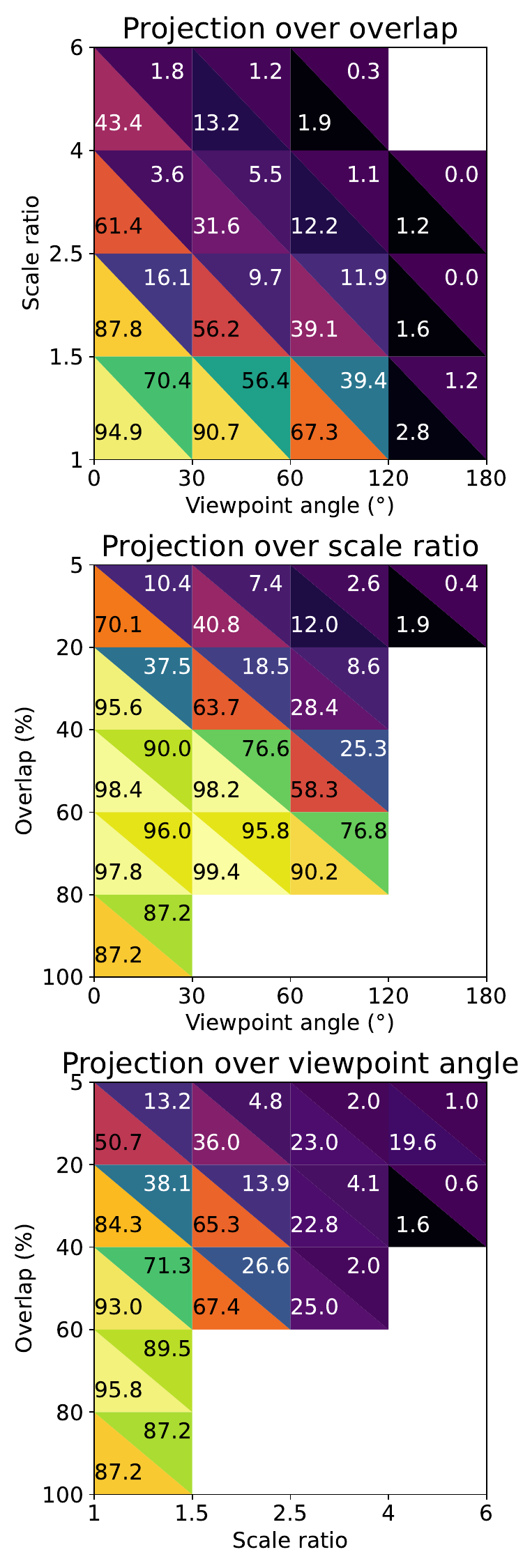}
        \caption{LoFTR~\cite{sun2021loftr}}
        \label{fig:loftr}
    \end{subfigure}
    \begin{subfigure}{0.195\linewidth}
        \centering
        \includegraphics[width=\linewidth]{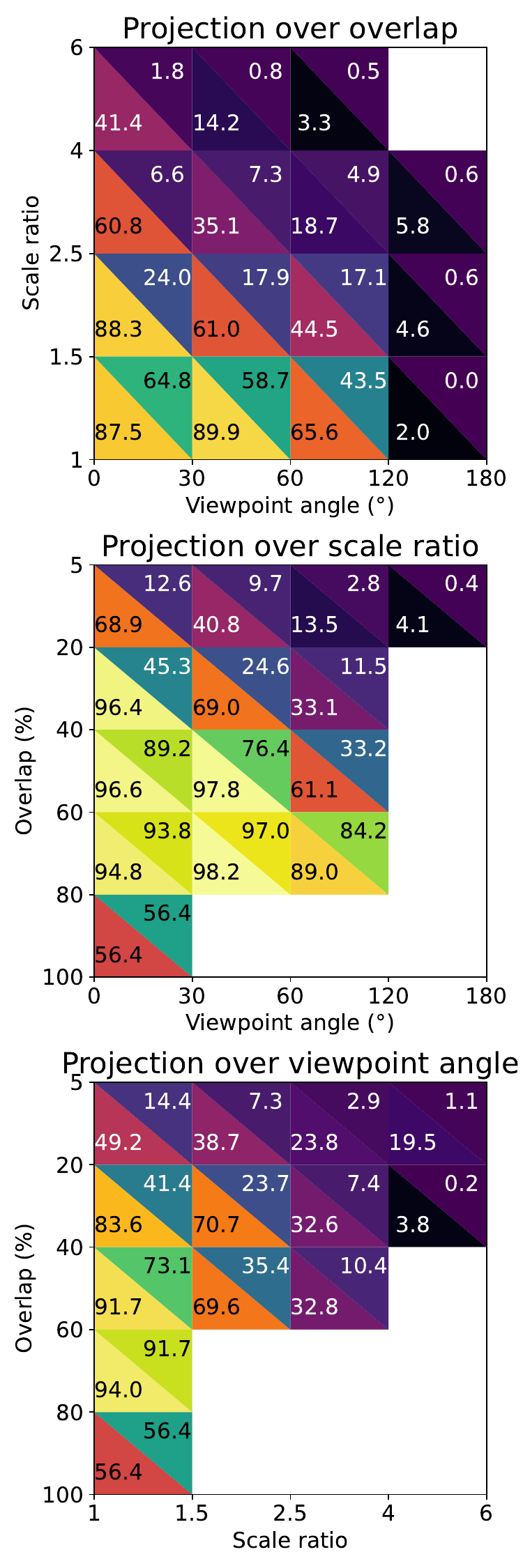}
        \caption{ELoFTR~\cite{wang2024efficient}}
        \label{fig:eloftr}
    \end{subfigure}
    \begin{subfigure}{0.195\linewidth}
        \centering
        \includegraphics[width=\linewidth]{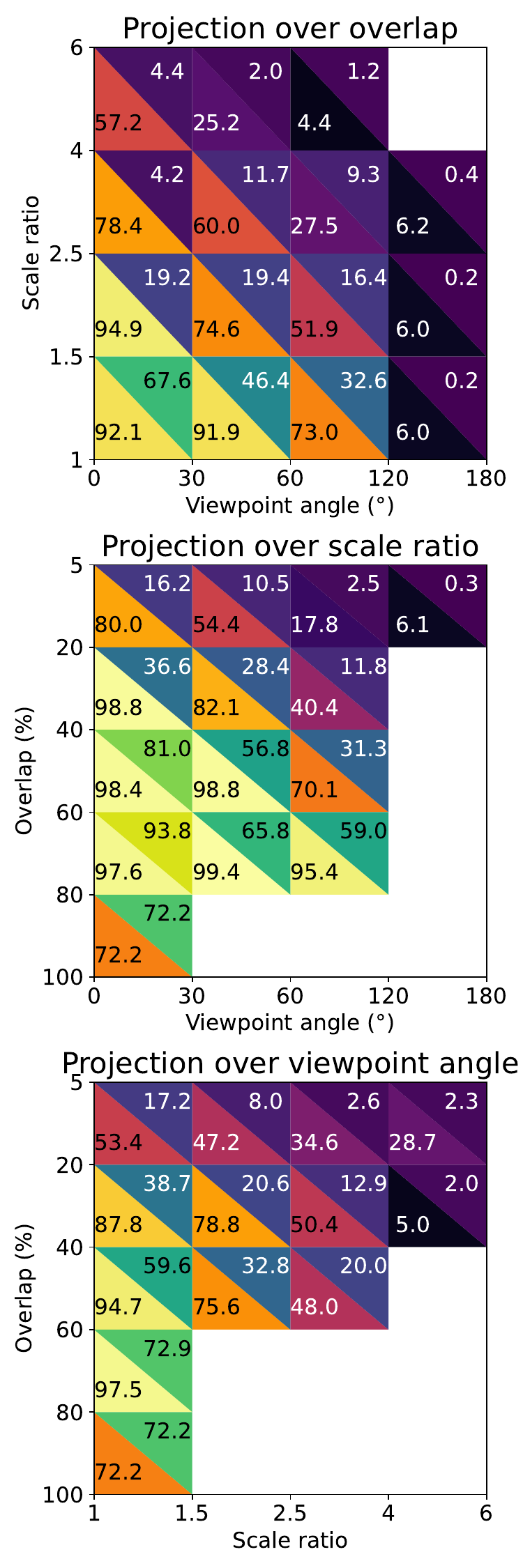}
        \caption{ASpanFormer~\cite{chen2022aspanformer}}
        \label{fig:aspanformer}
    \end{subfigure}
    \begin{subfigure}{0.195\linewidth}
        \centering
        \includegraphics[width=\linewidth]{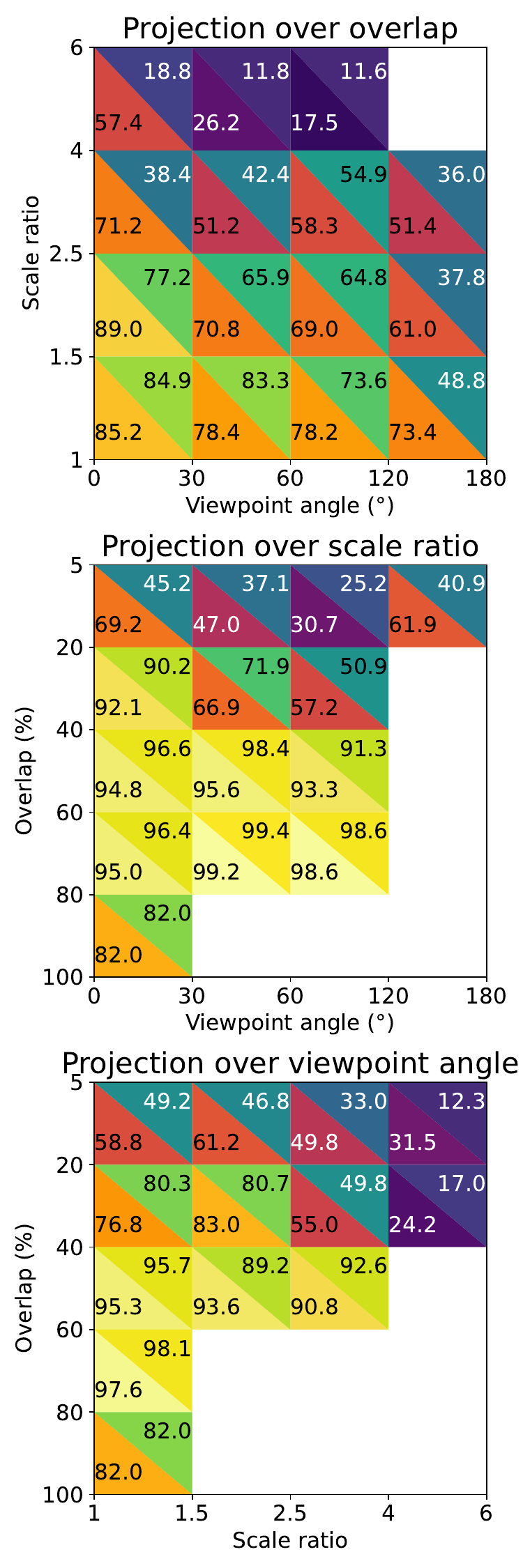}
        \caption{DUSt3R~\cite{wang2024dust3r}}
        \label{fig:dust3r}
    \end{subfigure}
    \caption{\textbf{Performance analysis across geometric criteria -- }Results for other methods not in the main paper, similar than~\cref{fig:fine_grained}.}
    \label{fig:fine_grained_suppl}
\end{figure*}

The performance analysis across geometric criteria for methods not shown in~\cref{fig:fine_grained} is presented in~\cref{fig:fine_grained_suppl}. These triangular plots follow the same visualization approach as in the main paper, with success rates for rotation (bottom-left) and translation (top-right) thresholds projected onto individual geometric criterion: overlap (top), scale ratio (middle), and viewpoint angle (bottom).

To provide additional context for the cumulative results analysis, we present in~\cref{tab:difficulty_ordering} the complete ordering of all 33 difficulty levels, sorted by decreasing average success rate across all methods. This ordering reveals clear patterns in what makes image pairs challenging: the easiest pairs typically combine high overlap (60-80\%), small scale changes (1.0-1.5), and small viewpoint changes (0-30°), while the most challenging pairs involve minimal overlap (5-20\%), large scale changes (4.0-6.0), and significant viewpoint changes (60-120°). This ordering was used to generate the cumulative plot in~\cref{fig:cumulative_results}, which shows how performance evolves when starting from the easiest geometric configurations (1 box) and gradually incorporating more difficult image pairs up to the complete benchmark (33 boxes). This visualization complements the fine-grained analysis by showing the overall robustness of each method across the full spectrum of geometric challenges.

These additional results further support and refine the conclusions drawn in the main paper. The detailed breakdown in~\cref{tab:detailed_results} reveal several noteworthy patterns:

\begin{enumerate}
    \item \textbf{Extreme conditions handling -- }While the best detector-free methods generally outperform the best detector-based ones, this gap becomes particularly pronounced in extreme geometric conditions. For instance, at very low overlap (5-20\%), DUSt3R and MASt3R maintain success rates of 30.4\% and 28.4\% respectively, while the best detector-based method (ALIKED+LightGlue) achieves only 12.7\%.

    \item \textbf{Detector-based methods vs LoFTR-like detector-free methods -- } LoFTR-like methods (LoFTR, ELoFTR and ASpanFormer) are almost systematically outperformed by several detector-based methods (DeDoDe v2, XFeat+LighterGlue, ALIKED+LighGLue, DISK+LightGlue, SP+LightGlue, SIFT+LightGlue).
    
    \item \textbf{Performance degradation patterns -- }The cumulative plot in~\cref{fig:cumulative_results} reveals distinct patterns in how different methods handle increasing geometric difficulty. Detector-free methods, particularly DUSt3R and MASt3R, show a more gradual performance degradation compared to detector-based approaches. This is quantitatively confirmed in~\cref{tab:detailed_results}, where these methods maintain relatively high success rates across all geometric criteria: overlap ($>$28\% even at 5-20\%), scale ratio ($>$40\% up to 4.0), and viewpoint angle ($>$50\% up to 120°). In contrast, detector-based methods show steeper performance drops, especially in challenging conditions, suggesting that recent dense matching approaches are inherently more robust to various geometric transformations (as some of the older detector-free approaches are beaten by most of the detector-based ones).

    \item \textbf{High overlap performance paradox -- }Interestingly, almost all methods perform better on image pairs with 60-80\% overlap compared to those with 80-100\% overlap. This seemingly counter-intuitive behavior could be explained by the geometric configuration of these pairs. Very high overlap ($>$80\%) often occurs in image pairs taken from nearly identical positions, resulting in very small baselines (i.e. small distance between camera centers). While these pairs have strong visual similarity, the small baseline makes both rotation and translation estimation challenging: small errors in matching lead to large uncertainties in triangulation geometry, affecting both the essential matrix estimation and the subsequent pose decomposition. In contrast, pairs with 60-80\% overlap typically have larger baselines while maintaining sufficient visual correspondences, creating more favorable conditions for pose estimation.
\end{enumerate}

These findings highlight the importance of comprehensive evaluation across different geometric criteria, as methods can exhibit significantly different behaviors depending on the specific challenges they encounter.

\begin{table*}[ht]
    \caption{\textbf{Difficulty Level Ordering -- }All 33 difficulty levels sorted by decreasing average success rate across all methods. Each level is defined by its overlap range (\%), scale ratio range, and viewpoint angle range (°).}
    \label{tab:difficulty_ordering}
    \centering
    \begin{tabular}{ccccc}
        \toprule
        Level & Overlap (\%) & Scale Ratio & Viewpoint (°) & Success (\%) \\
        \midrule
        1 & 60--80 & 1.0--1.5 & 0--30 & 95.2 \\
        2 & 40--60 & 1.0--1.5 & 0--30 & 89.9 \\
        3 & 60--80 & 1.0--1.5 & 30--60 & 88.0 \\
        4 & 60--80 & 1.0--1.5 & 60--120 & 82.2 \\
        5 & 40--60 & 1.0--1.5 & 30--60 & 75.5 \\
        6 & 80--100 & 1.0--1.5 & 0--30 & 68.5 \\
        7 & 20--40 & 1.0--1.5 & 0--30 & 60.6 \\
        8 & 40--60 & 1.0--1.5 & 60--120 & 57.9 \\
        9 & 20--40 & 1.0--1.5 & 30--60 & 52.7 \\
        10 & 40--60 & 1.5--2.5 & 60--120 & 47.1 \\
        11 & 5--20 & 1.0--1.5 & 0--30 & 40.6 \\
        12 & 20--40 & 1.5--2.5 & 0--30 & 40.4 \\
        13 & 20--40 & 1.5--2.5 & 30--60 & 36.7 \\
        14 & 20--40 & 1.0--1.5 & 60--120 & 33.0 \\
        15 & 40--60 & 2.5--4.0 & 60--120 & 28.3 \\
        16 & 5--20 & 1.0--1.5 & 30--60 & 27.6 \\
        17 & 20--40 & 1.5--2.5 & 60--120 & 25.3 \\
        18 & 5--20 & 1.5--2.5 & 0--30 & 22.5 \\
        19 & 20--40 & 2.5--4.0 & 30--60 & 22.2 \\
        20 & 5--20 & 1.5--2.5 & 30--60 & 20.5 \\
        21 & 20--40 & 2.5--4.0 & 60--120 & 12.2 \\
        22 & 5--20 & 1.0--1.5 & 60--120 & 10.6 \\
        23 & 5--20 & 2.5--4.0 & 30--60 & 9.3 \\
        24 & 5--20 & 2.5--4.0 & 0--30 & 9.0 \\
        25 & 5--20 & 1.5--2.5 & 60--120 & 6.4 \\
        26 & 5--20 & 4.0--6.0 & 0--30 & 5.4 \\
        27 & 5--20 & 1.0--1.5 & 120--180 & 5.0 \\
        28 & 5--20 & 2.5--4.0 & 60--120 & 4.1 \\
        29 & 5--20 & 1.5--2.5 & 120--180 & 4.1 \\
        30 & 5--20 & 2.5--4.0 & 120--180 & 3.8 \\
        31 & 5--20 & 4.0--6.0 & 30--60 & 3.0 \\
        32 & 20--40 & 4.0--6.0 & 60--120 & 2.9 \\
        33 & 5--20 & 4.0--6.0 & 60--120 & 1.0 \\
        \bottomrule
    \end{tabular}
\end{table*}

\section{Limitations}
\begin{figure*}
    \centering
    \begin{subfigure}{0.48\linewidth}
        \fbox{\includegraphics[width=\linewidth]{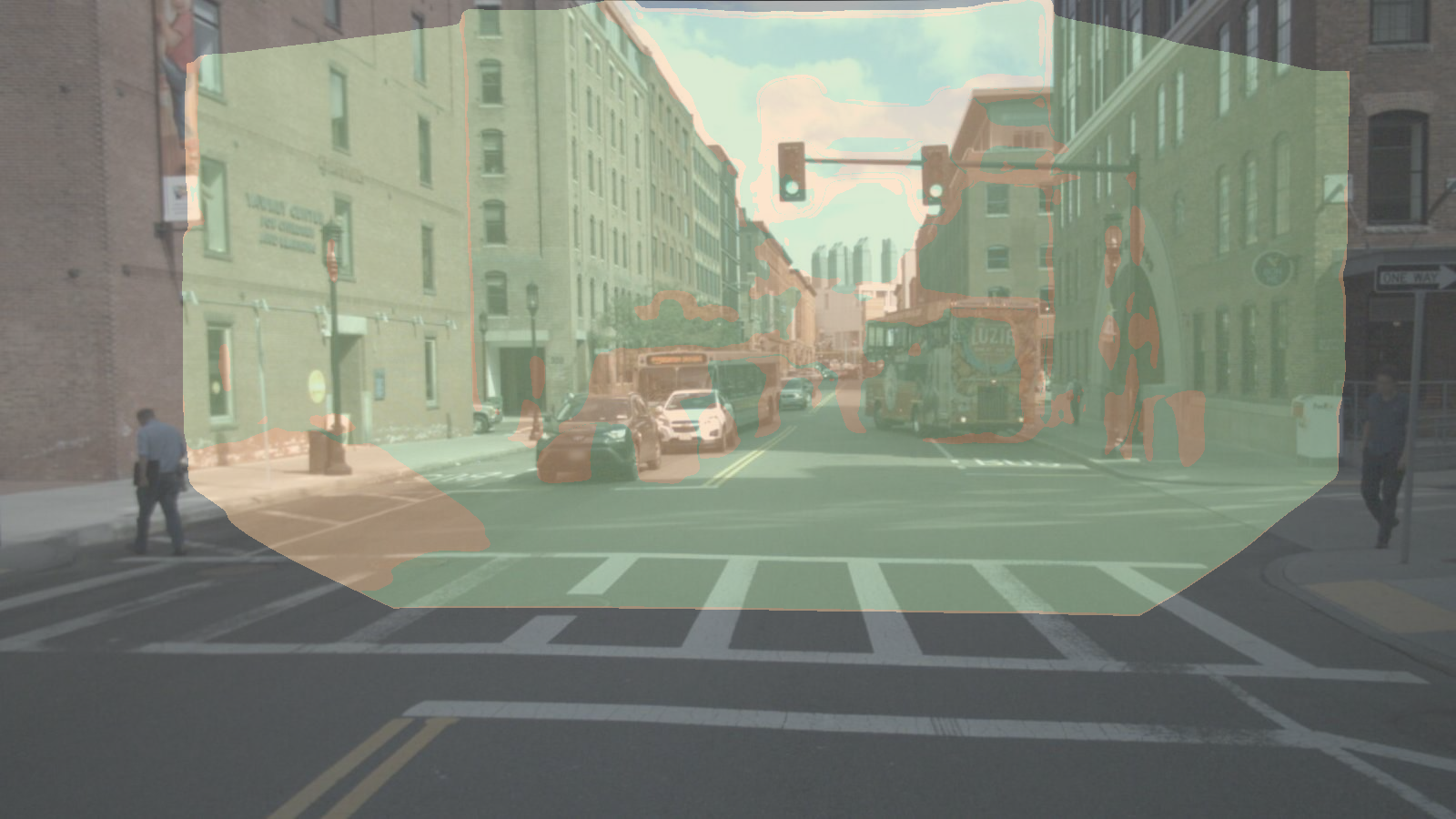}}
    \end{subfigure}
    \begin{subfigure}{0.48\linewidth}
        \fbox{\includegraphics[width=\linewidth]{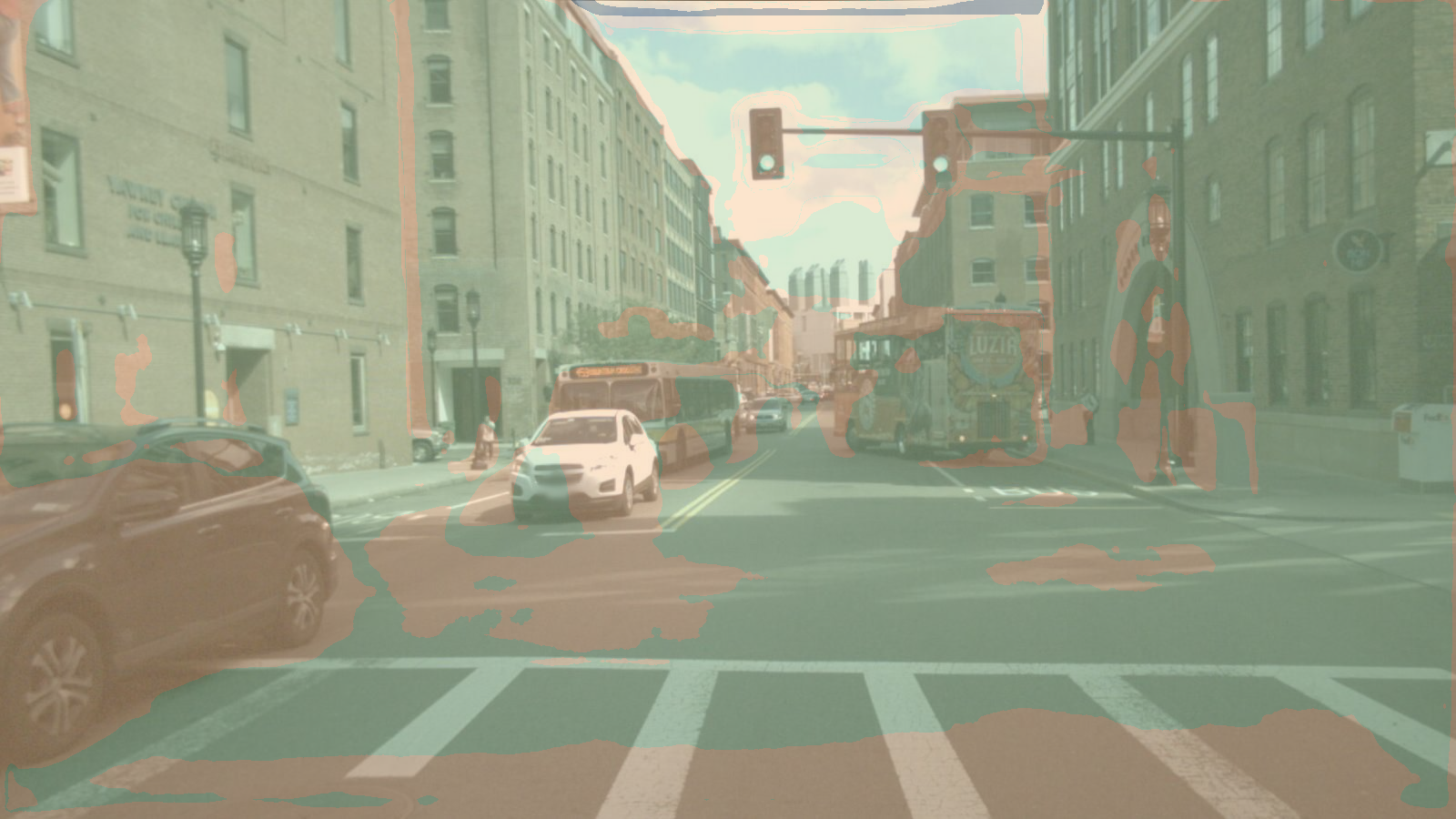}}
    \end{subfigure} \makebox[\linewidth]{\rule{\linewidth}{1pt}}
    \begin{subfigure}{0.48\linewidth}
        \fbox{\includegraphics[width=\linewidth]{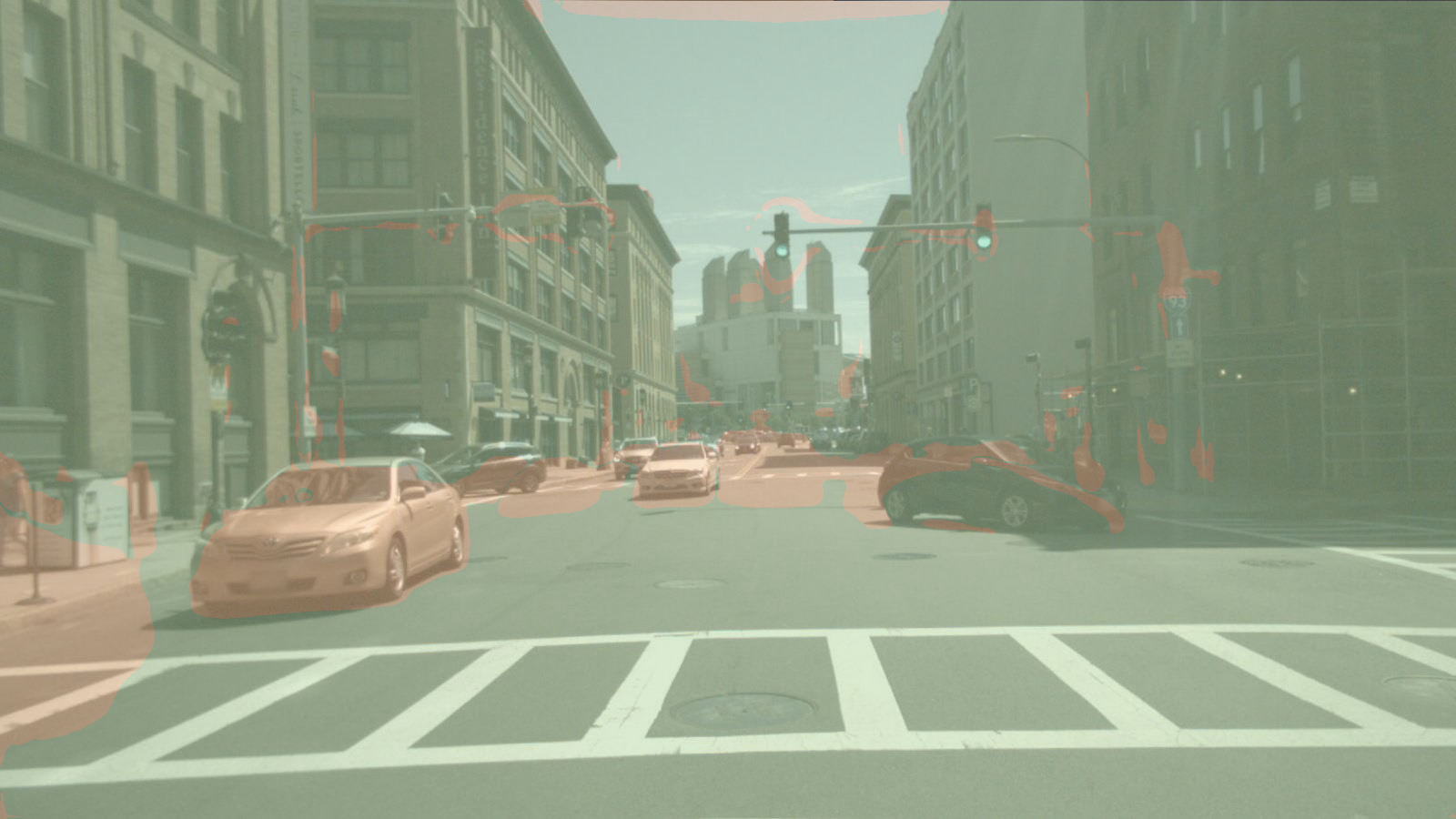}}
        \caption{First view}
    \end{subfigure}
    \begin{subfigure}{0.48\linewidth}
        \fbox{\includegraphics[width=\linewidth]{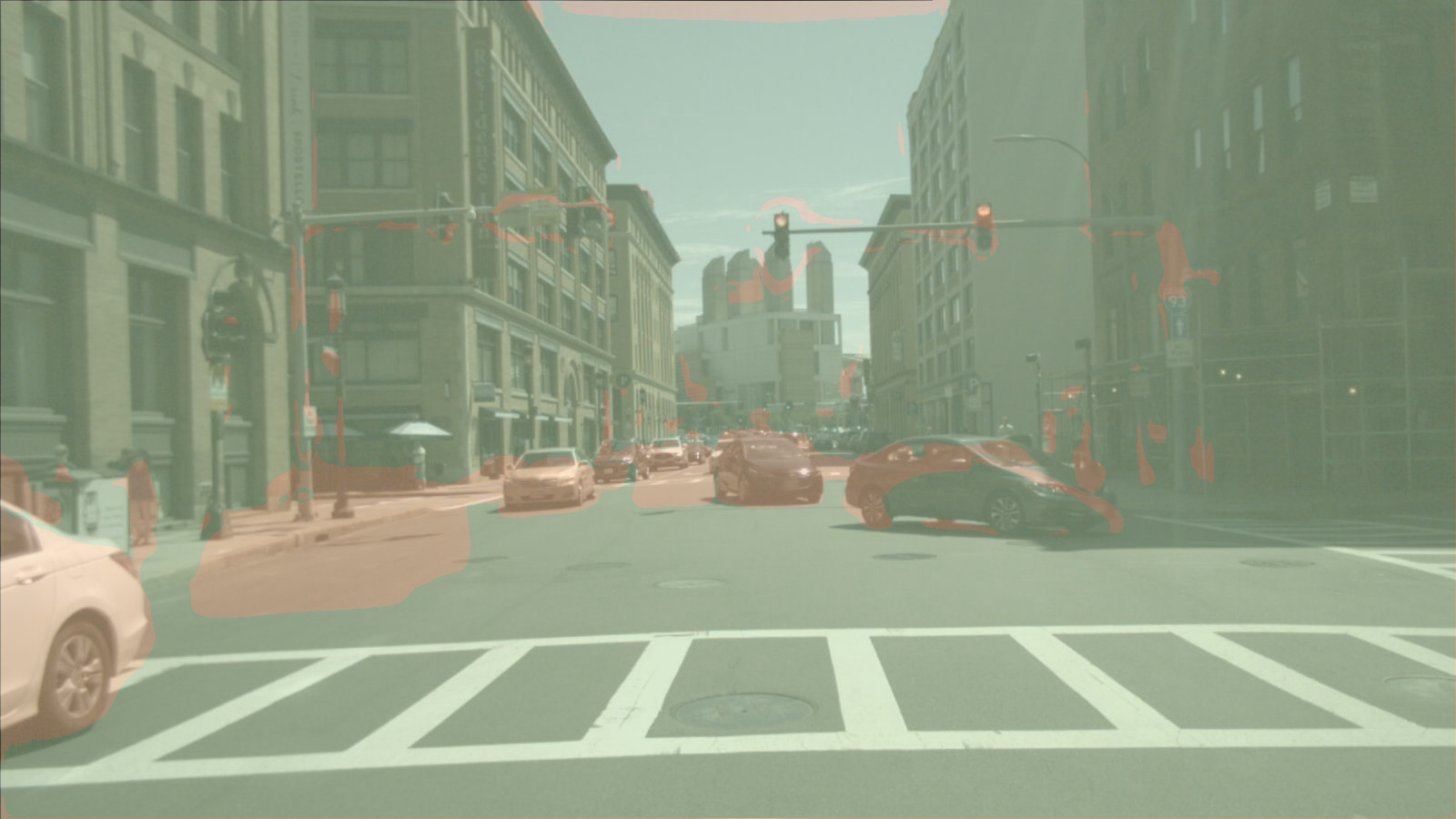}}
        \caption{Second view}
    \end{subfigure}

    \caption{\textbf{Limitations in co-visibility estimation -- }Our method for determining co-visible regions can be affected by dynamic objects in the scene. In these examples, different cars occupy the same space in two temporally separated views. On the top pair, the white car replaces the gray car, and part of both cars are marked as co-visible. On the bottom pair, the cars turning in both views are different, but marked as co-visible as well. This highlights a limitation in handling dynamic scene elements when computing co-visibility maps.}
    \label{fig:limitations}
\end{figure*}

While our benchmark provides comprehensive evaluations across various geometric challenges, there are some inherent limitations in how we determine co-visibility between image pairs. The main challenge stems from dynamic objects in the scenes, as illustrated in~\cref{fig:limitations}.

Our co-visibility computation relies on static scene geometry, which cannot properly account for moving objects. When dynamic objects (such as vehicles or pedestrians) appear in different positions in image pairs, our method may incorrectly label pixels as co-visible simply because they occupy the same 3D space, even though they correspond to different objects. This limitation particularly affects urban scenes where temporary occlusions and moving objects are common.

While this does not invalidate our benchmark's utility for evaluating the methods, it does suggest potential areas for improvement in co-visibility estimation, particularly for dynamic scene understanding. Future work could explore incorporating instance segmentation or temporal consistency checks to better handle dynamic objects when computing co-visibility maps.

\section{Visualization of Geometric Criteria}
\label{sec:visualizations}
\begin{figure*}[ht]
    \centering
    \textbf{Overlap (\%)}
    \begin{minipage}{\linewidth}
        \centering
        \caption*{Very high overlap (80--100\%)}
        \vspace{-0.3cm}
        \begin{subfigure}{0.3\linewidth}
            \includegraphics[width=\linewidth]{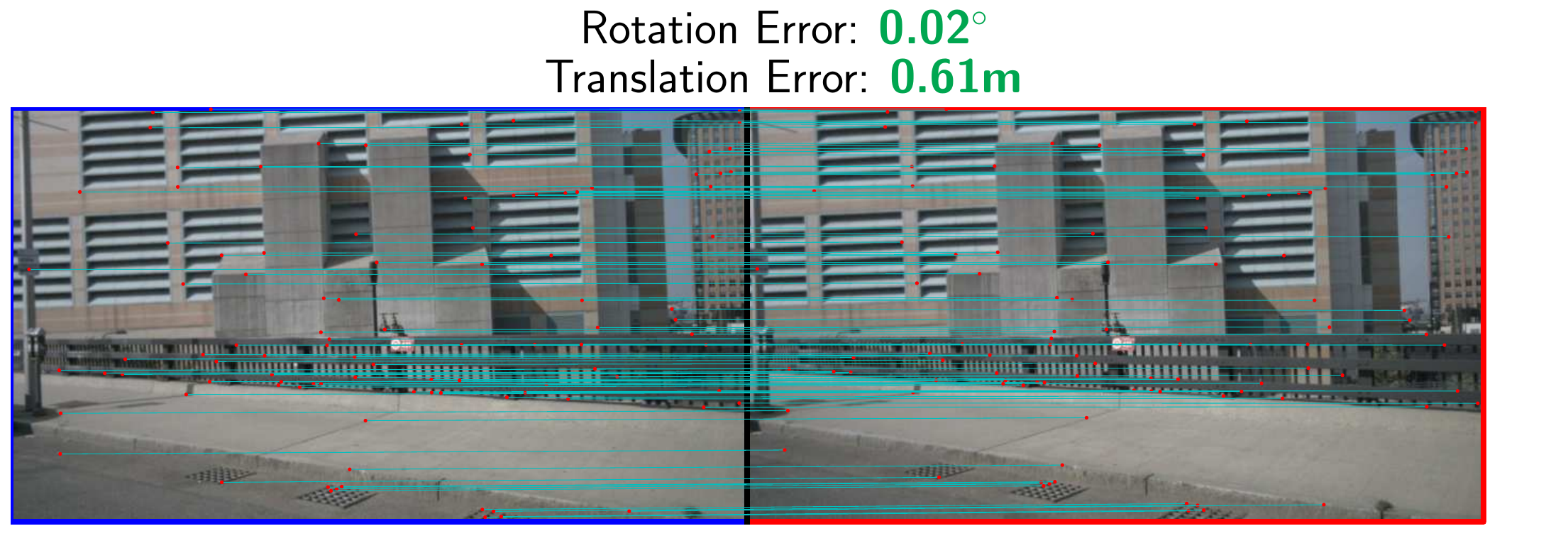}
        \end{subfigure}
        \begin{subfigure}{0.3\linewidth}
            \includegraphics[width=\linewidth]{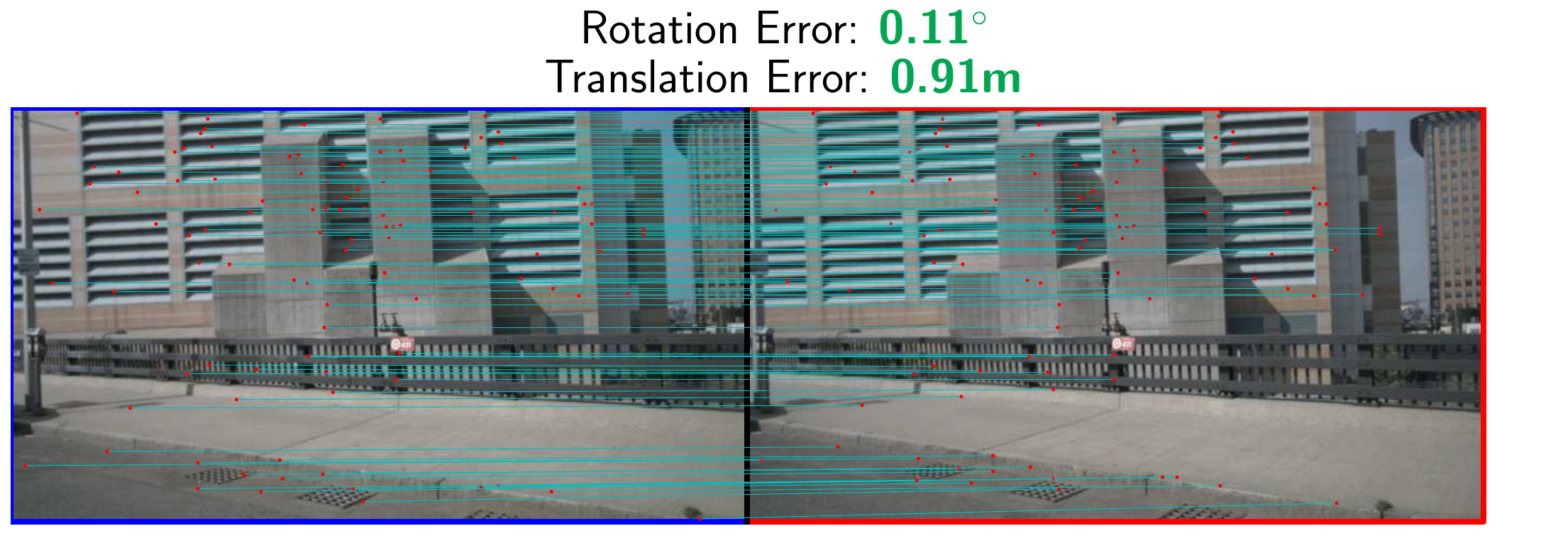}
        \end{subfigure} 
        \\
        \begin{subfigure}{0.3\linewidth}
            \includegraphics[width=\linewidth]{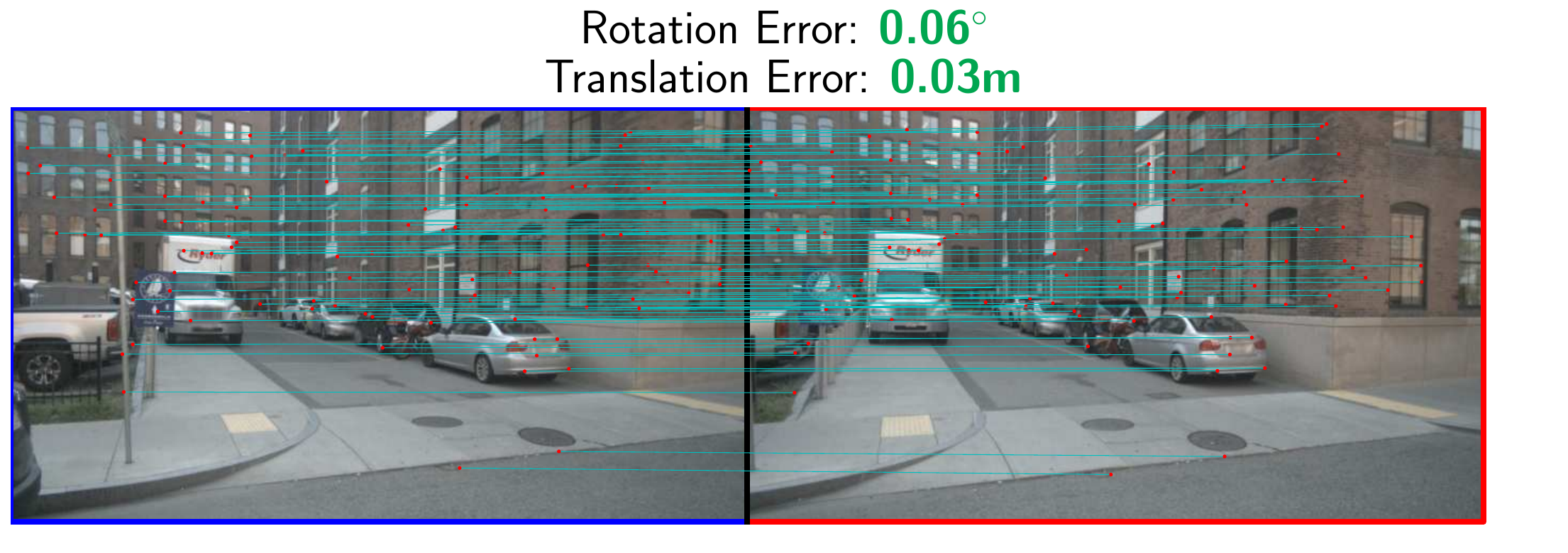}
        \end{subfigure}
        \begin{subfigure}{0.3\linewidth}
            \includegraphics[width=\linewidth]{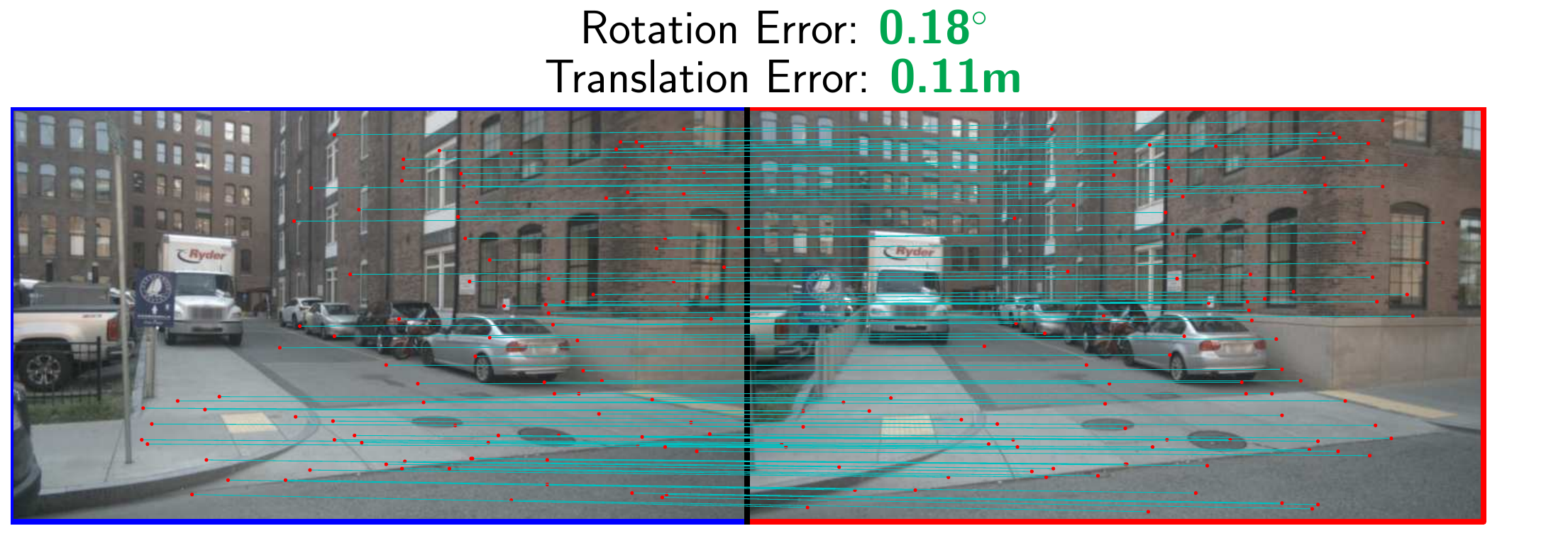}
        \end{subfigure}
        \vspace{-0.3cm}       
    \end{minipage}

    \begin{minipage}{\linewidth}
        \centering
        \caption*{High overlap (60--80\%)}
        \begin{subfigure}{0.3\linewidth}
            \includegraphics[width=\linewidth]{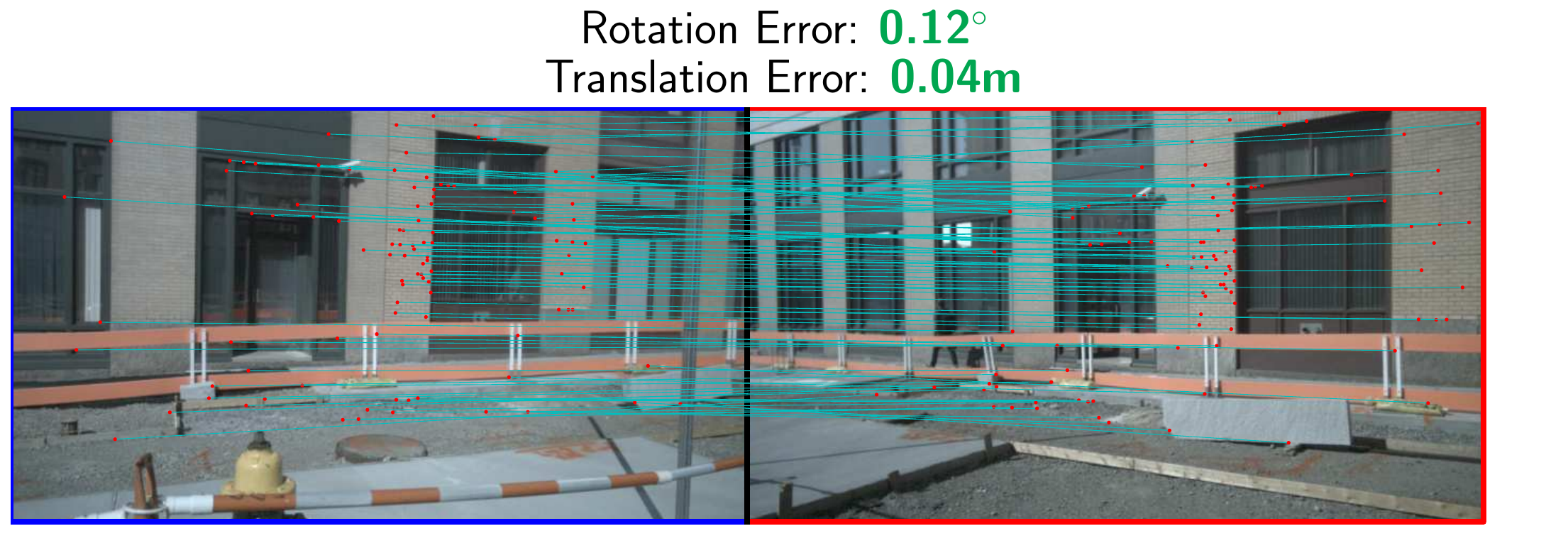}
        \end{subfigure}
        \begin{subfigure}{0.3\linewidth}
            \includegraphics[width=\linewidth]{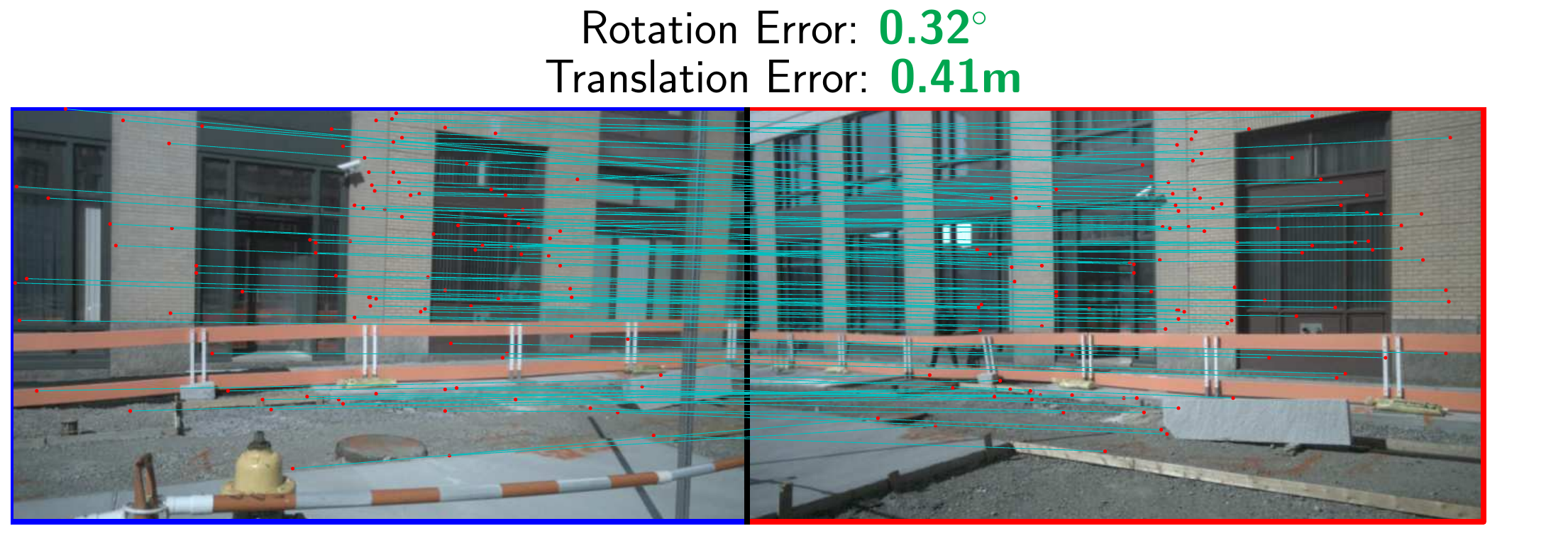}
        \end{subfigure} 
        \\
        \begin{subfigure}{0.3\linewidth}
            \includegraphics[width=\linewidth]{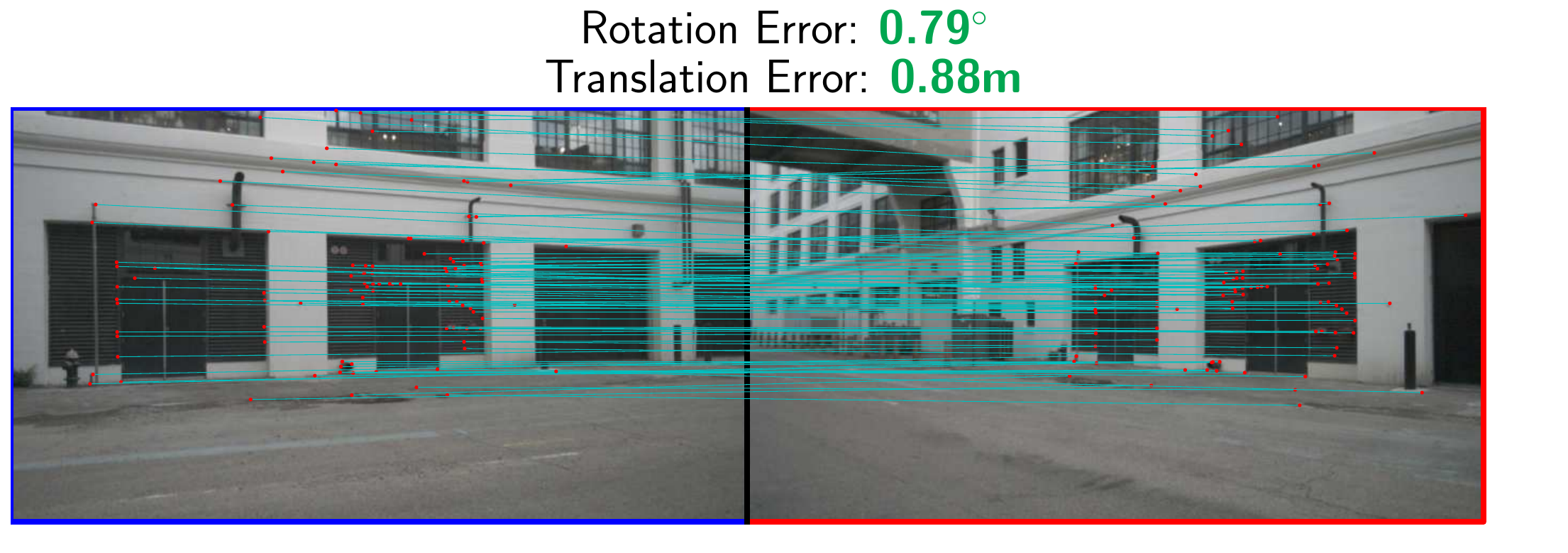}
        \end{subfigure}
        \begin{subfigure}{0.3\linewidth}
            \includegraphics[width=\linewidth]{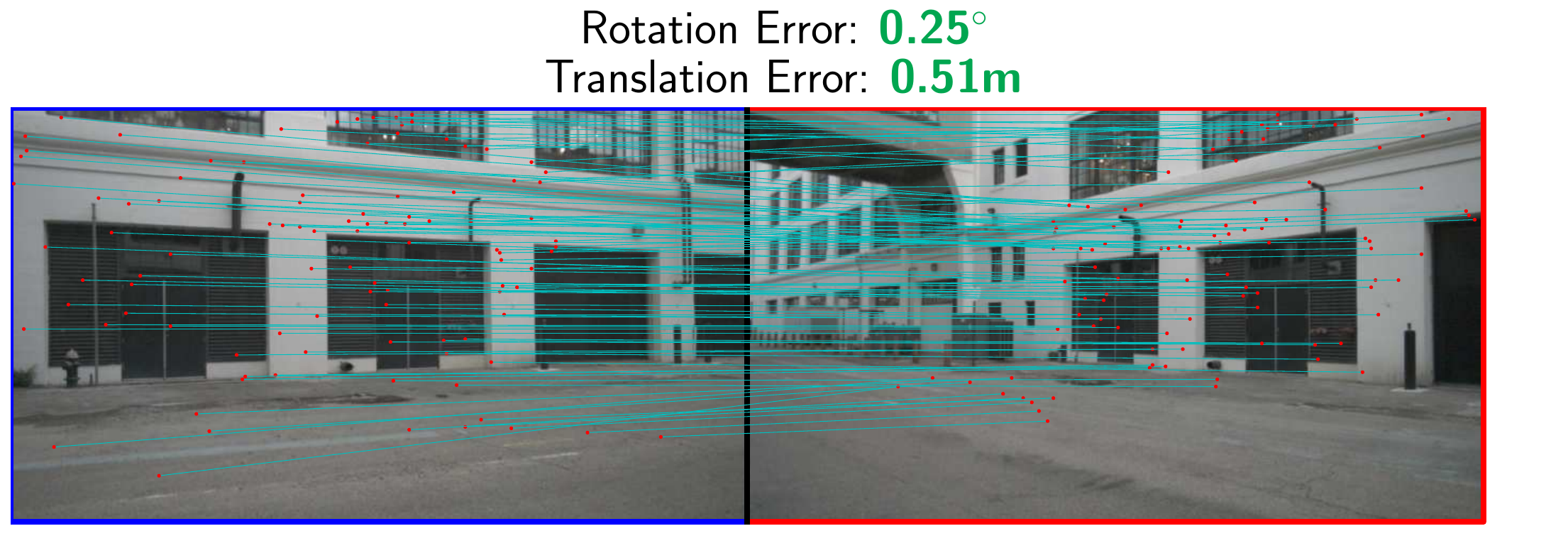}
        \end{subfigure} 
        \vspace{-0.3cm}
    \end{minipage}

    \begin{minipage}{\linewidth}
        \centering
        \caption*{Medium overlap (40--60\%)}
        \begin{subfigure}{0.3\linewidth}
            \includegraphics[width=\linewidth]{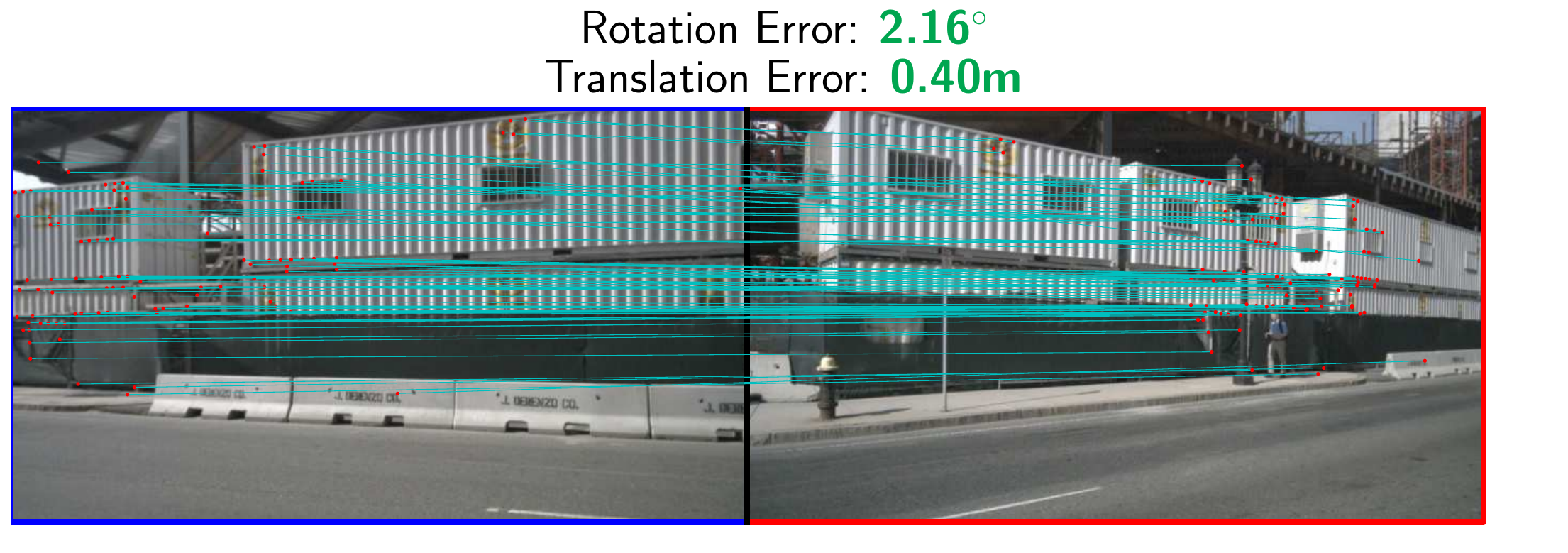}
        \end{subfigure}
        \begin{subfigure}{0.3\linewidth}
            \includegraphics[width=\linewidth]{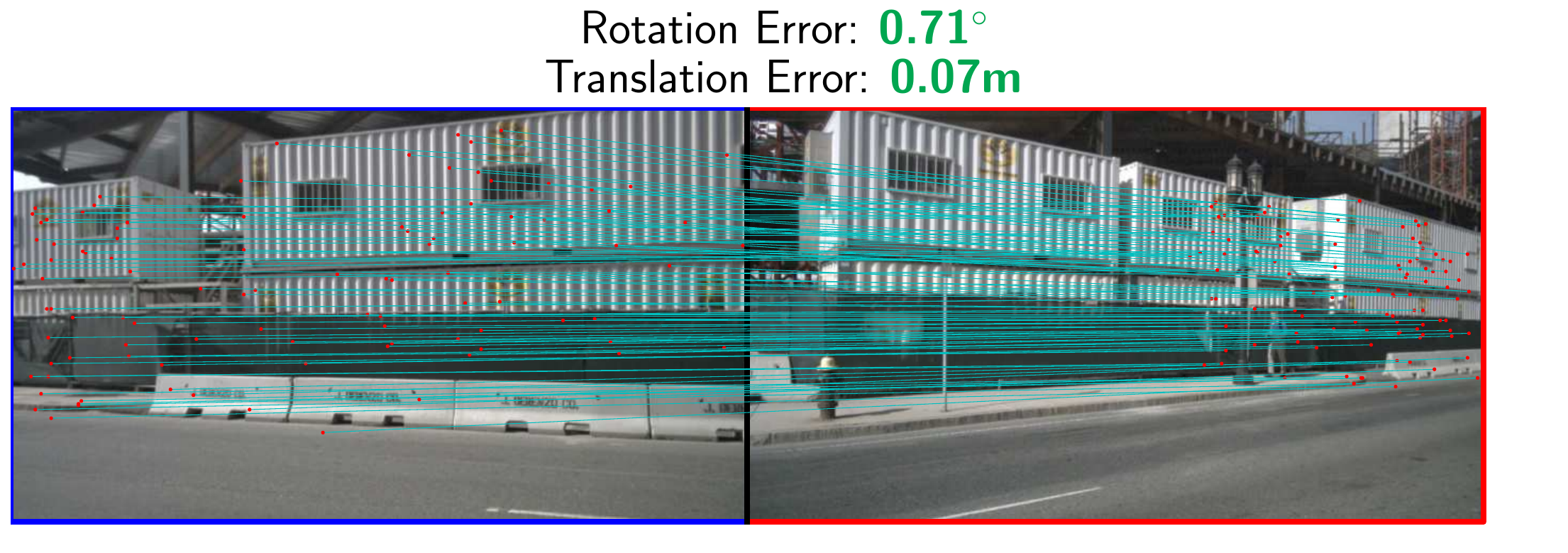}
        \end{subfigure} 
        \\
        \begin{subfigure}{0.3\linewidth}
            \includegraphics[width=\linewidth]{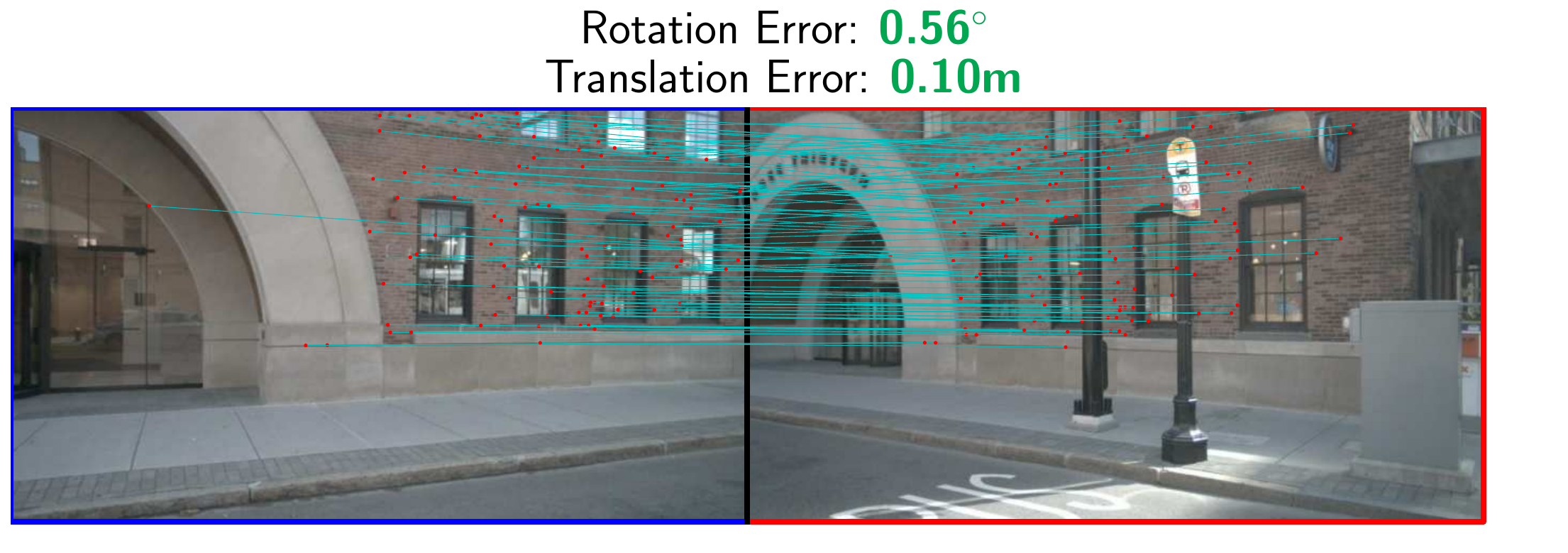}
        \end{subfigure}
        \begin{subfigure}{0.3\linewidth}
            \includegraphics[width=\linewidth]{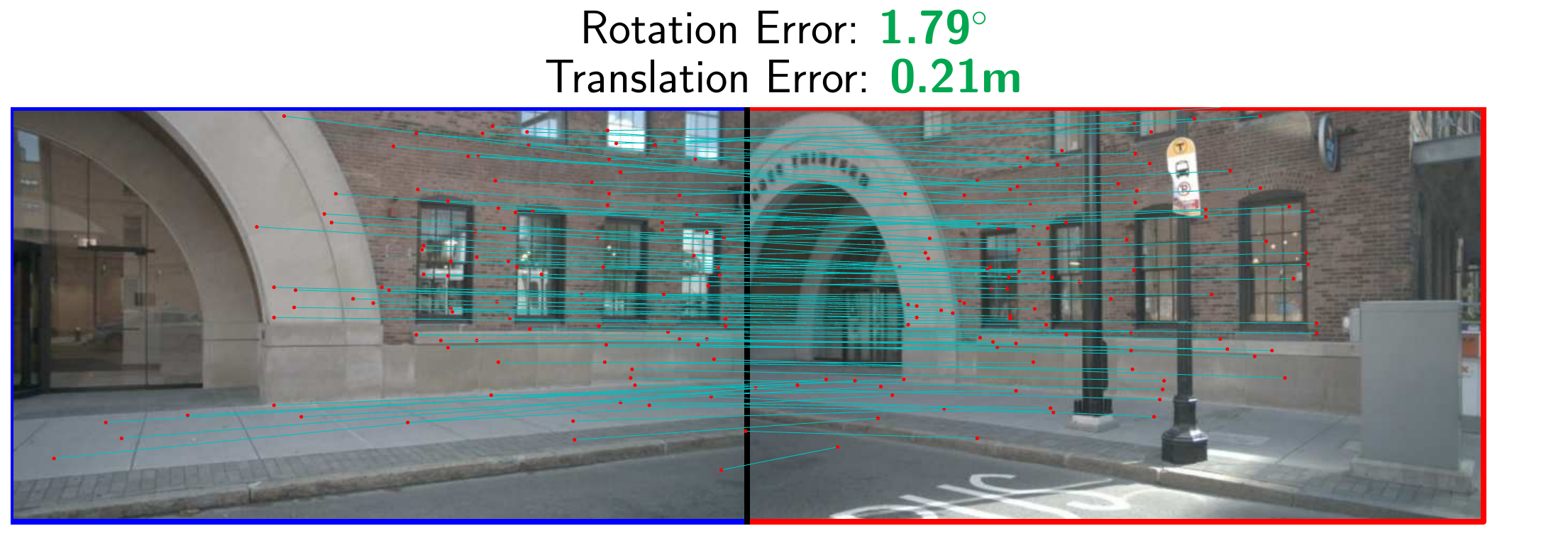}
        \end{subfigure} 
        \vspace{-0.3cm}
    \end{minipage}

    \begin{minipage}{\linewidth}
        \centering
        \caption*{Low overlap (20--40\%)}
        \begin{subfigure}{0.3\linewidth}
            \includegraphics[width=\linewidth]{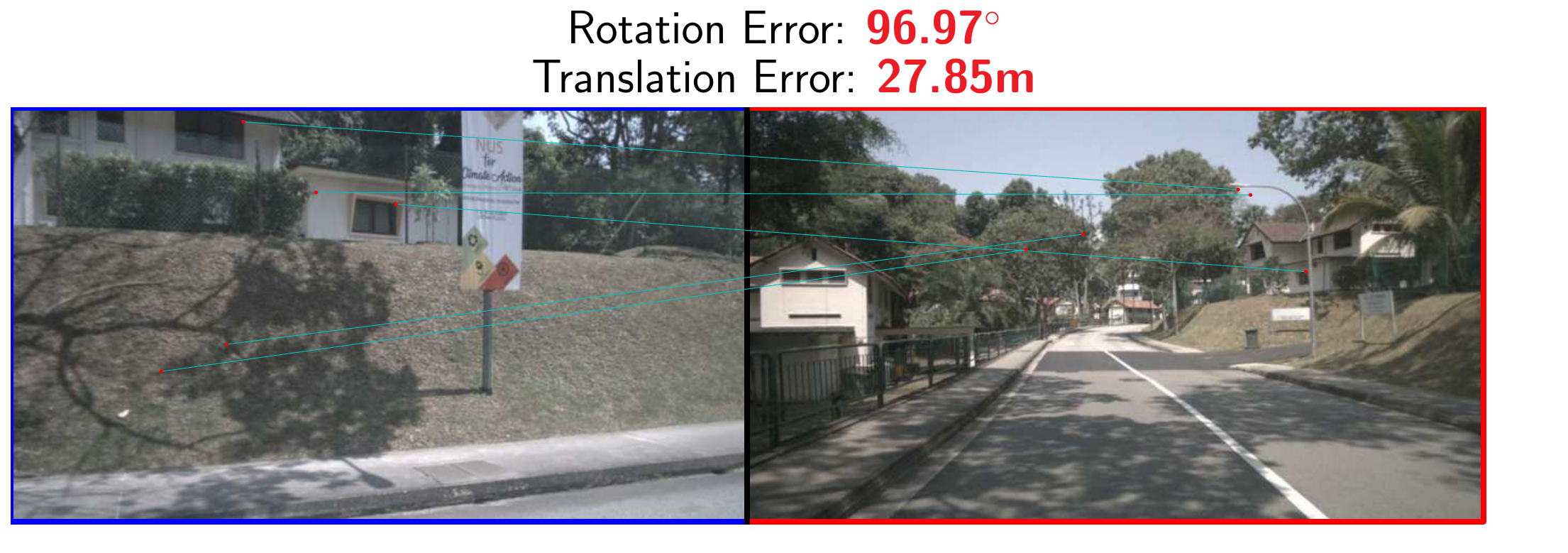}
        \end{subfigure}
        \begin{subfigure}{0.3\linewidth}
            \includegraphics[width=\linewidth]{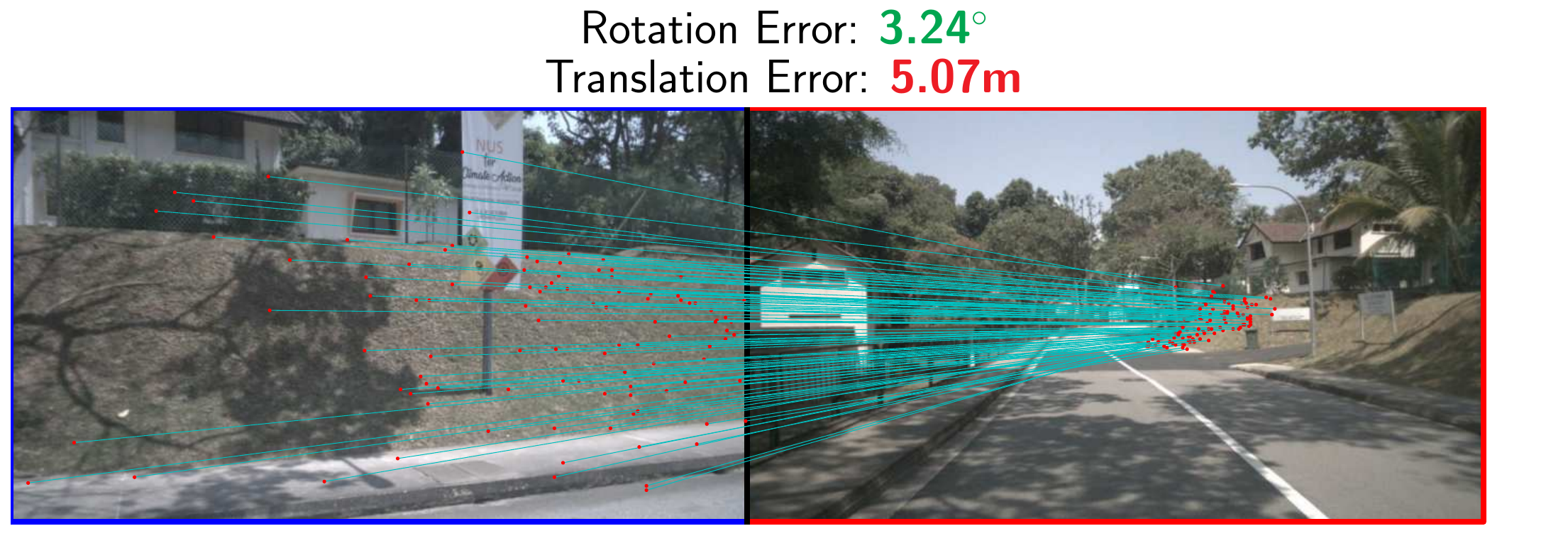}
        \end{subfigure} 
        \\
        \begin{subfigure}{0.3\linewidth}
            \includegraphics[width=\linewidth]{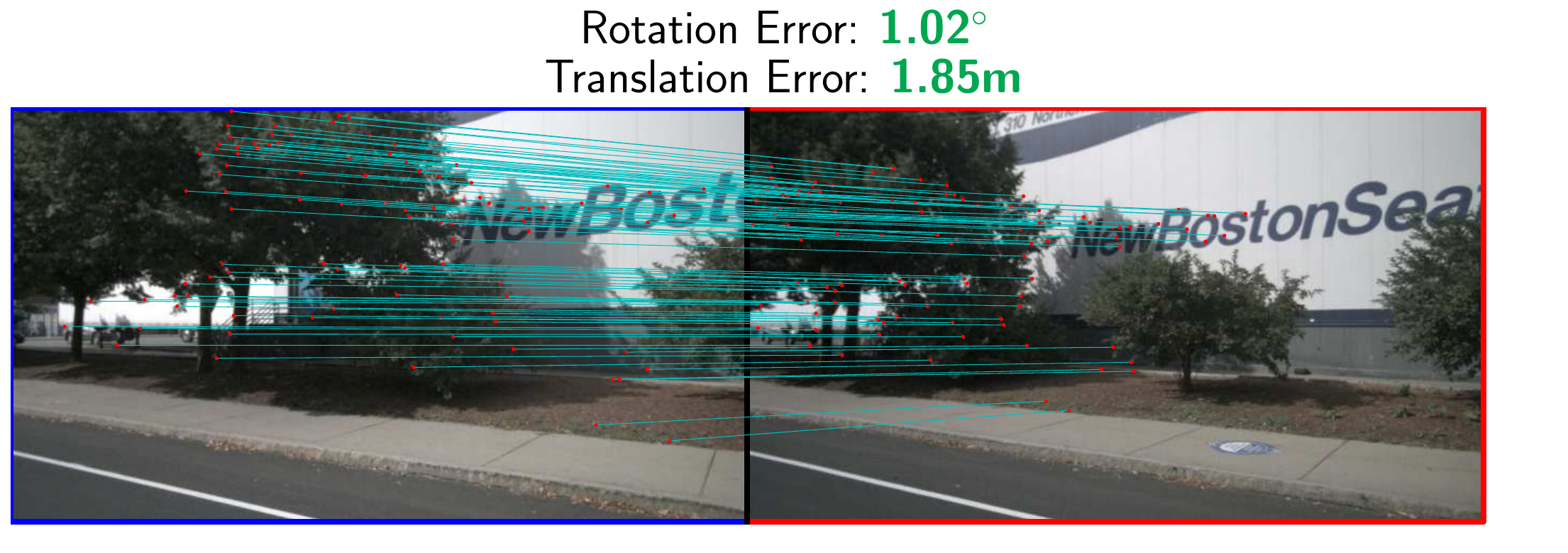}
        \end{subfigure}
        \begin{subfigure}{0.3\linewidth}
            \includegraphics[width=\linewidth]{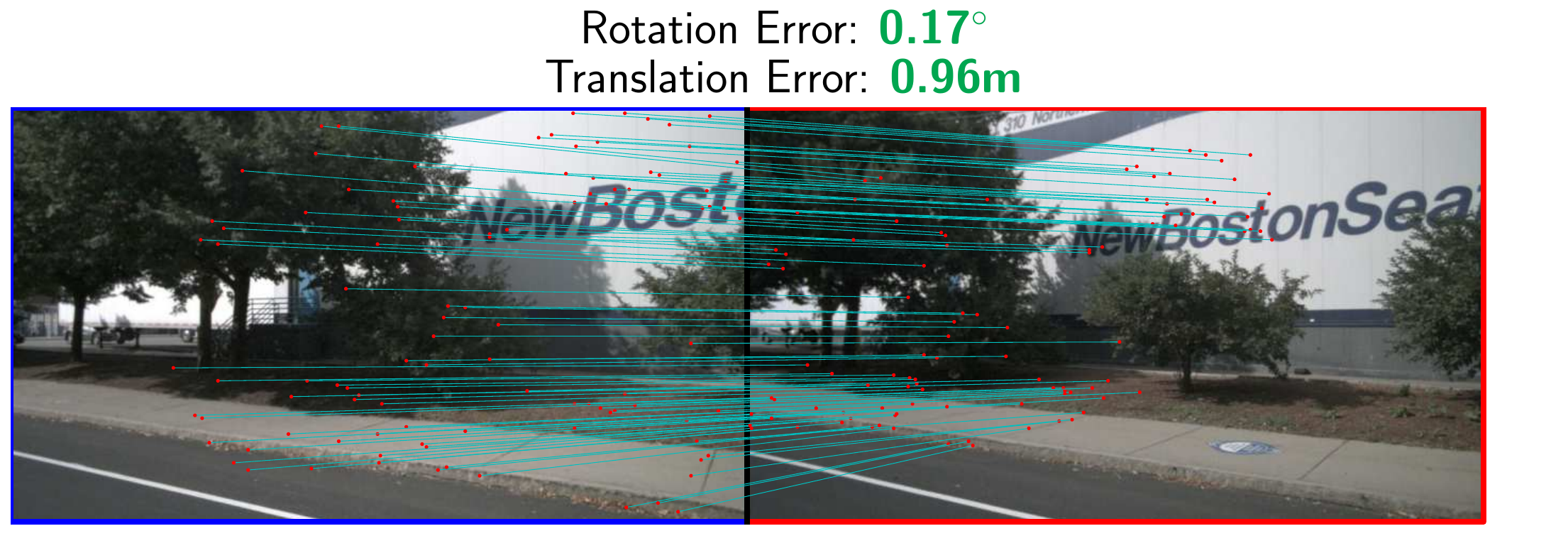}
        \end{subfigure} 
        \vspace{-0.3cm}
    \end{minipage}

    \begin{minipage}{\linewidth}
        \centering
        \caption*{Very low overlap (5--20\%)}
        \begin{subfigure}{0.3\linewidth}
            \includegraphics[width=\linewidth]{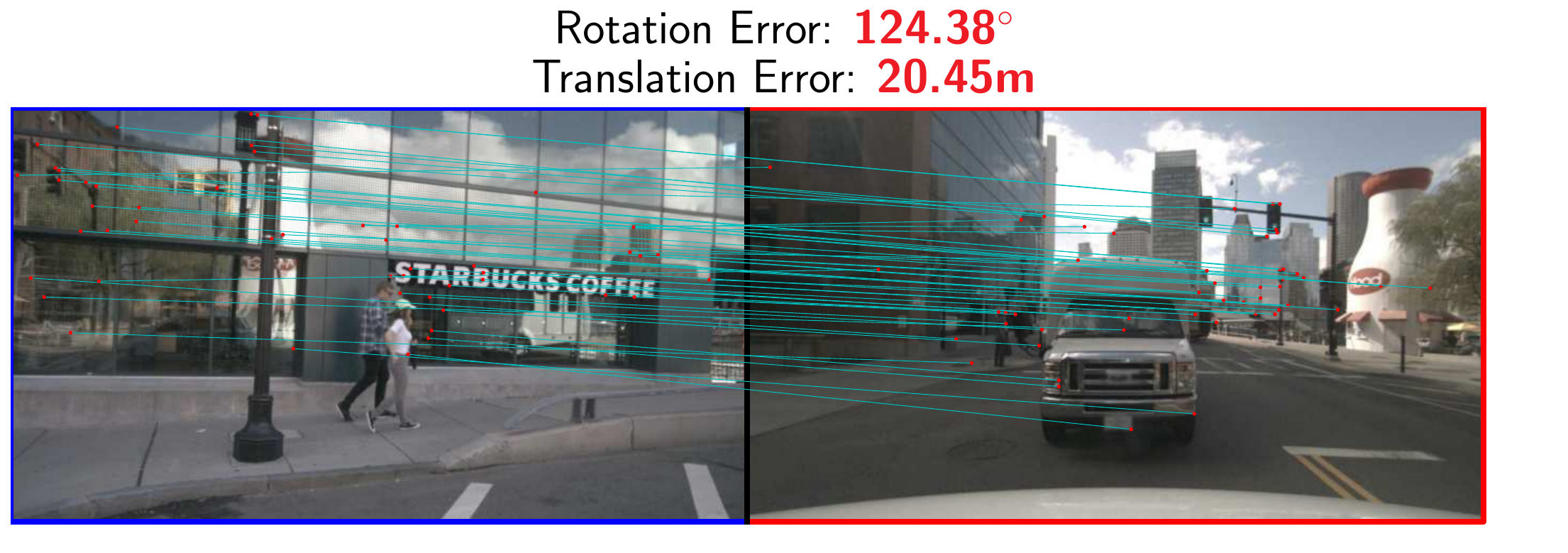}
        \end{subfigure}
        \begin{subfigure}{0.3\linewidth}
            \includegraphics[width=\linewidth]{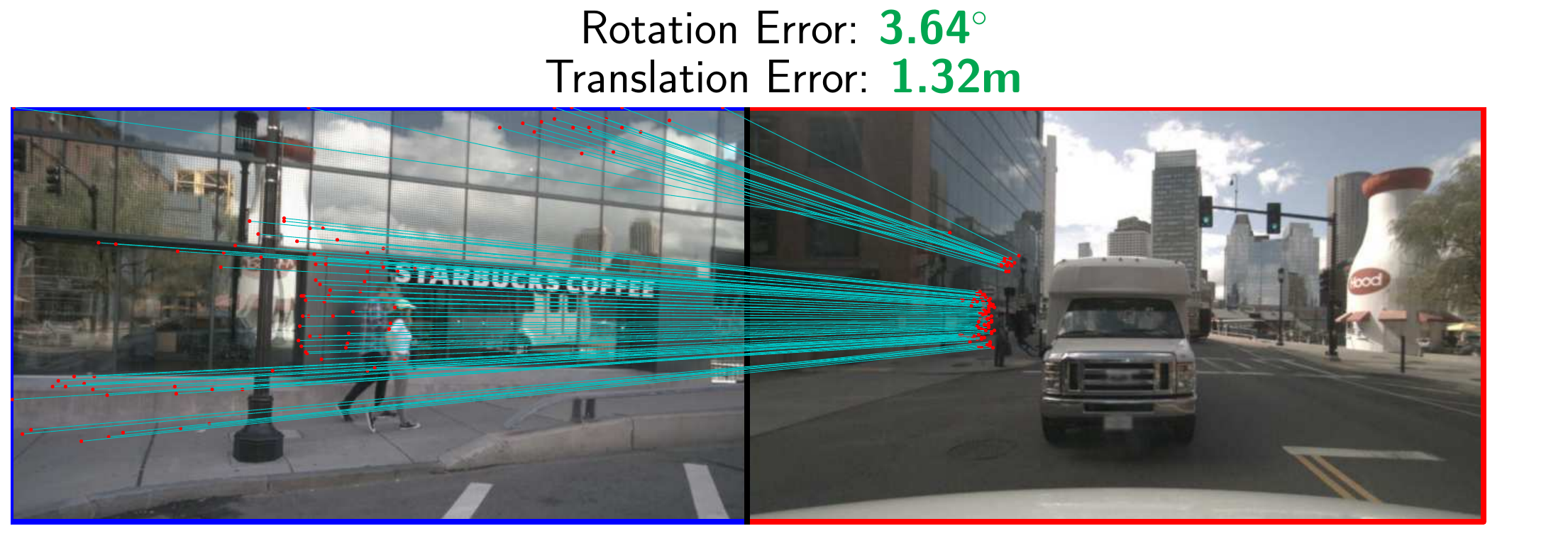}
        \end{subfigure} 
        \\
        \begin{subfigure}{0.3\linewidth}
            \includegraphics[width=\linewidth]{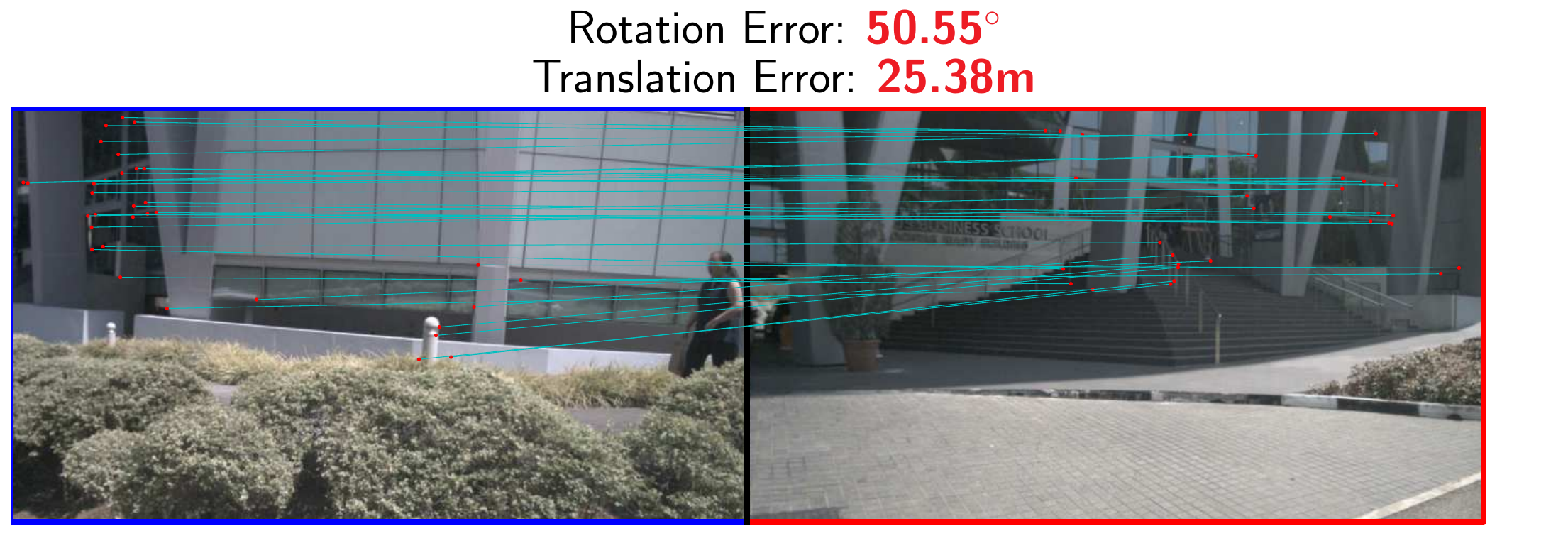}
            \caption{ALIKED+LightGlue}
        \end{subfigure}
        \begin{subfigure}{0.3\linewidth}
            \includegraphics[width=\linewidth]{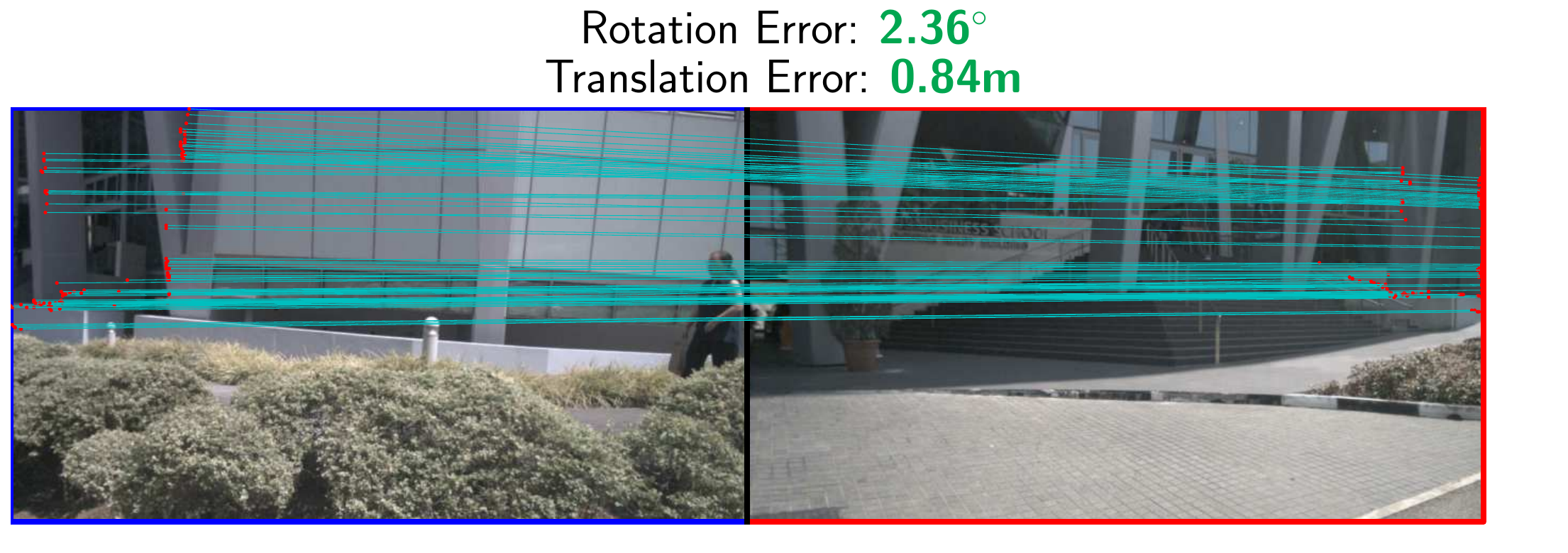}
            \caption{DUSt3R}
        \end{subfigure} 
        \vspace{-0.3cm}
    \end{minipage}
    
    \caption{\textbf{Examples of image pairs with varying overlap -- }For each overlap range, we show two random image pairs for the best methods in either detector-based (ALIKED+LightGlue on the left) or detector-free (DUSt3R on the right) approaches.}
    \label{fig:overlap_examples}
\end{figure*}

\begin{figure*}[ht]
    \centering
    \textbf{Scale ratio}
    \begin{minipage}{\linewidth}
        \centering
        \caption*{Small scale change (1.0--1.5)}
        \vspace{-0.3cm}
        \begin{subfigure}{0.3\linewidth}
            \includegraphics[width=\linewidth]{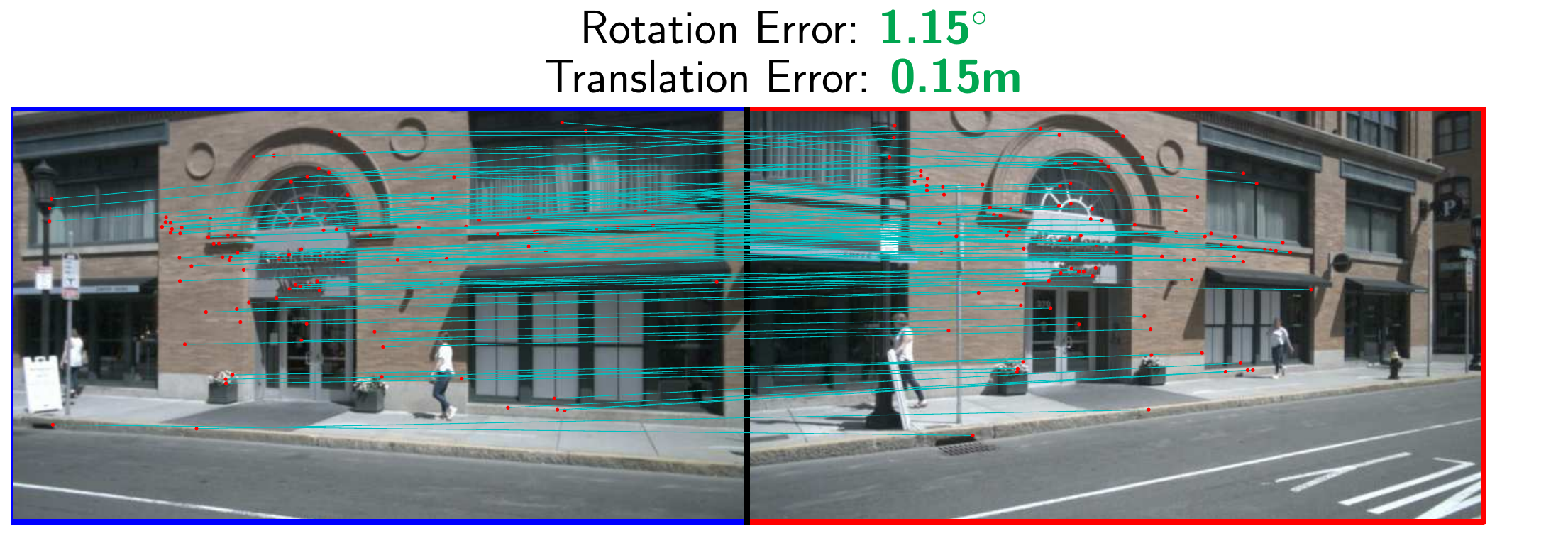}
        \end{subfigure}
        \begin{subfigure}{0.3\linewidth}
            \includegraphics[width=\linewidth]{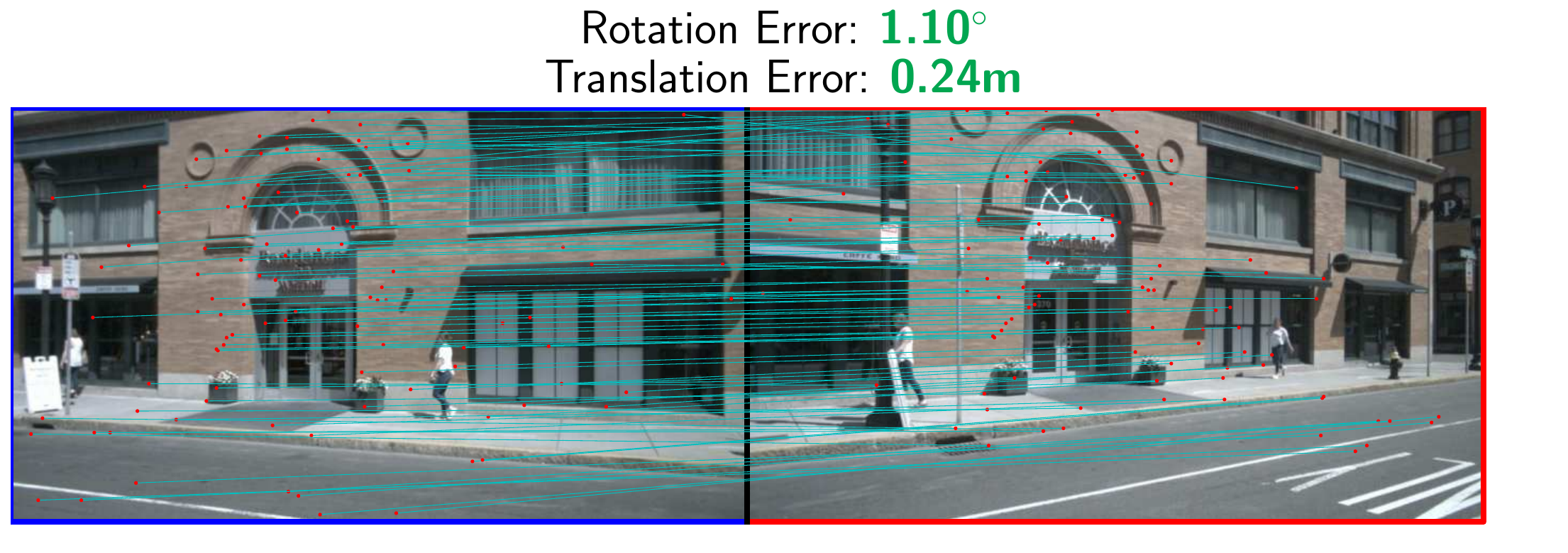}
        \end{subfigure} 
        \\
        \begin{subfigure}{0.3\linewidth}
            \includegraphics[width=\linewidth]{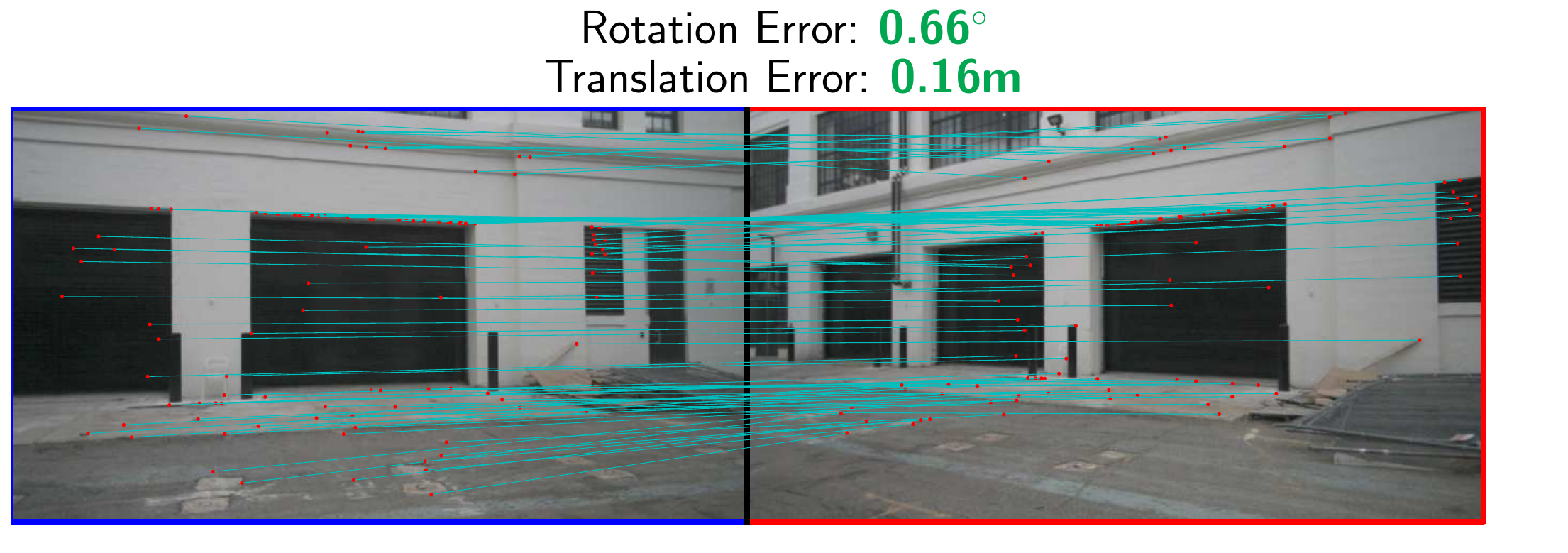}
        \end{subfigure}
        \begin{subfigure}{0.3\linewidth}
            \includegraphics[width=\linewidth]{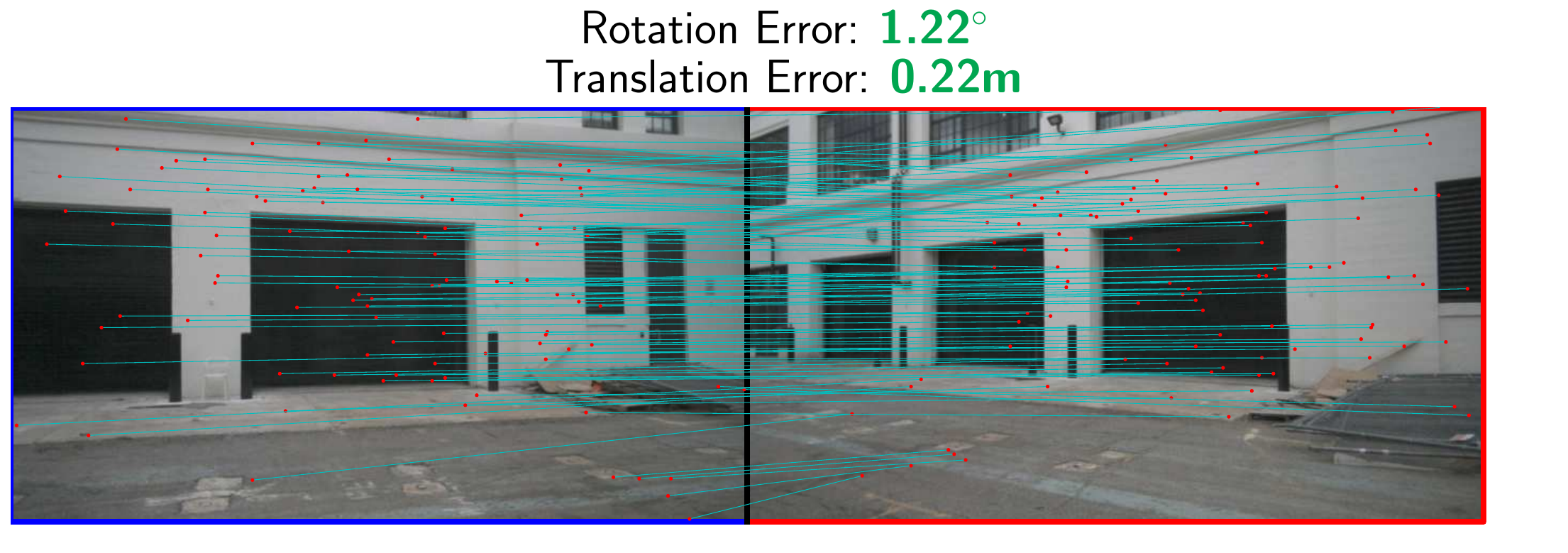}
        \end{subfigure}
        \vspace{-0.3cm}        
    \end{minipage}

    \begin{minipage}{\linewidth}
        \centering
        \caption*{Moderate scale change (1.5--2.5)}
        \begin{subfigure}{0.3\linewidth}
            \includegraphics[width=\linewidth]{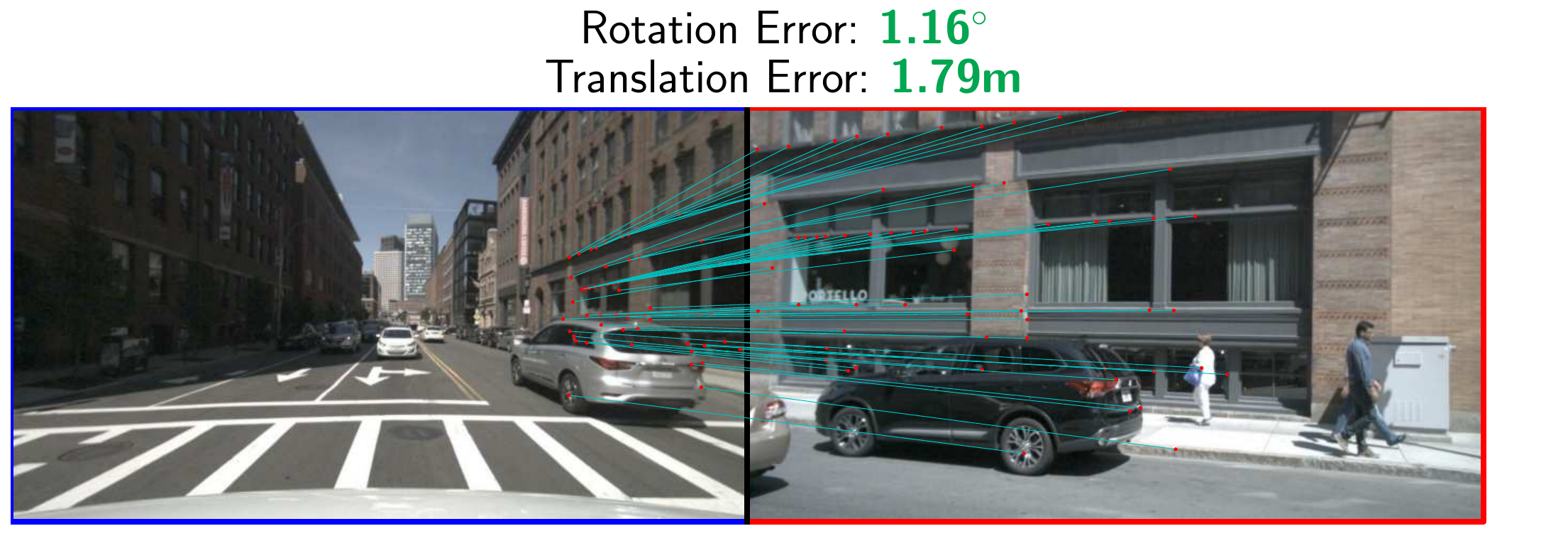}
        \end{subfigure}
        \begin{subfigure}{0.3\linewidth}
            \includegraphics[width=\linewidth]{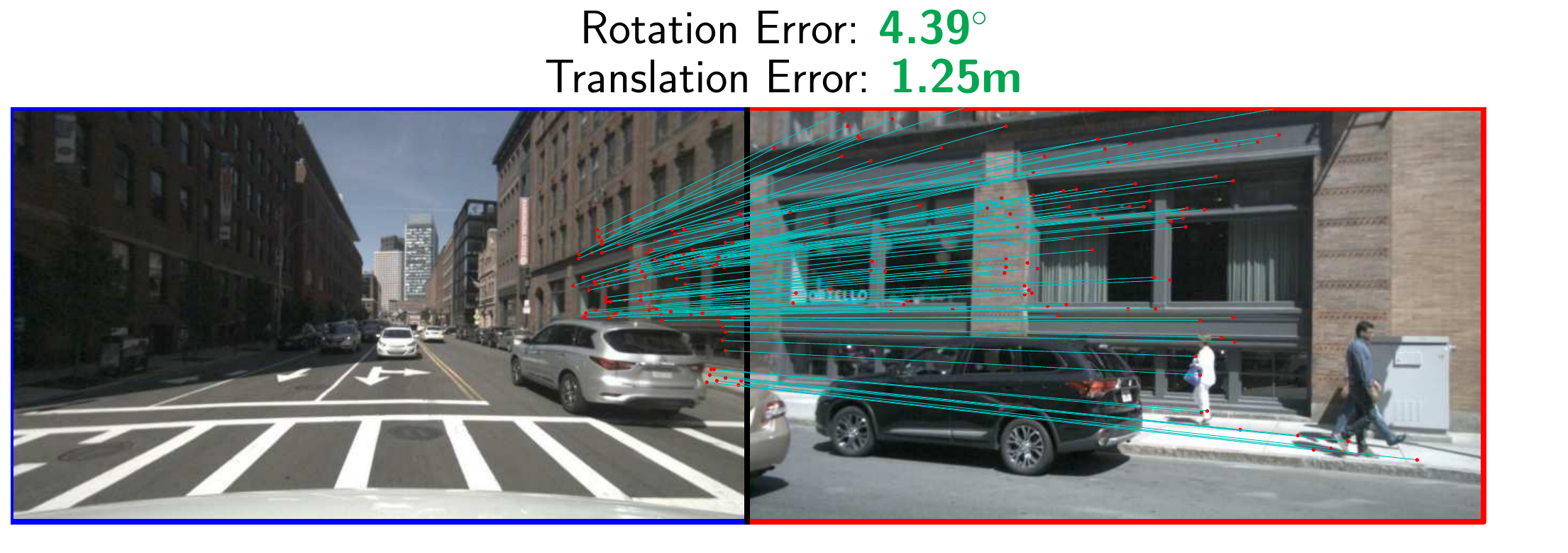}
        \end{subfigure} 
        \\
        \begin{subfigure}{0.3\linewidth}
            \includegraphics[width=\linewidth]{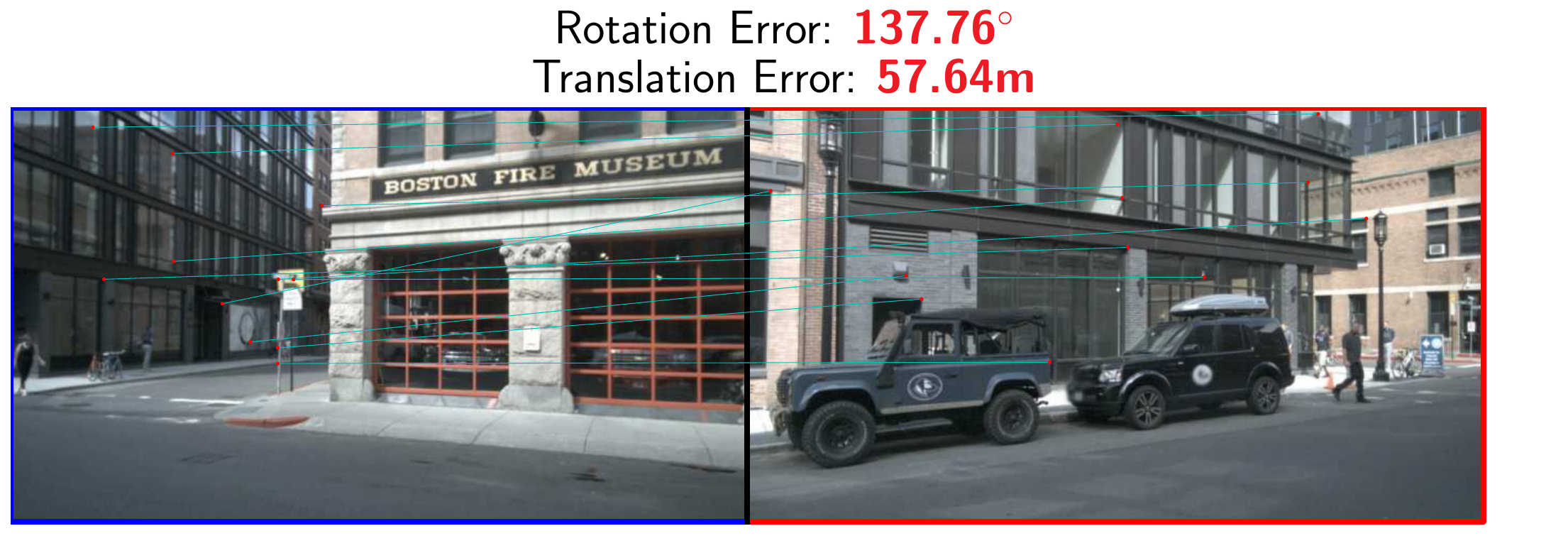}
        \end{subfigure}
        \begin{subfigure}{0.3\linewidth}
            \includegraphics[width=\linewidth]{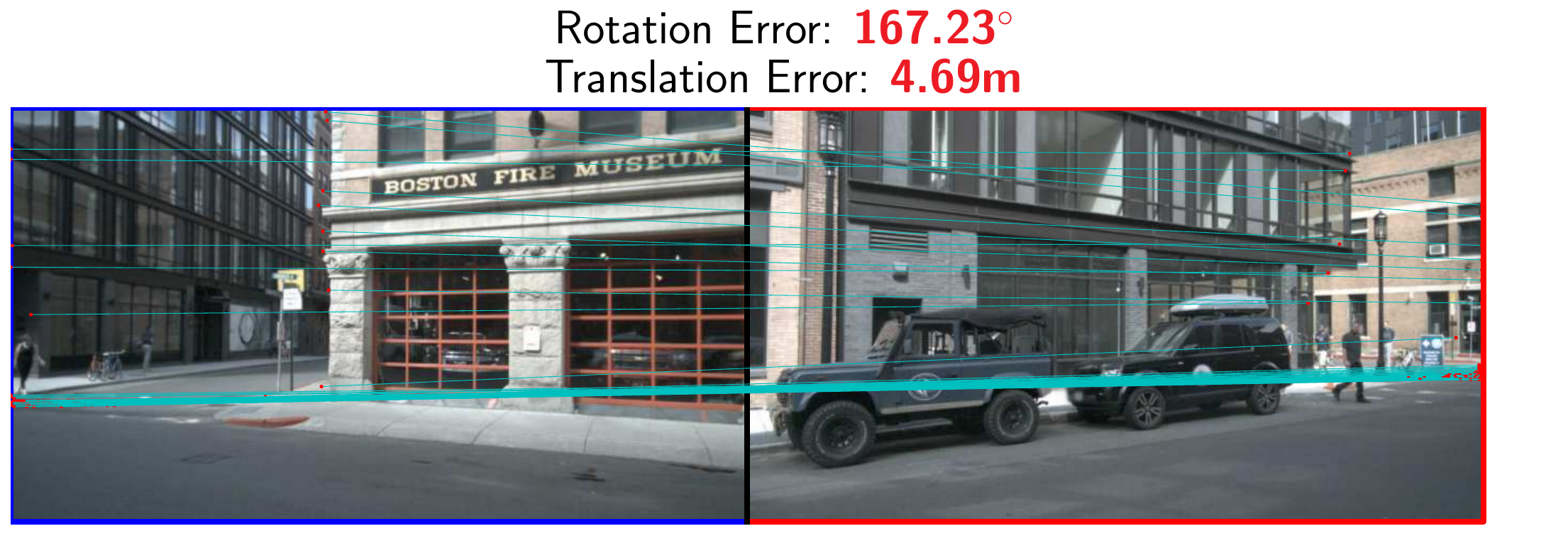}
        \end{subfigure} 
        \vspace{-0.3cm}
    \end{minipage}

    \begin{minipage}{\linewidth}
        \centering
        \caption*{Large scale change (2.5--4.0)}
        \begin{subfigure}{0.3\linewidth}
            \includegraphics[width=\linewidth]{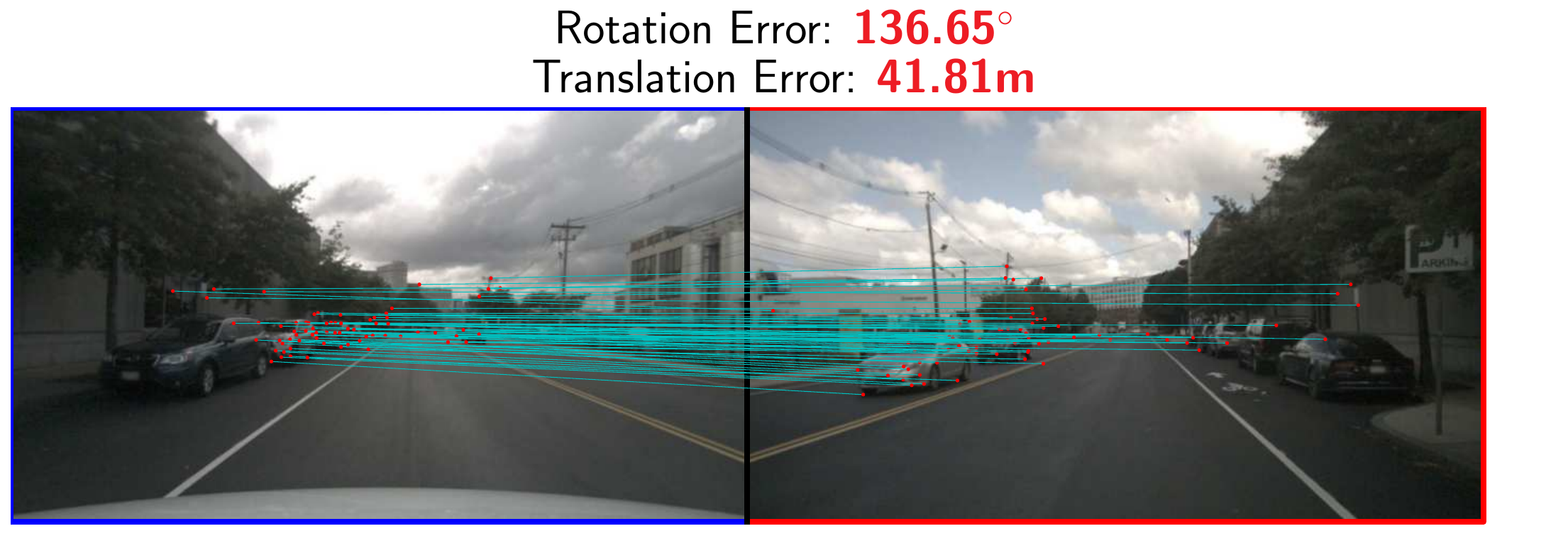}
        \end{subfigure}
        \begin{subfigure}{0.3\linewidth}
            \includegraphics[width=\linewidth]{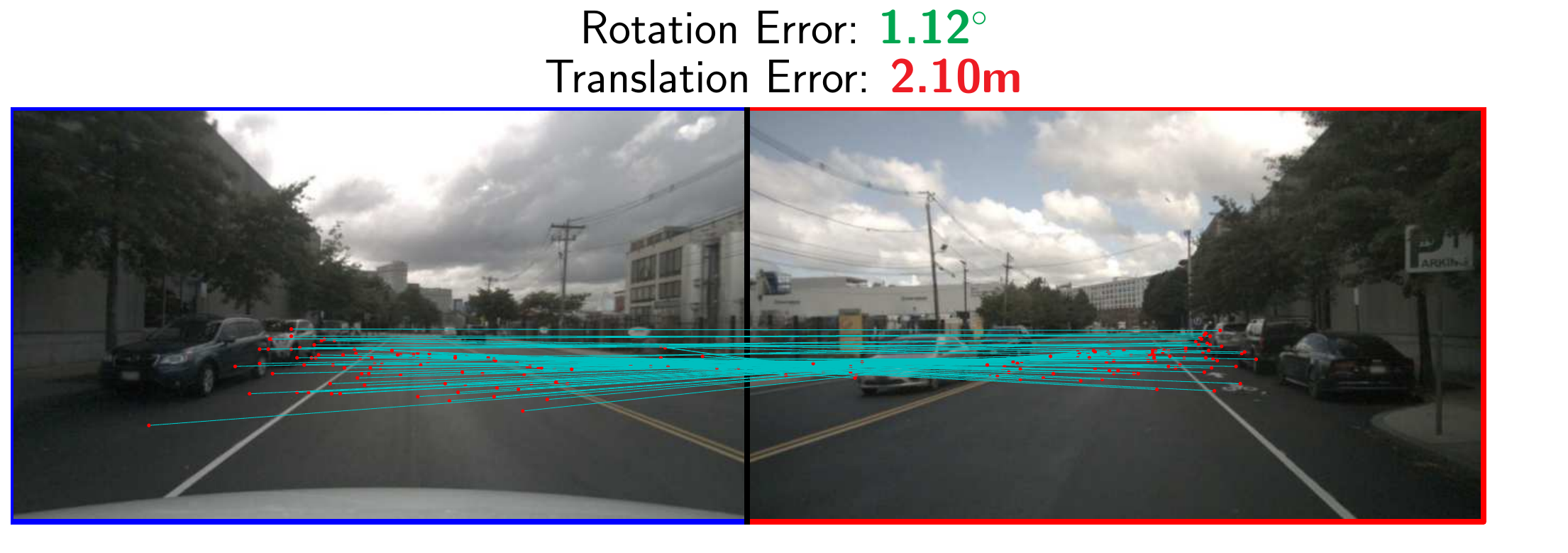}
        \end{subfigure} 
        \\
        \begin{subfigure}{0.3\linewidth}
            \includegraphics[width=\linewidth]{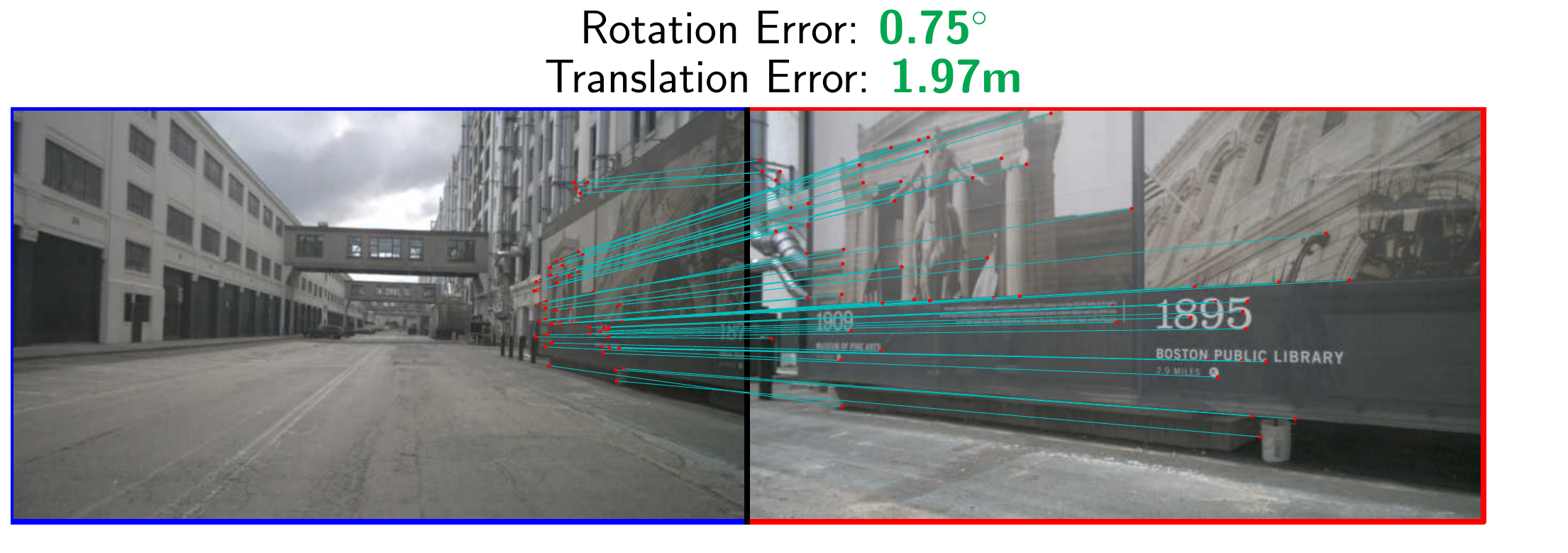}
        \end{subfigure}
        \begin{subfigure}{0.3\linewidth}
            \includegraphics[width=\linewidth]{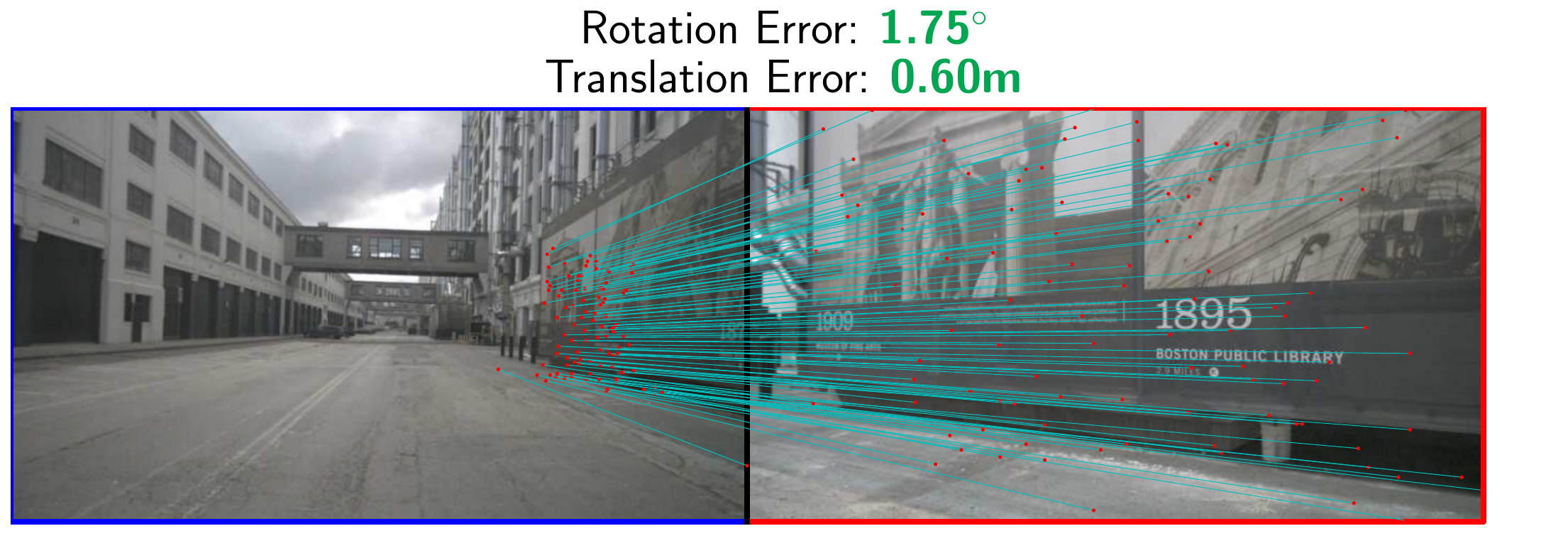}
        \end{subfigure} 
        \vspace{-0.3cm}
    \end{minipage}

    \begin{minipage}{\linewidth}
        \centering
        \caption*{Very large scale change (4.0--6.0)}
        \begin{subfigure}{0.3\linewidth}
            \includegraphics[width=\linewidth]{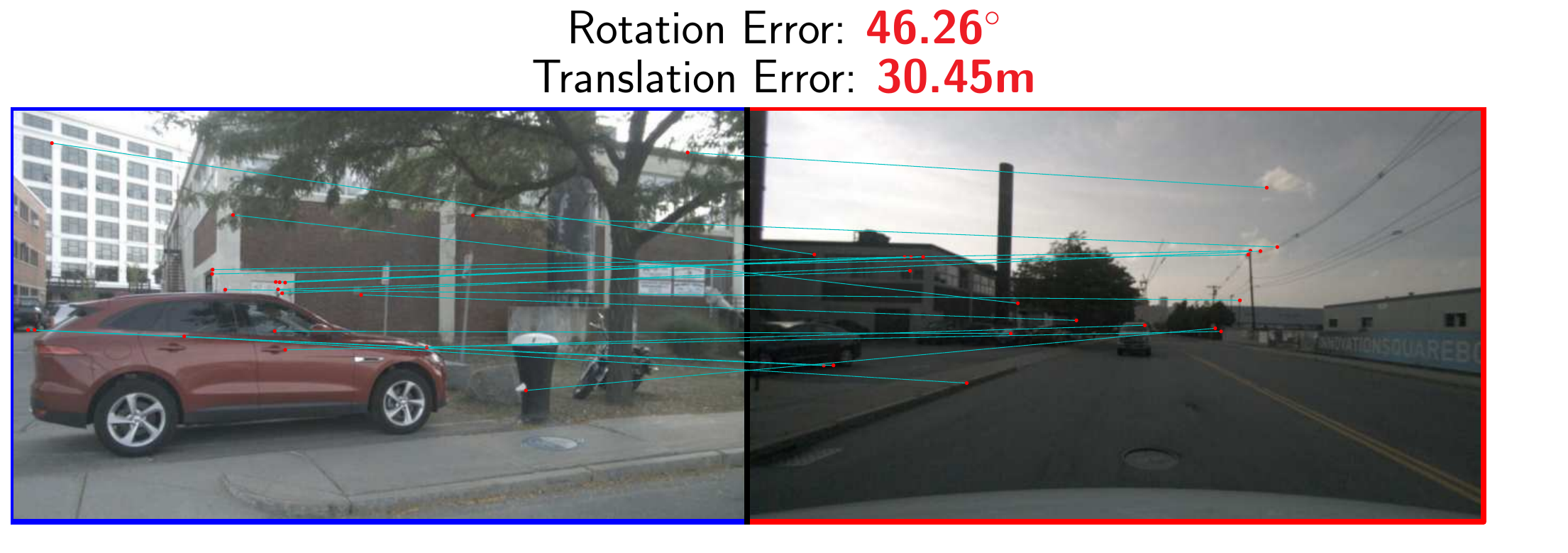}
        \end{subfigure}
        \begin{subfigure}{0.3\linewidth}
            \includegraphics[width=\linewidth]{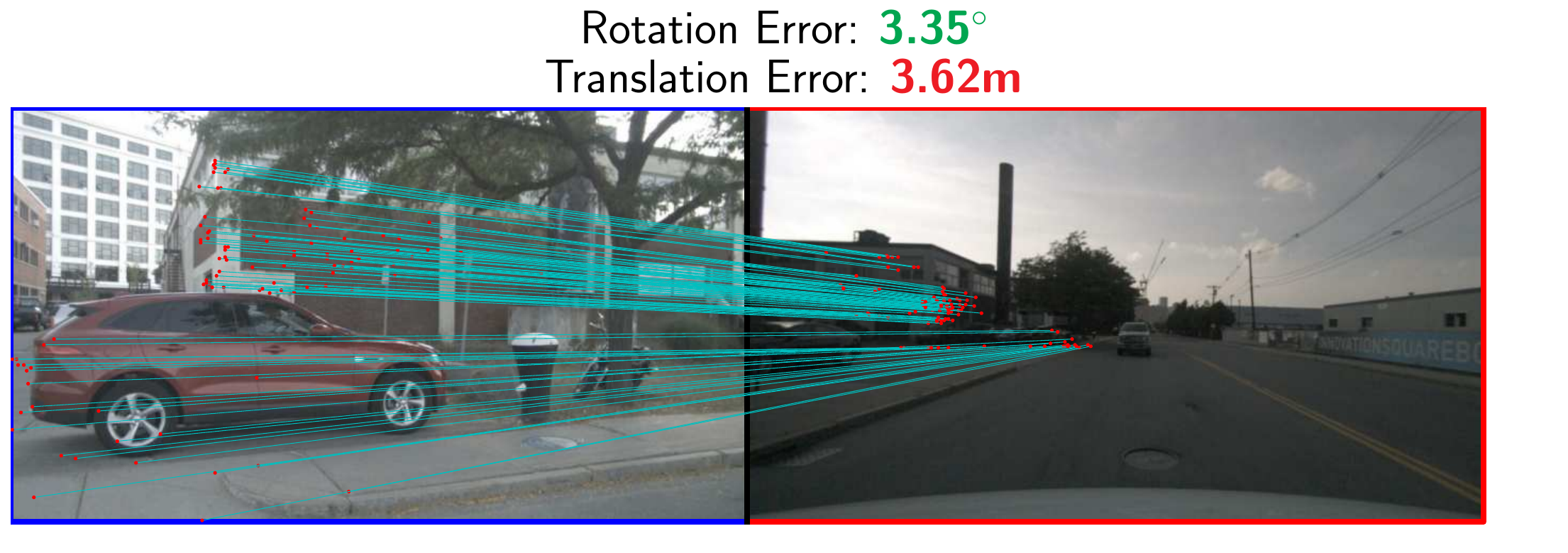}
        \end{subfigure} 
        \\
        \begin{subfigure}{0.3\linewidth}
            \includegraphics[width=\linewidth]{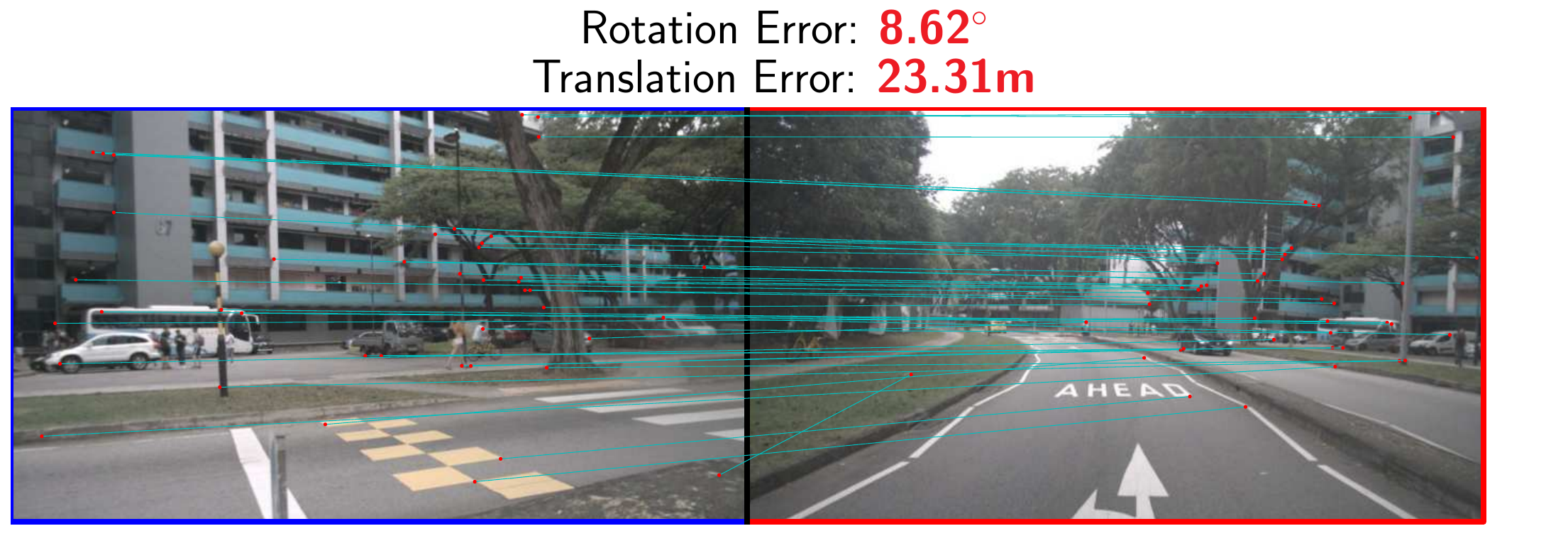}
            \caption{ALIKED+LightGlue}
        \end{subfigure}
        \begin{subfigure}{0.3\linewidth}
            \includegraphics[width=\linewidth]{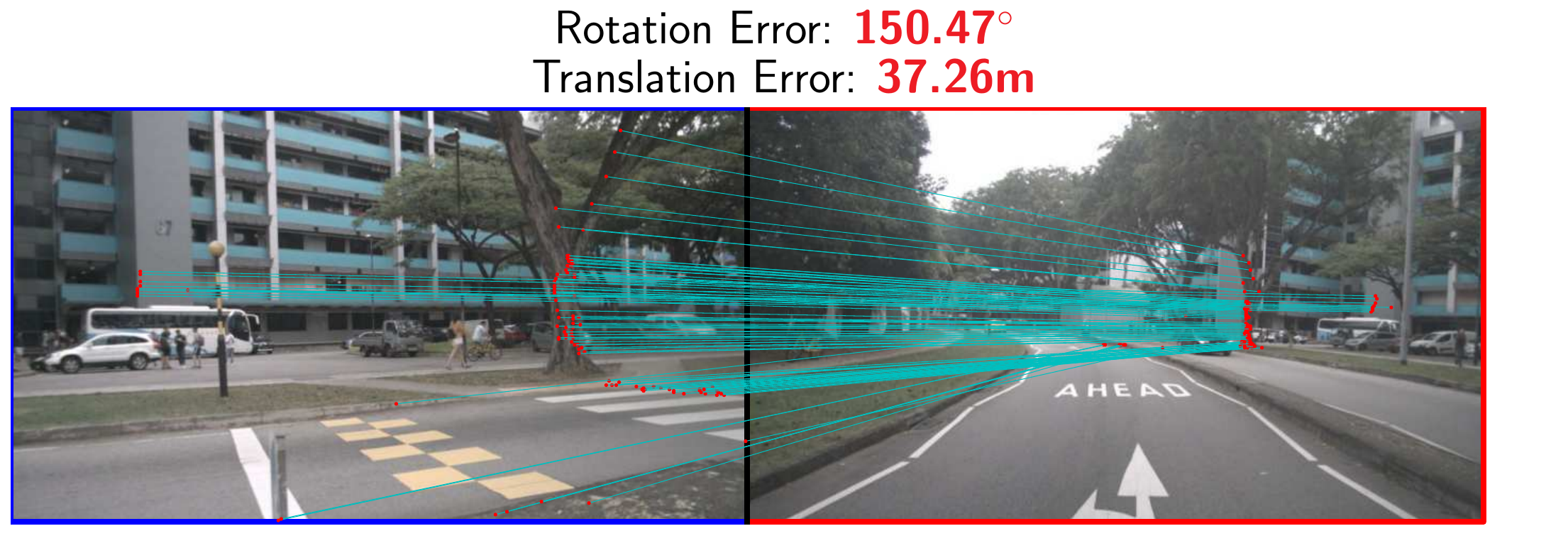}
            \caption{DUSt3R}
        \end{subfigure} 
        \vspace{-0.3cm}
    \end{minipage}
    
    \caption{\textbf{Examples of image pairs with varying scale ratios -- }For each scale ratio range, we show two random image pairs for the best methods in either detector-based (ALIKED+LightGlue on the left) or detector-free (DUSt3R on the right) approaches.}
    \label{fig:scale_examples}
\end{figure*}

\begin{figure*}[ht]
    \centering
    \textbf{Viewpoint Angle (°)} \\[0.5em]
    \begin{minipage}{\linewidth}
        \centering
        \caption*{Small viewpoint change (0--30°)}
        \vspace{-0.3cm}
        \begin{subfigure}{0.3\linewidth}
            \includegraphics[width=\linewidth]{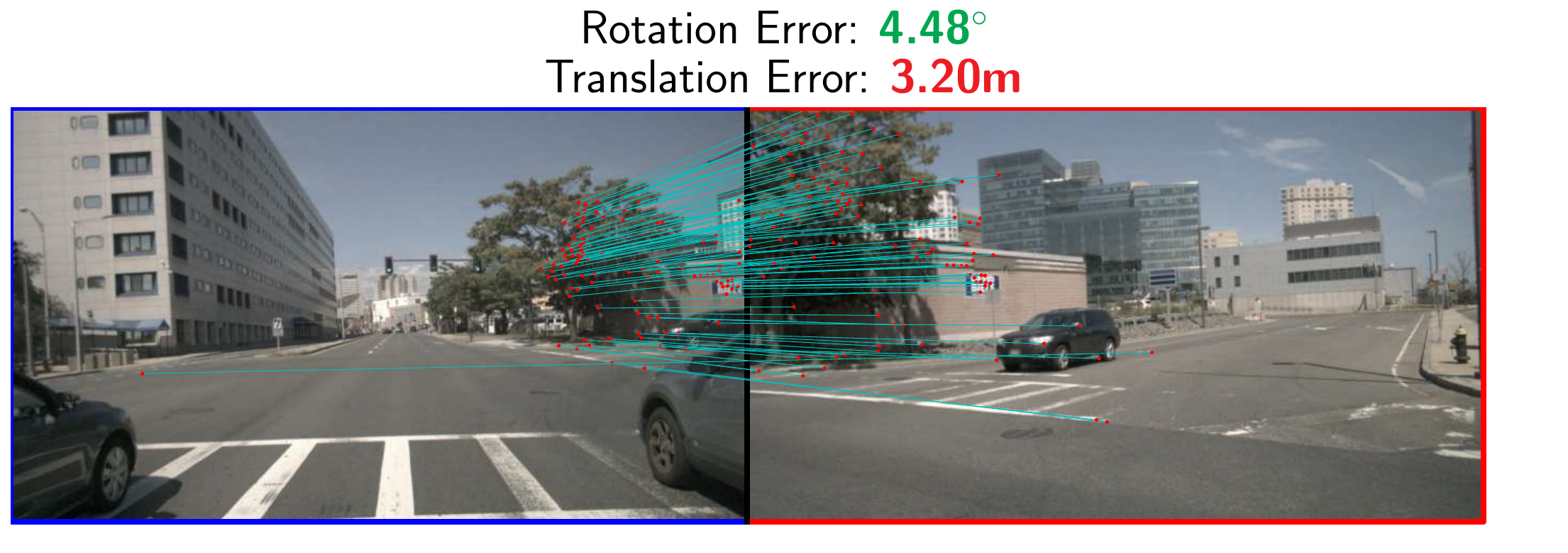}
        \end{subfigure}
        \begin{subfigure}{0.3\linewidth}
            \includegraphics[width=\linewidth]{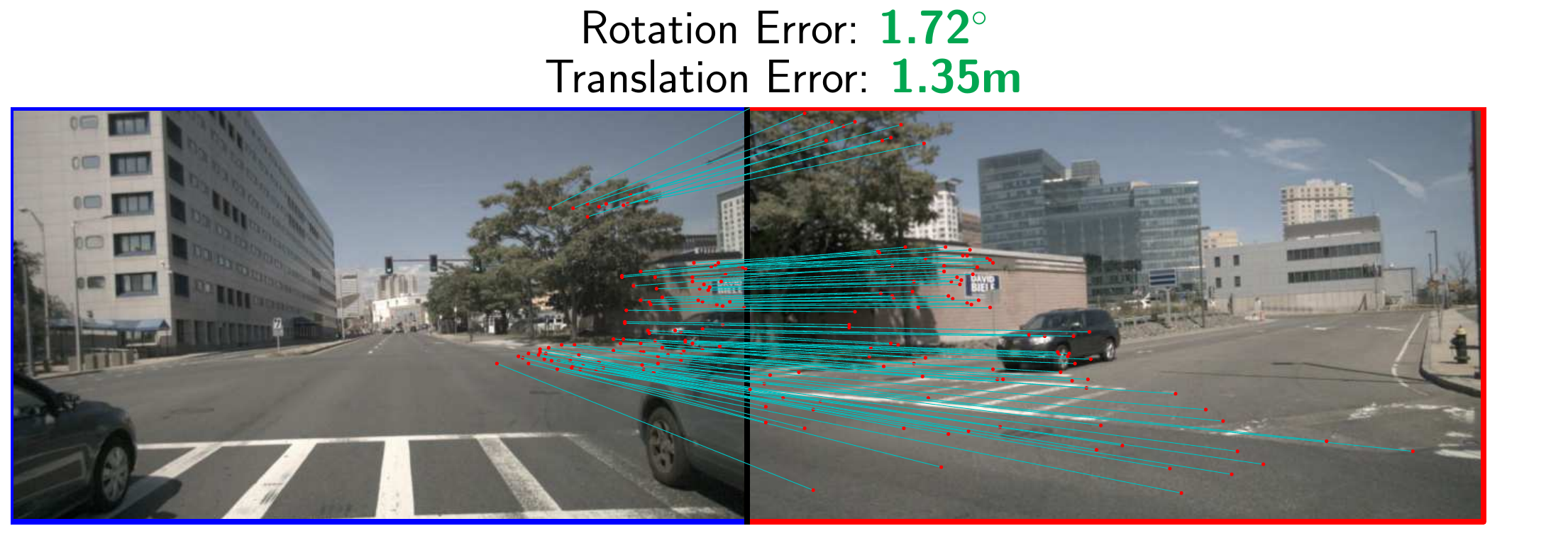}
        \end{subfigure} 
        \\
        \begin{subfigure}{0.3\linewidth}
            \includegraphics[width=\linewidth]{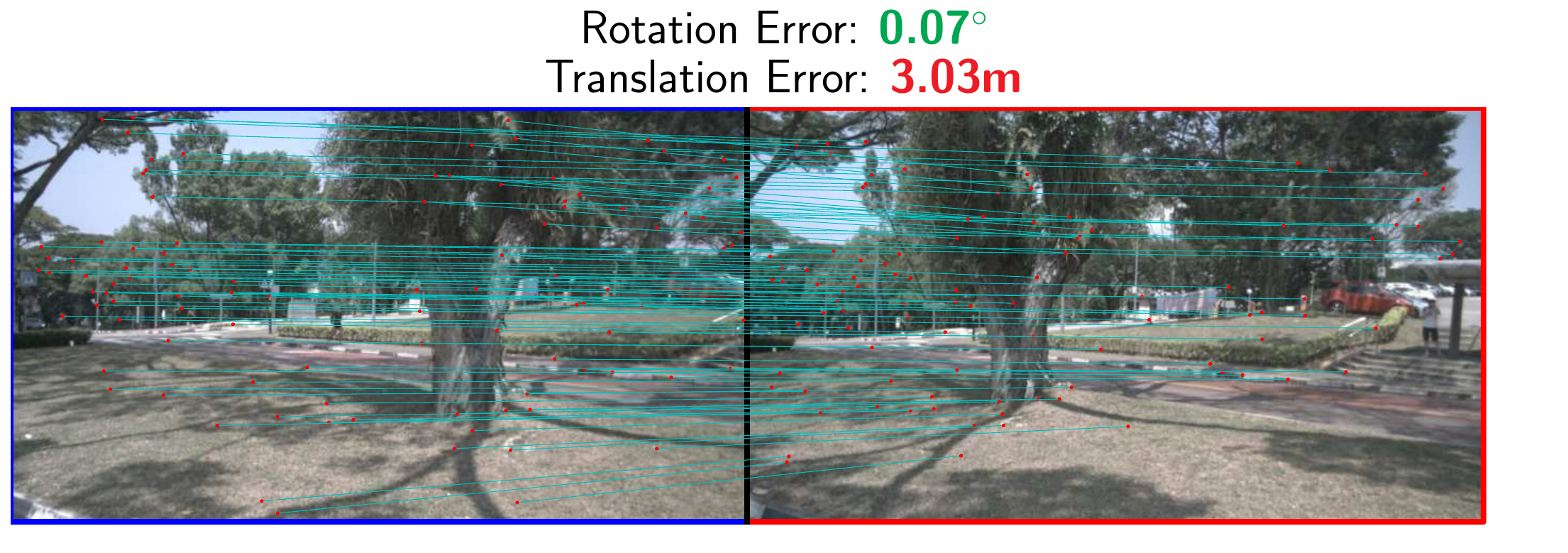}
        \end{subfigure}
        \begin{subfigure}{0.3\linewidth}
            \includegraphics[width=\linewidth]{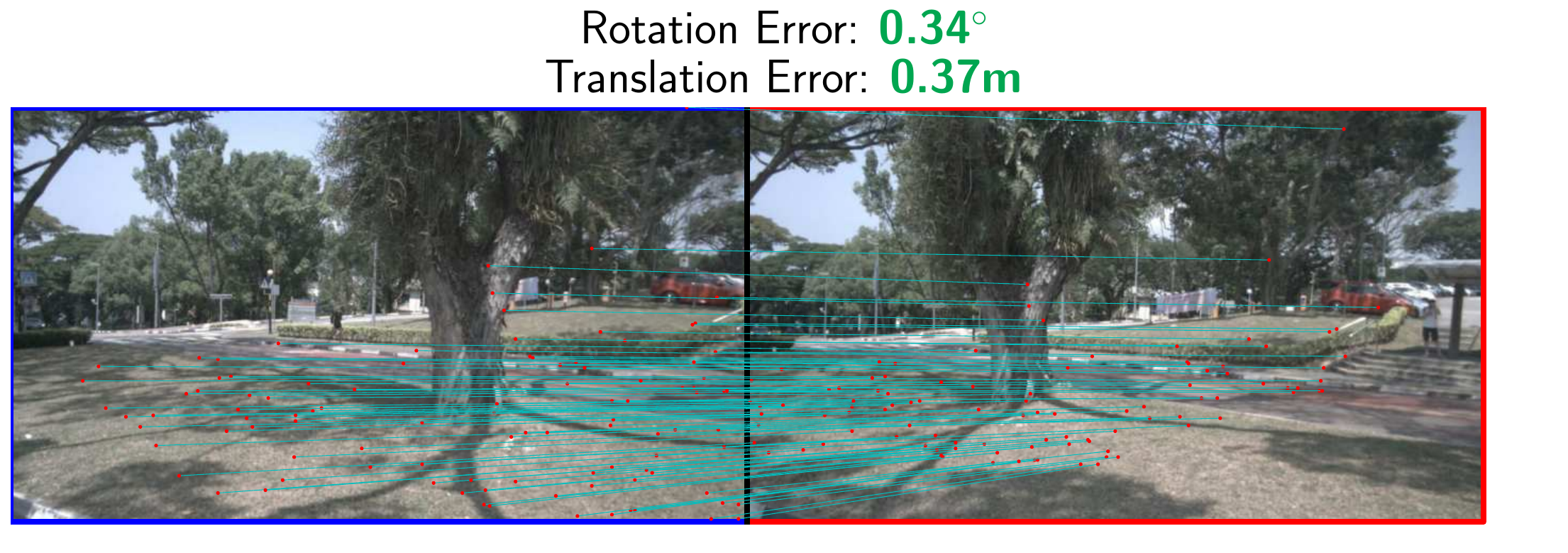}
        \end{subfigure}
        \vspace{-0.3cm}        
    \end{minipage}

    \begin{minipage}{\linewidth}
        \centering
        \caption*{Moderate viewpoint change (30--60°)}
        \begin{subfigure}{0.3\linewidth}
            \includegraphics[width=\linewidth]{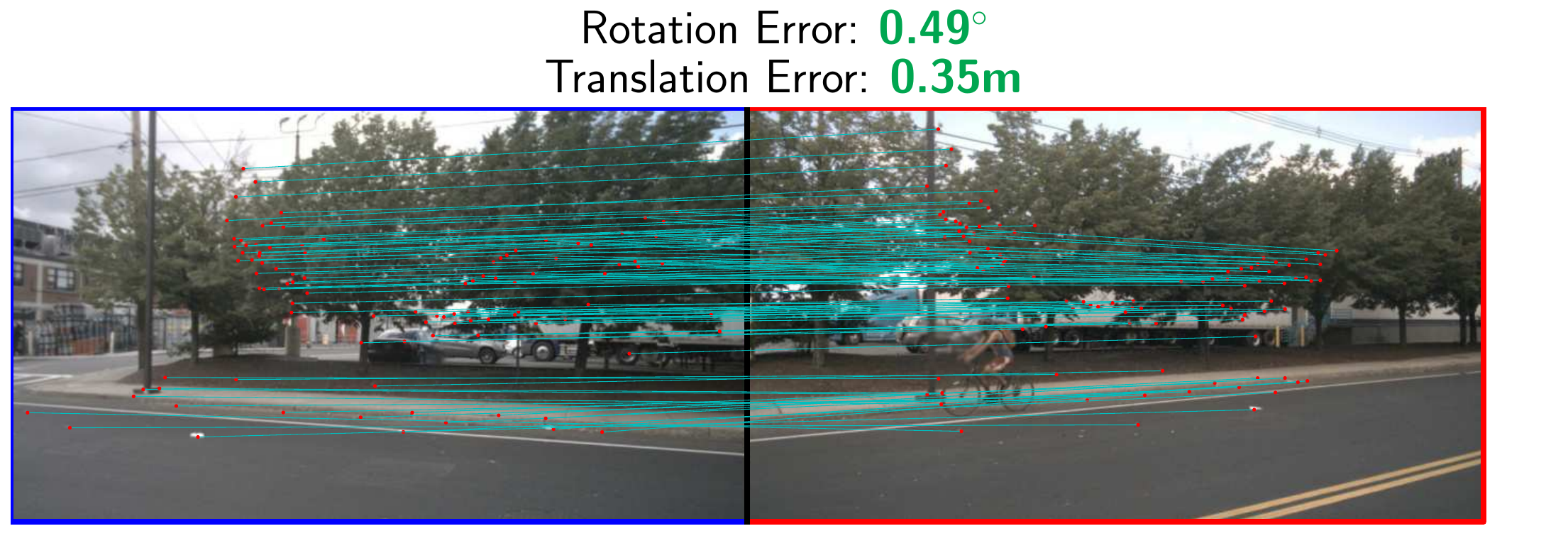}
        \end{subfigure}
        \begin{subfigure}{0.3\linewidth}
            \includegraphics[width=\linewidth]{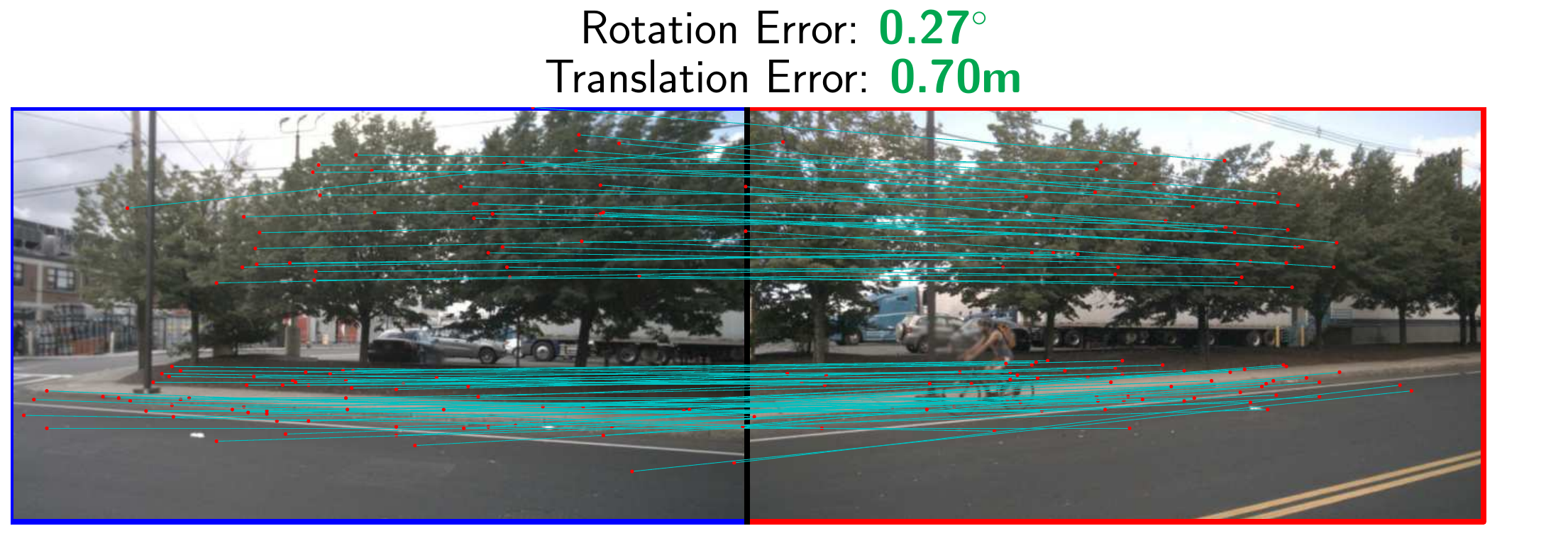}
        \end{subfigure} 
        \\
        \begin{subfigure}{0.3\linewidth}
            \includegraphics[width=\linewidth]{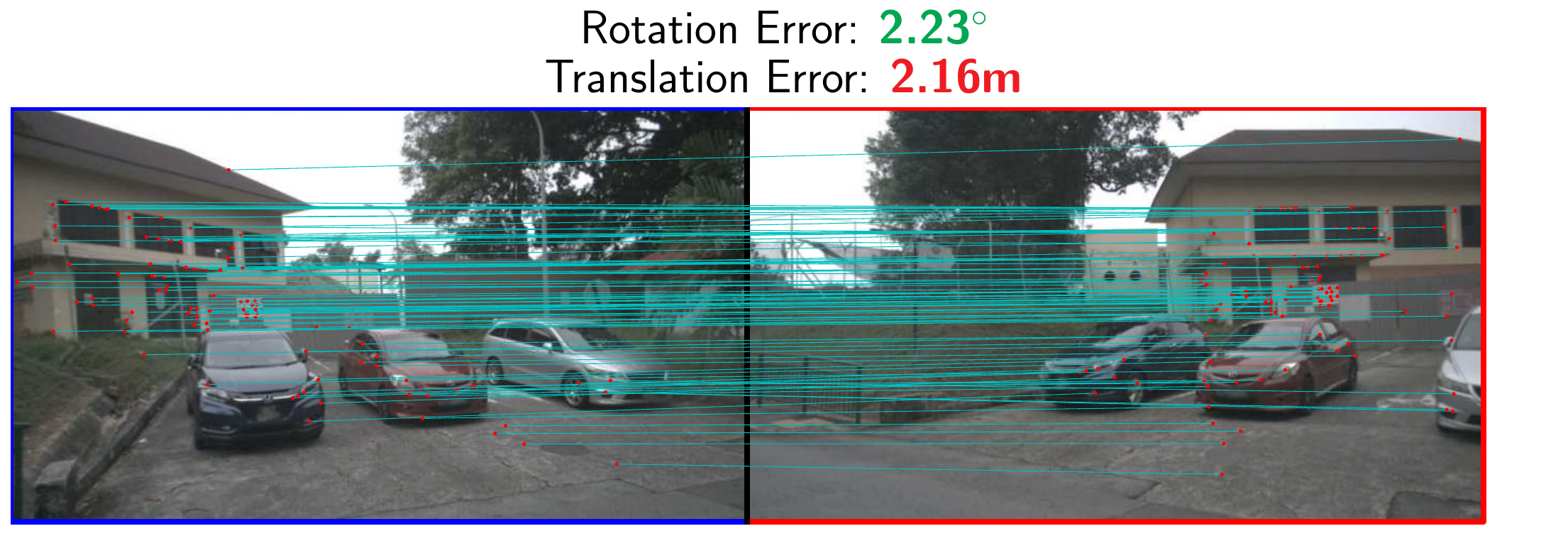}
        \end{subfigure}
        \begin{subfigure}{0.3\linewidth}
            \includegraphics[width=\linewidth]{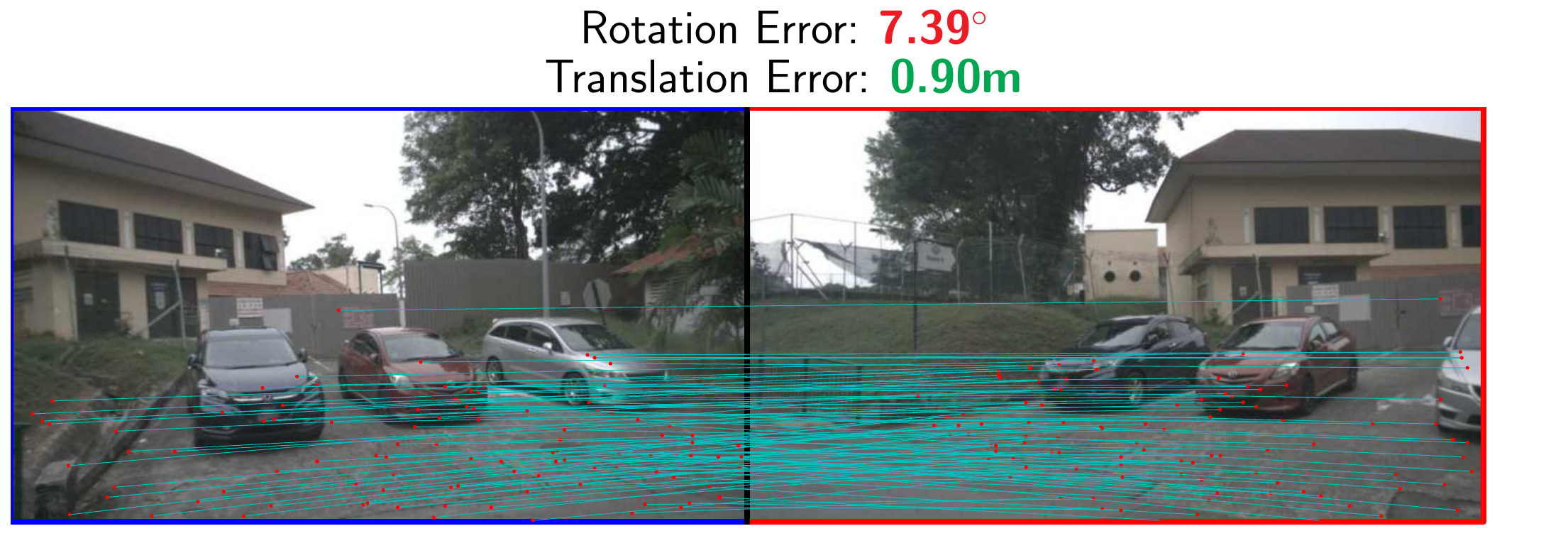}
        \end{subfigure} 
        \vspace{-0.3cm}
    \end{minipage}

    \begin{minipage}{\linewidth}
        \centering
        \caption*{Large viewpoint change (60--120°)}
        \begin{subfigure}{0.3\linewidth}
            \includegraphics[width=\linewidth]{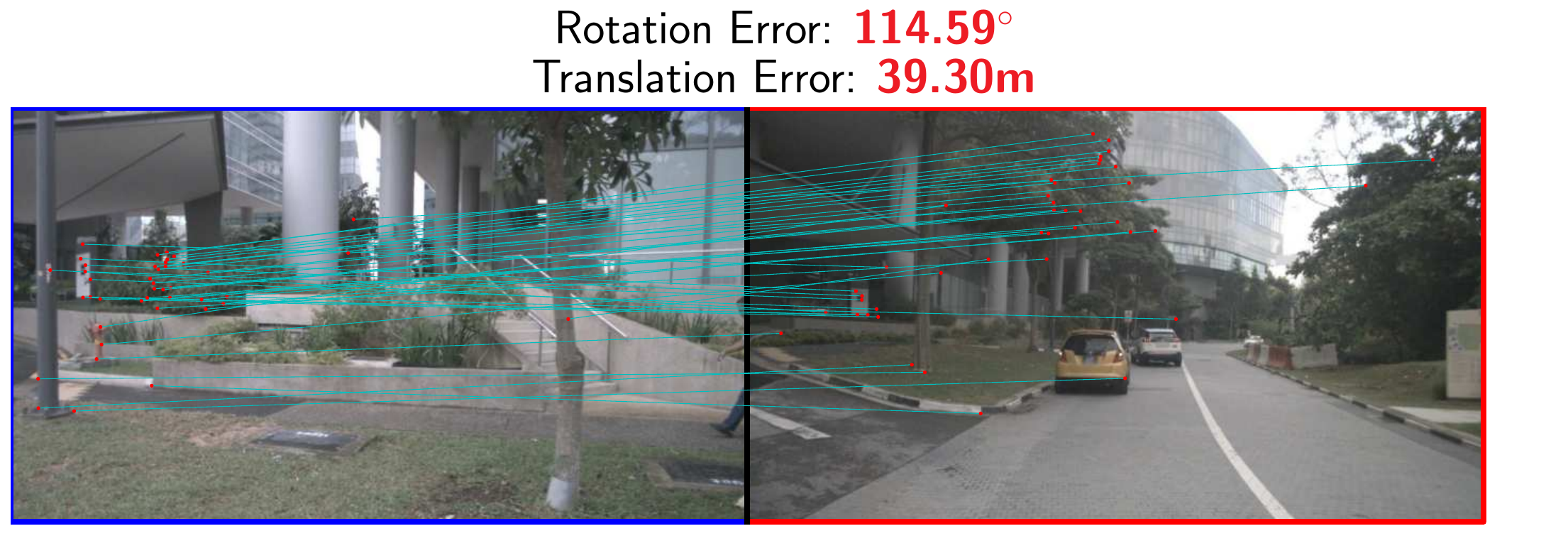}
        \end{subfigure}
        \begin{subfigure}{0.3\linewidth}
            \includegraphics[width=\linewidth]{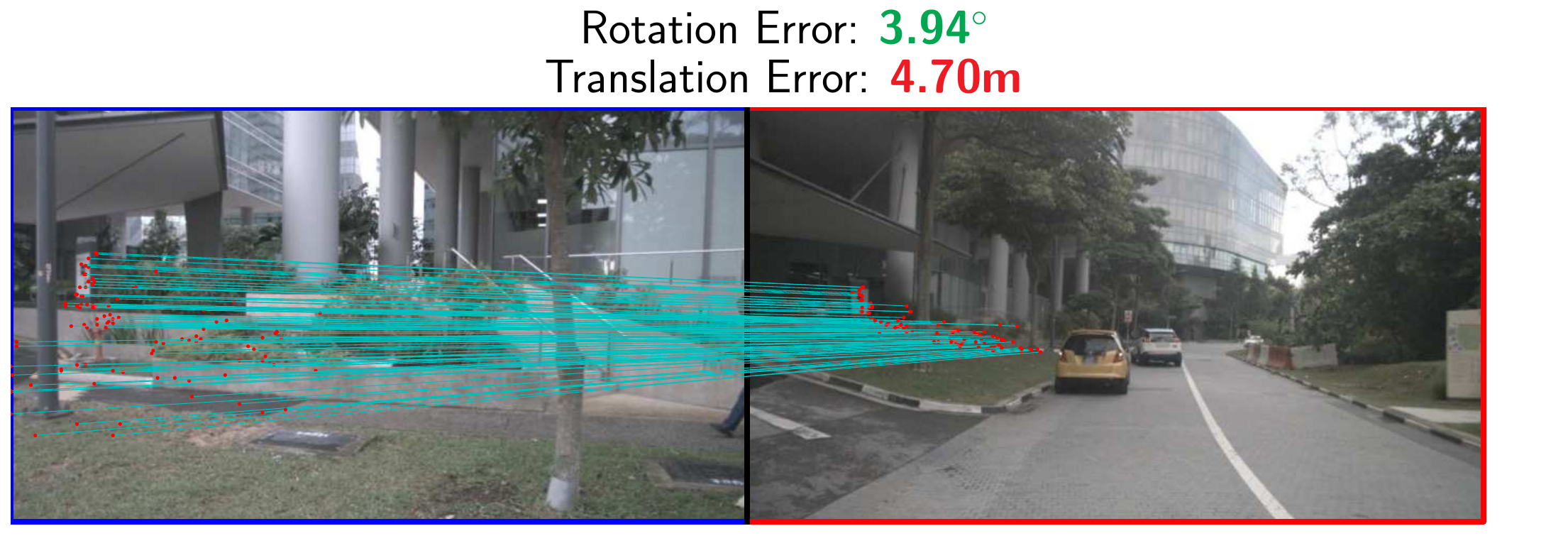}
        \end{subfigure} 
        \\
        \begin{subfigure}{0.3\linewidth}
            \includegraphics[width=\linewidth]{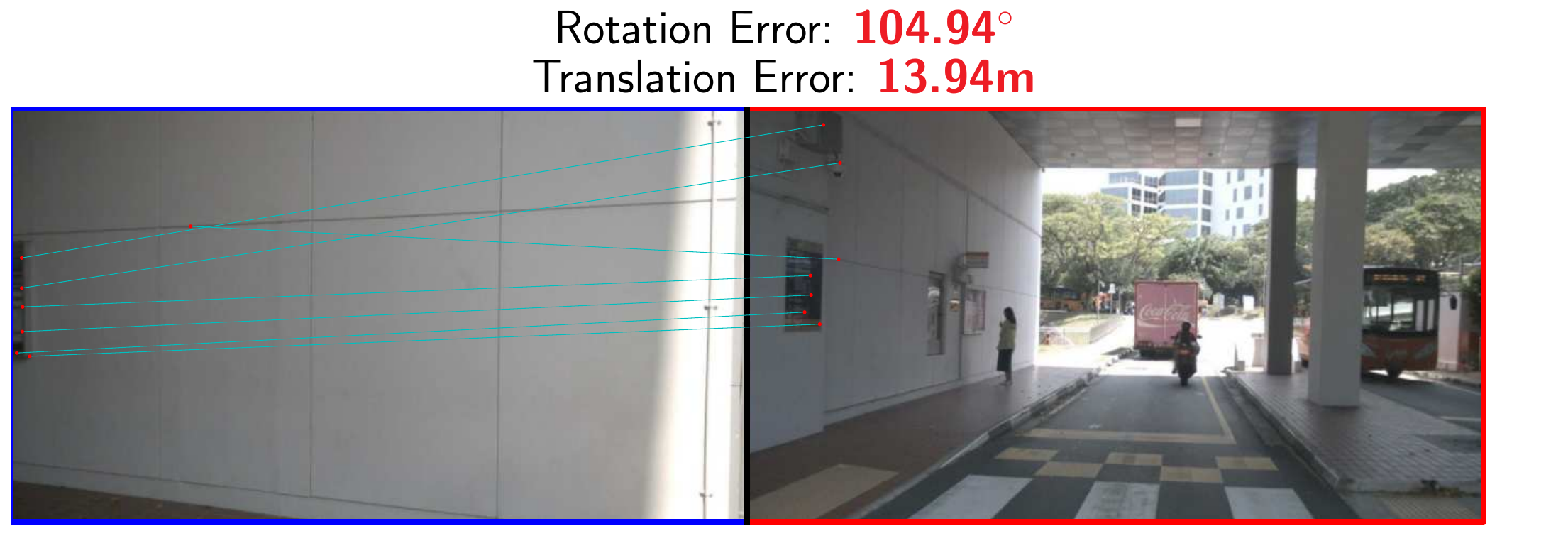}
        \end{subfigure}
        \begin{subfigure}{0.3\linewidth}
            \includegraphics[width=\linewidth]{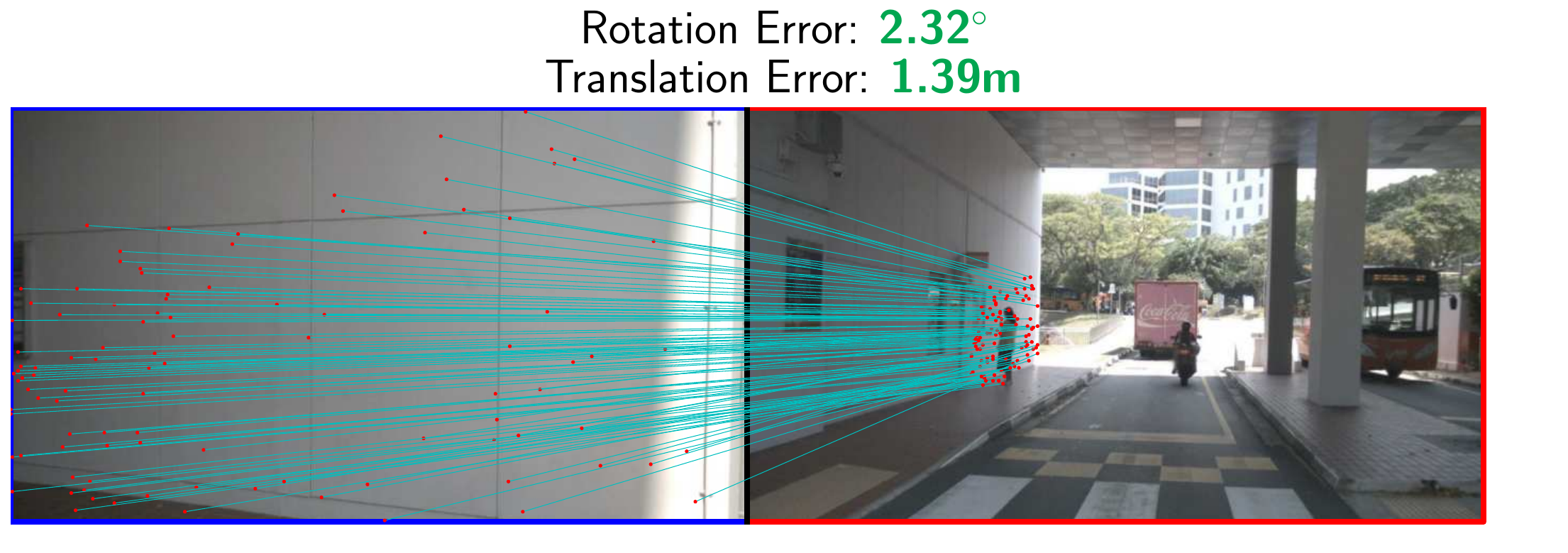}
        \end{subfigure} 
        \vspace{-0.3cm}
    \end{minipage}

    \begin{minipage}{\linewidth}
        \centering
        \caption*{Extreme viewpoint change (120--180°)}
        \begin{subfigure}{0.3\linewidth}
            \includegraphics[width=\linewidth]{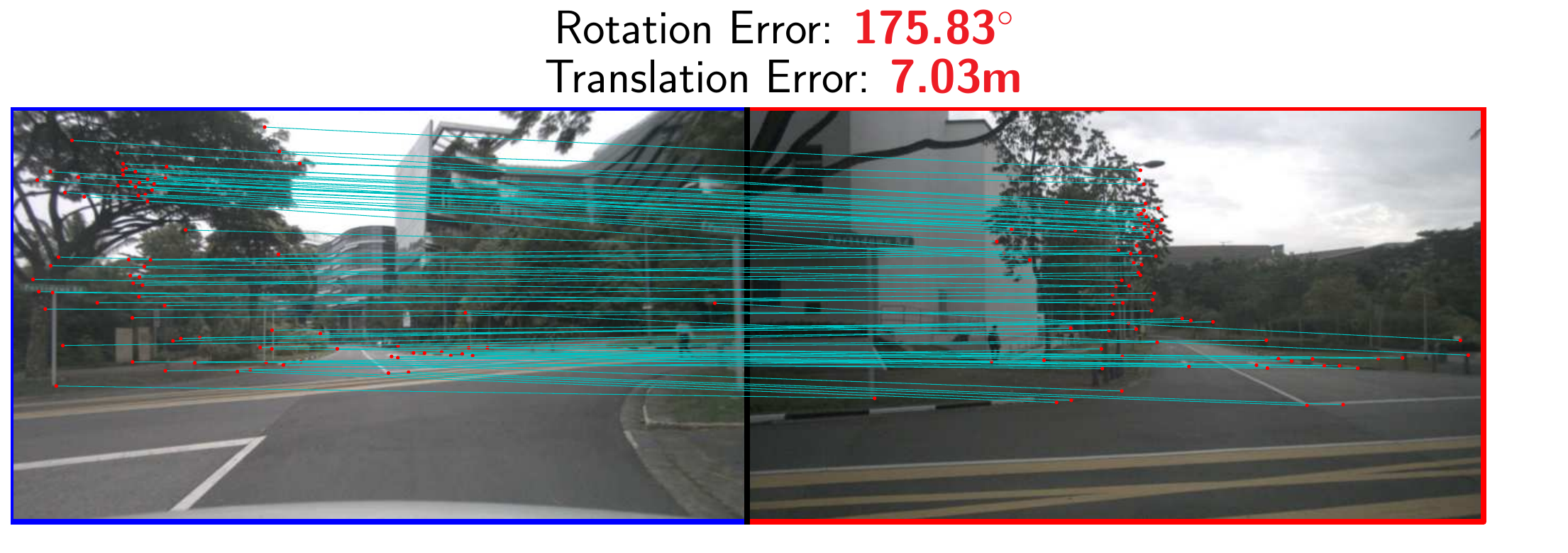}
        \end{subfigure}
        \begin{subfigure}{0.3\linewidth}
            \includegraphics[width=\linewidth]{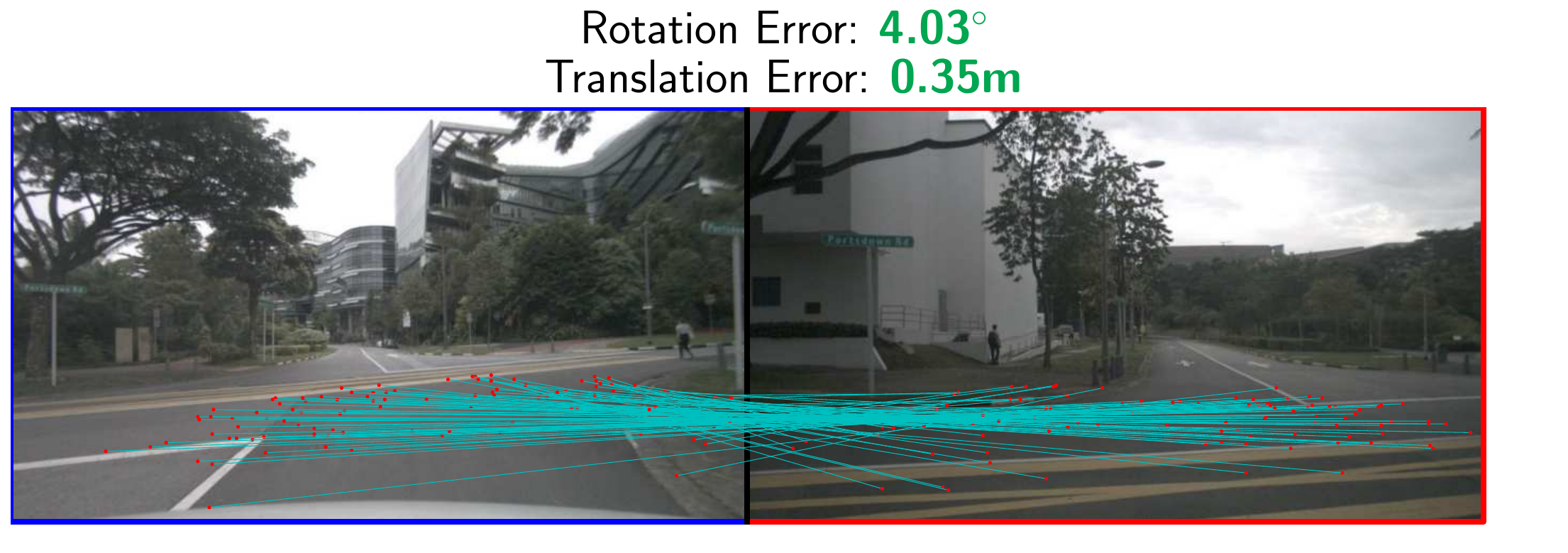}
        \end{subfigure} 
        \\
        \begin{subfigure}{0.3\linewidth}
            \includegraphics[width=\linewidth]{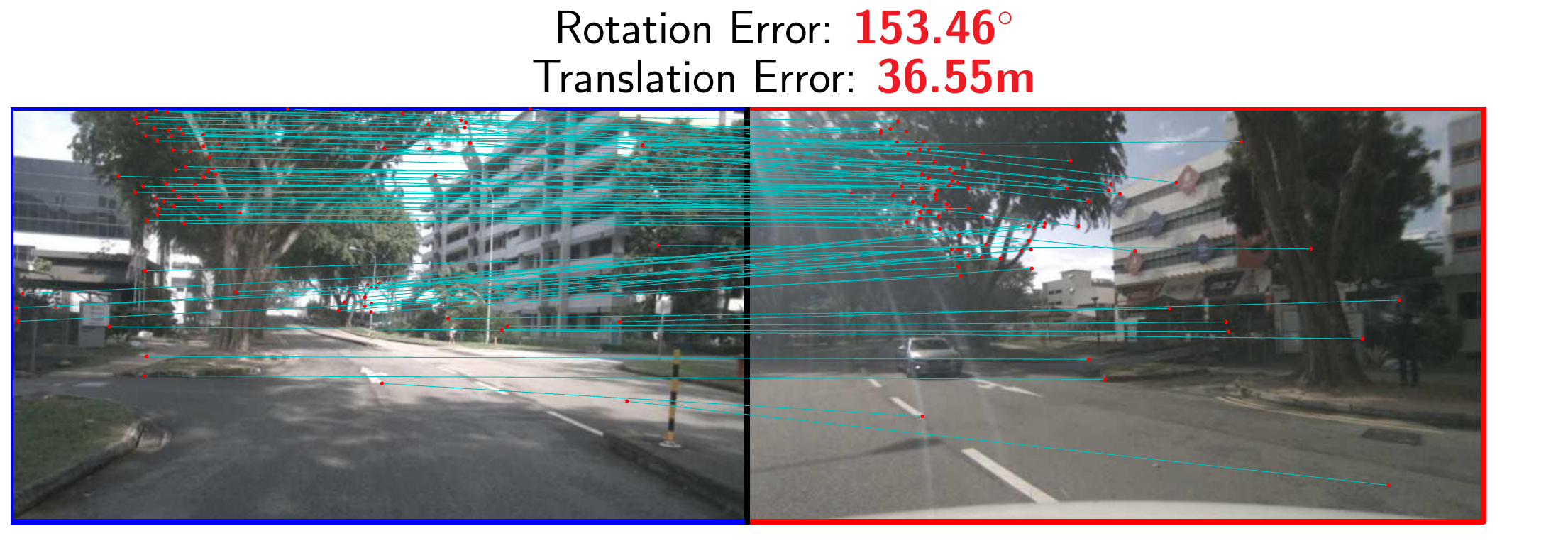}
            \caption{ALIKED+LightGlue}
        \end{subfigure}
        \begin{subfigure}{0.3\linewidth}
            \includegraphics[width=\linewidth]{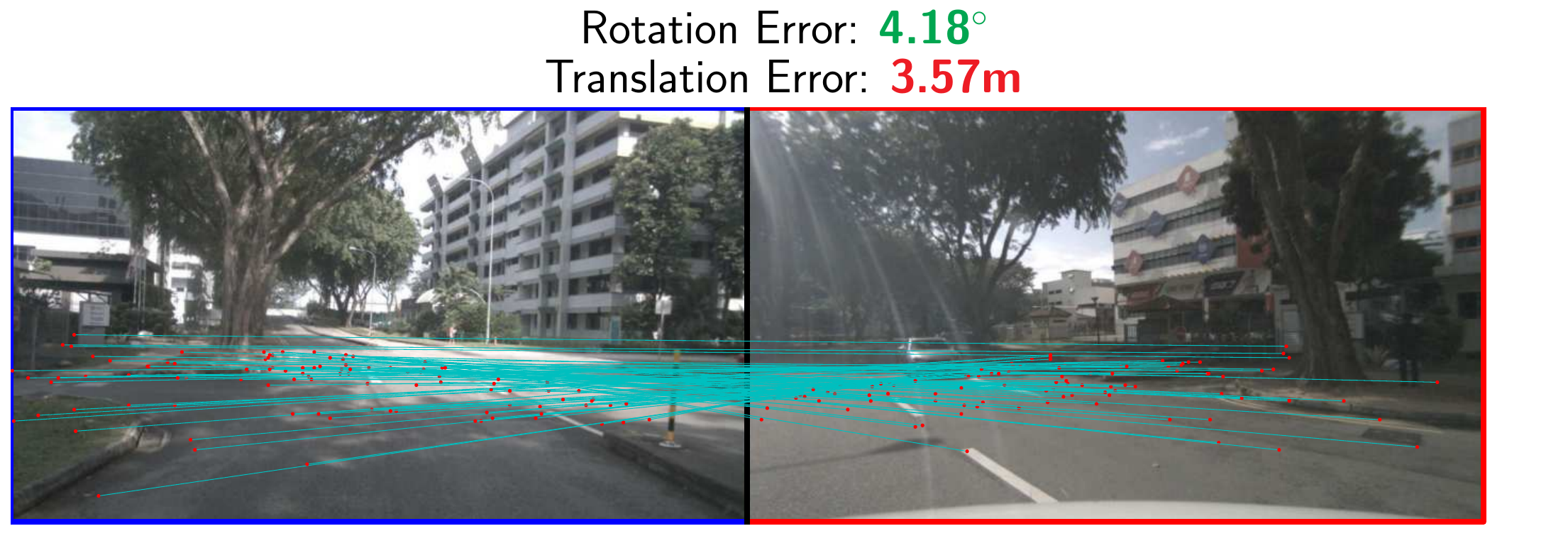}
            \caption{DUSt3R}
        \end{subfigure} 
        \vspace{-0.3cm}
    \end{minipage}
    
    \caption{\textbf{Examples of image pairs with varying viewpoint angles -- }For each viewpoint angle range, we show two random image pairs for the best methods in either detector-based (ALIKED+LightGlue on the left) or detector-free (DUSt3R on the right) approaches.}
    \label{fig:angle_examples}
\end{figure*}

We provide visual examples of image pairs for each geometric criterion bin, along with 100 randomly sampled matches from different methods in~\cref{fig:overlap_examples,fig:scale_examples,fig:angle_examples}. For each bin, we show results on two image pairs, from the two best methods in either detector-based (ALIKED+LightGlue) or detector-free (DUSt3R) approaches.

